\pdfoutput=1
\documentclass[11pt]{article}

\usepackage{emnlp2021}

\usepackage{times}
\usepackage{latexsym}

\usepackage[T1]{fontenc}
\usepackage[utf8]{inputenc}
\usepackage{CJKutf8} % for Japanese

\usepackage{microtype}

\usepackage{graphicx}
\usepackage{booktabs} % for professional tables

\usepackage{hyperref}

\usepackage{url}
\usepackage{latexsym}
\usepackage{multirow}
\usepackage{xspace}
\usepackage{caption}
\usepackage{subcaption}

\usepackage{amssymb}  % for checkmark
\newcommand{\cmark}{\checkmark}

\newcommand{\xtreme}{\textsc{xtreme}\xspace}
\newcommand{\xglue}{\textsc{xglue}\xspace}
\newcommand{\name}{\mbox{\textsc{xtreme-r}}\xspace}
\newcommand{\checklist}{\textsc{CheckList}\xspace}
\newcommand{\multichecklist}{\textsc{MultiCheckList}\xspace}
\newcommand{\explainaboard}{\textsc{ExplainaBoard}\xspace}
\definecolor{brinkpink}{rgb}{0.98, 0.38, 0.5}
\definecolor{aogreen}{rgb}{0.0, 0.5, 0.0}
\definecolor{ballblue}{rgb}{0.13, 0.67, 0.8}

\definecolor{ballblue}{rgb}{0.13, 0.67, 0.8}
\definecolor{brinkpink}{rgb}{0.98, 0.38, 0.5}

\title{XTREME-R: Towards More Challenging \\and Nuanced Multilingual Evaluation}

\author{Sebastian Ruder$^{1}$, Noah Constant$^{2}$, Jan Botha$^{2}$, Aditya Siddhant$^{2}$, Orhan Firat$^{2}$, \\
        {\bf Jinlan Fu$^{3}$, Pengfei Liu$^{4}$, Junjie Hu$^{4}$, Dan Garrette$^{2}$, Graham Neubig$^{4}$, Melvin Johnson$^{2}$} \\
        $^1$DeepMind $\:$ $^2$Google Research $\:$ $^3$Fudan University $\:$ $^4$Carnegie Mellon University \\
        }

\begin{document}
\maketitle
\begin{abstract}
Machine learning has brought striking advances in multilingual natural language processing capabilities over the past year. For example, the latest techniques have improved the state-of-the-art performance on the \xtreme multilingual benchmark by more than 13 points. While a sizeable gap to human-level performance remains, improvements have been easier to achieve in some tasks than in others. This paper analyzes the current state of cross-lingual transfer learning and summarizes some lessons learned. In order to catalyze meaningful progress, we extend \xtreme to \name, which consists of an improved set of ten natural language understanding tasks, including challenging language-agnostic retrieval tasks, and covers 50 typologically diverse languages. In addition, we provide a massively multilingual diagnostic suite (\multichecklist) and fine-grained multi-dataset evaluation capabilities through an interactive public leaderboard to gain a better understanding of such models.
\footnote{The leaderboard and code for \name will be made available at \url{https://sites.research.google/xtreme} and \url{https://github.com/google-research/xtreme} respectively.} 
\end{abstract}

\section{Introduction}

% \gn{In general, I really like to have a figure or table on the top right of the first page visually highlighting the new stuff in the paper to catch the reader's eye. For example, it could be a figure or table contrasting the features of XTREME (and XGLUE?) with XTREME-R, maybe with an ``Analysis'' arrow pointing from the former to the latter to demonstrate that the analysis was the basis for the changes? Just a suggestion, not mandatory by any means.}

% \lr{The conclusion talks about driving meaningful progress, avoiding easy victories, and encouraging learning of cross-linguistic syntax and semantics. Would be great to frame in the introduction what this means. How do we know when meaningful progress has been made, using the tasks and diagnostics available in XTREME or XTREME-R? Is it with success on certain task types, or across the board, or certain languages, or with certain methods? This would help with the analysis of retrieval results in Section 2.2. where I struggled to know whether you thought such progress has been made or not.}

% \lr{If it seems reasonable to you, my preference would also be to see the zero-shot evaluation setting mentioned earlier than Section 4, and for it to be defined in the XTREME-R context (e.g. any general multilingual training data is allowed, task fine-tuning is English only). This could be in the introduction or in Section 2 (maybe 2.2). It's probably worth assuming some readers are not already familiar with XTREME or its component tasks.}

Most research in natural language processing (NLP) to date has focused on developing methods that work well for English and a small set of other high-resource languages \citep{Joshi2020}. In contrast, methods for other languages can be vastly more beneficial as they enable access to language technology for more than three billion speakers of low-resource languages and prevent the NLP community from overfitting to English. Motivated by these benefits, the area of multilingual NLP has attracted increasing interest recently. 
% Most state-of-the-art models \pfliu{Do we need some typical refs here?} focus on learning general-purpose multilingual representations based on Transformer \citep{Vaswani2017} models pre-trained on massive amounts of multilingual data \gn{This sentence seemed like it could be skipped if you want to save space}.

\begin{table}[t]
\centering
\resizebox{\columnwidth}{!}{%
\begin{tabular}{lll}
\toprule
 & \xtreme & \name \\ \midrule
\vspace{2pt}\# of languages & 40 & 50 \\
\vspace{2pt}\# of tasks & 9 & $-$ 2 + 3 = 10 \\
\vspace{2pt}\begin{tabular}[l]{@{}l@{}}Task categories\\~ \end{tabular} & 
\begin{tabular}[l]{@{}l@{}}Classification, structured\\ prediction, QA, retrieval\end{tabular} & 
\begin{tabular}[l]{@{}l@{}}+language-agnostic\\ retrieval\end{tabular} \\
\vspace{2pt}\begin{tabular}[l]{@{}l@{}}Analysis tools\\~ \end{tabular} & 
\begin{tabular}[l]{@{}l@{}}---\\~ \end{tabular} & 
\begin{tabular}[l]{@{}l@{}}\multichecklist,\\ Explainaboard\end{tabular} \\
\begin{tabular}[l]{@{}l@{}}Leaderboard \end{tabular} & 
\begin{tabular}[l]{@{}l@{}}Static \end{tabular} & 
\begin{tabular}[l]{@{}l@{}}Interactive, +metadata\end{tabular} \\
\bottomrule
\end{tabular}%
}
\caption{Overview of \xtreme and \name.
% \nc{This table is hard to read. Maybe add separating divider rules, or extra vertical space between the rows?}\dhg{I top-left-aligned the text, added space between rows, and reduced the number of linebreaks. What do we think?} \sr{Thanks! This looks good to me. Let me know if anything should still be changed.} 
}
\label{tab:xtreme-vs-xtreme-r}
\end{table}

However, evaluating multilingual models is challenging as it requires assessing performance on a wide range of typologically distinct languages in the face of limited heterogeneous data sources. Recently large-scale benchmarks such as \xtreme \citep{Hu2020xtreme} and \textsc{xglue} \cite{liang2020xglue} have been introduced, consolidating existing multilingual tasks and covering tens of languages.
% Recently, the \xtreme benchmark \citep{Hu2020xtreme}, which consolidates nine existing tasks covering 40 languages has been introduced for the evaluation of state-of-the-art multilingual models \gn{Cite XGLUE also for completeness, it might make the paper seem a little bit less focused only on XTREME and emphasize the general interestingness?}. 
When \xtreme was released, the gap between the best-performing baseline, XLM-R Large \citep{Conneau2020xlmr}, and human-level performance was roughly 25. This has since shrunk to less than 12 points, a much smaller but still substantial gap compared to the difference from human-level performance observed in English transfer learning \citep{Wang2019superglue}, which has recently been closed entirely on some evaluation suites \cite{He2021deberta}.
% \gn{changed}.

In order to examine the nature of this progress, we first perform an analysis of state-of-the-art multilingual models on \xtreme. We observe that progress has not been uniform, but concentrated on cross-lingual retrieval tasks where fine-tuning on other tasks and pre-training with parallel data lead to large gains. On other task categories improvements are more modest. Models still generally perform poorly on languages with limited data and non-Latin scripts. Fine-tuning on additional translated data generally leads to the best performance.
% \gn{Added a paragraph break here to highlight the analysis contributions a little more. In doing so, it does seem that the selling of the analysis results is a little bit weaker than the other parts. From the current writing seems that the only result of the ``extensive'' analysis was the identification that particular tasks had improved more than others?}.

Based on this analysis, we propose \name (\textsc{xtreme} \textsc{r}evisited), a new benchmark with the dual purpose of ensuring that research in multilingual NLP focuses on the most challenging problems and equipping researchers with a broader set of tools to better understand their models (see Table \ref{tab:xtreme-vs-xtreme-r} for a brief overview). 
% \gn{changed}. 
\name follows in its predecessor's footsteps by being massively multilingual, diverse, and accessible. It expands on \xtreme by covering 50 typologically diverse languages and 10 challenging, diverse tasks. To make retrieval more difficult, we introduce two new tasks that focus on ``language-agnostic'' retrieval \cite{roy-etal-2020-lareqa}, where targets must be retrieved from a large multilingual candidate pool. We additionally establish new state-of-the-art mT5 \cite{Xue2020mt5} and translate-train baselines for our tasks.

\name aims to move away from a single aggregate metric summarizing a model's performance and towards a more nuanced evaluation and comparison of multilingual models \cite{ethayarajh2020utility, linzen2020can}. To this end, we introduce an extensible multilingual diagnostic and evaluation suite that consists of two main components: a) \multichecklist, a test suite \citep{Ribeiro2020checklist} for probing question answering capabilities in 50 languages. This test suite is the first of its kind and enables direct evaluation of fine-grained capabilities in a massively multilingual setting. b) We extend the multi-dataset evaluation framework \explainaboard \citep{Fu2020explainaboard,liu2021explainaboard} to additional tasks and the multilingual setting. This framework allows us to break down performance based on language and task-specific attributes, which enables a more nuanced diagnosis of a model's behaviour.

We also make several logistic improvements to improve \name's utility as a leaderboard. 
% \gn{I added this sentence to summarize the paragraph}.
To make it easier to choose the best model for a use case, each submission is required to provide metadata such as the number of parameters and amount of pre-training data,
% . We collect this information for existing models and make it
which we make available via an interactive leaderboard. We also introduce task and language-specific sub-leaderboards to invite submissions of dedicated models.

% \gn{Added a paragraph break here to split the discussion of logistics and discussion of overall contributions.}
In sum, we make the following contributions: a) an analysis of progress in cross-lingual modeling; b) an improved benchmark covering 50 languages, including a newly created retrieval task (Mewsli-X); c) a massively multilingual diagnostic suite; d) fine-grained evaluation capabilities; e) experiments and analyses of state-of-the-art models; and f) an interactive metadata-rich leaderboard. 
\section{Examining the State of Multilingual Benchmarking}

\subsection{Background}

% \gn{I feel the structure of this paragraph could be improved. Maybe describe the need for benchmarking first, motivated by GLUE and SuperGLUE. Then mention limitations of benchmarking on English only, which motivates the need for benchmarks in other languages, and also multilingual benchmarks.}
% \mj{Took a stab at restructuring this. ruder@ please refine this as needed.}
Benchmarking is critical to evaluate general-purpose language understanding technologies. To this end, benchmarks like GLUE \citep{Wang2019glue} and SuperGLUE \citep{Wang2019superglue} provide a way to assess the transfer learning capabilities of various models. However, these benchmarks focus only on English. On the other hand, cross-lingual approaches have been evaluated on a wide range of disparate tasks \cite{Hu2020xtreme}. \xtreme was proposed as a platform to unify this fragmented evaluation landscape and to catalyze advances in cross-lingual learning by including a diverse set of tasks and languages. It consists of 9 tasks covering 40 diverse languages, which can be grouped into 4 broad task types (see \textsection \ref{sec:retained} for details): classification (XNLI, PAWS-X),  structured prediction (UD-POS, WikiANN-NER), question answering (XQuAD, MLQA, TyDiQA-GoldP), and retrieval (Tatoeba, BUCC). \xtreme focuses on \emph{zero-shot cross-lingual transfer}, i.e. models can be pre-trained on any multilingual data and are fine-tuned only in English.
Similarly, \xglue \citep{liang2020xglue}, another cross-lingual benchmark focuses on a smaller number of less typologically diverse languages, and includes generation tasks. Other non-English benchmarks focus on specific linguistic phenomena, e.g.~code-switching \citep{Khanuja2020gluecos}; languages, e.g.~Indonesian \citep{Willie2020indonlu} and Persian \citep{Khashabi2020parsinlu}; and language families, e.g.~Indian languages \cite{kakwani-etal-2020-indicnlpsuite}. 

\subsection{An Analysis of \xtreme}

% \gn{If you add the above-suggested explanation I would suggest adding a subsection break here to split the ``Background'' and ``An Analysis of XTREME'' subsections and make our actual contributions more clear.}

% So far, progress both in cross-lingual transfer and representation learning and on \xtreme has largely mirrored its English counterpart: Early approaches focused on learning cross-lingual word embeddings \citep{mikolov2013exploiting,gouws2015bilbowa,Artetxe2017learning,Ruder2019survey}. Recently, advances in monolingual transfer learning \citep{howard2018universal,peters2018deep,Devlin2019bert} have been extended to the multilingual setting \citep{eriguchi2018zero,Pires2019,Wu2019,Lample2019xlm,Siddhant2019evaluating}. On \xtreme, XLM-R \citep{Conneau2020xlmr}, a multilingual version of RoBERTa \citep{liu2019roberta} was initially the best-performing model. Subsequent improvements were achieved by intermediate-task training \citep{phang2020english}.

\begin{figure*}[!h]
\centering
\begin{subfigure}{\columnwidth}
  \centering
  \includegraphics[width=\columnwidth]{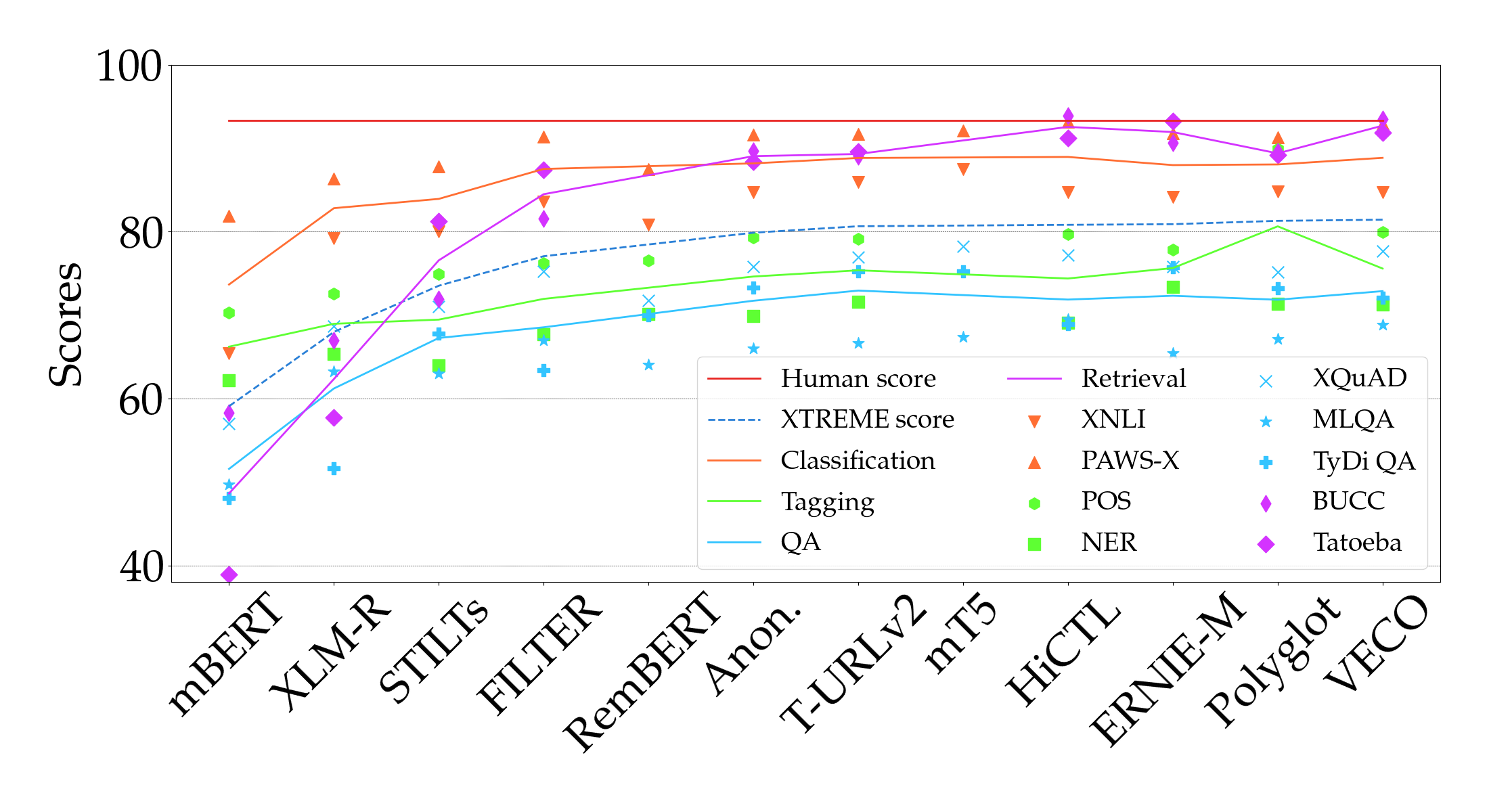}
  \caption{Performance on \xtreme}
  \label{fig:xtreme_performance}
\end{subfigure}%
\begin{subfigure}{\columnwidth}
  \centering
  \includegraphics[width=\columnwidth]{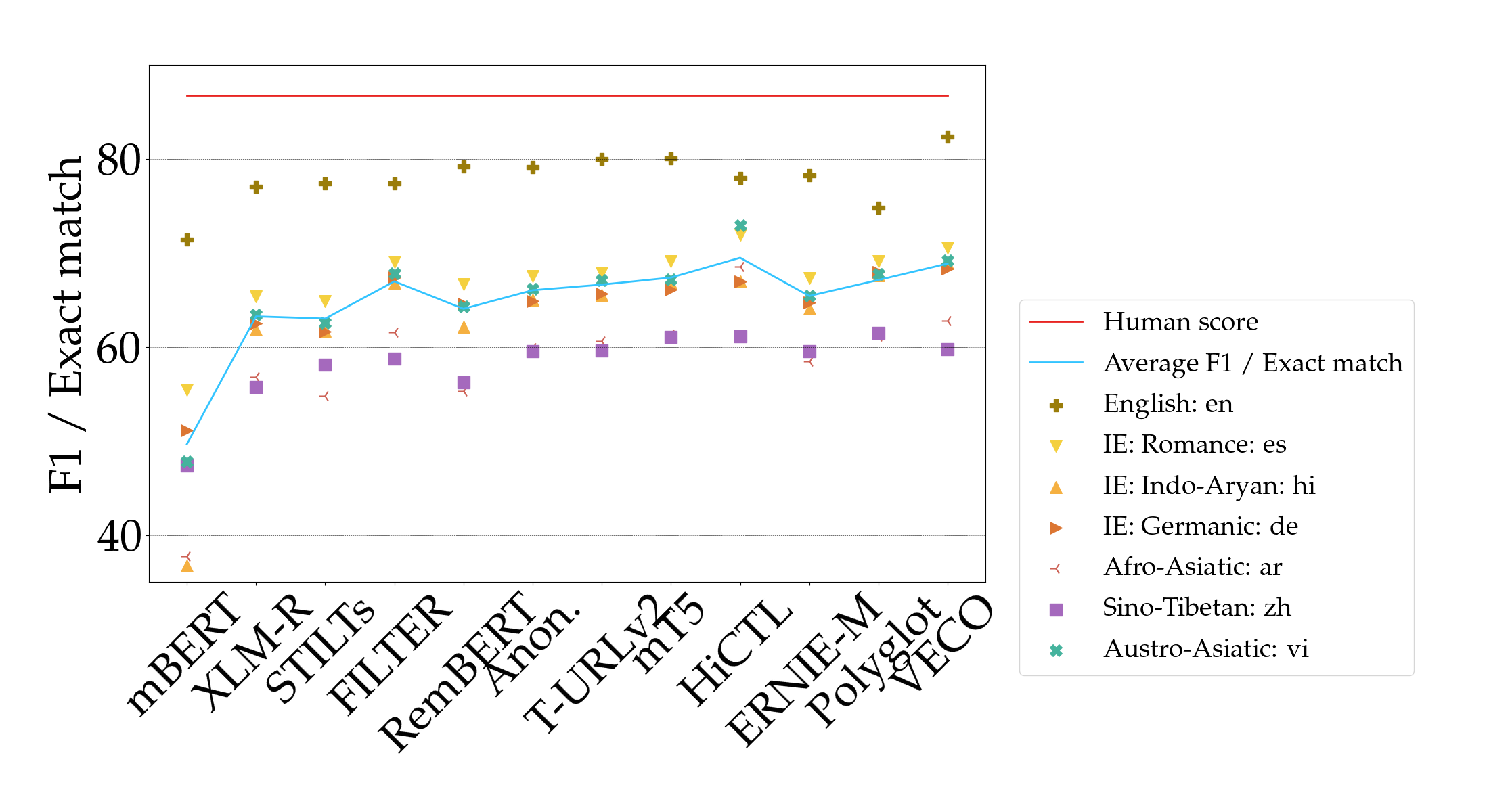}
  \caption{Performance on MLQA}
  \label{fig:xtreme_mlqa_performance}
\end{subfigure}
\begin{subfigure}{\columnwidth}
  \centering
  \includegraphics[width=\linewidth]{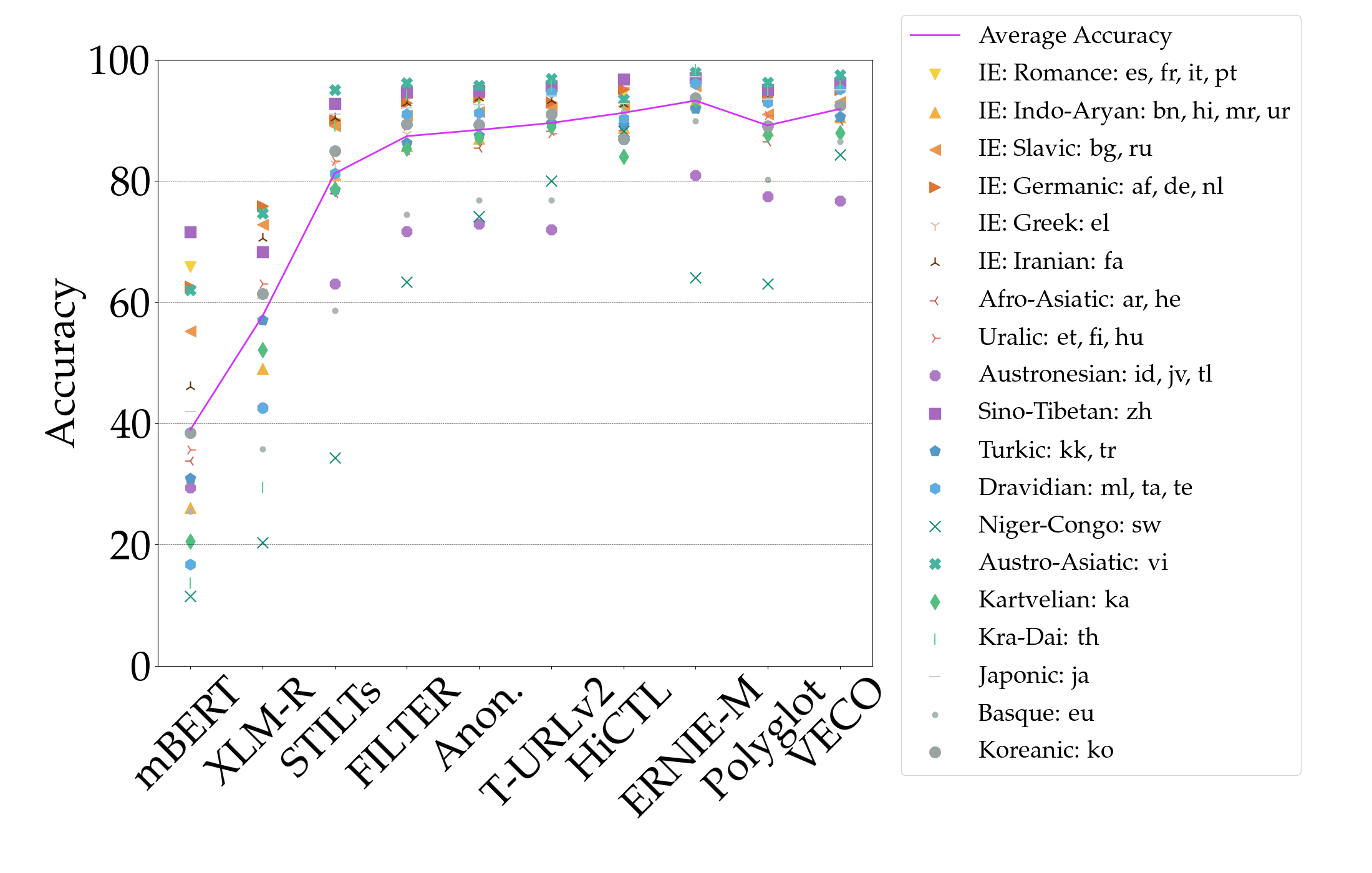}
  \caption{Performance on Tatoeba}
  \label{fig:xtreme_tatoeba_performance}
\end{subfigure}%
\begin{subfigure}{\columnwidth}
  \centering
  \includegraphics[width=\linewidth]{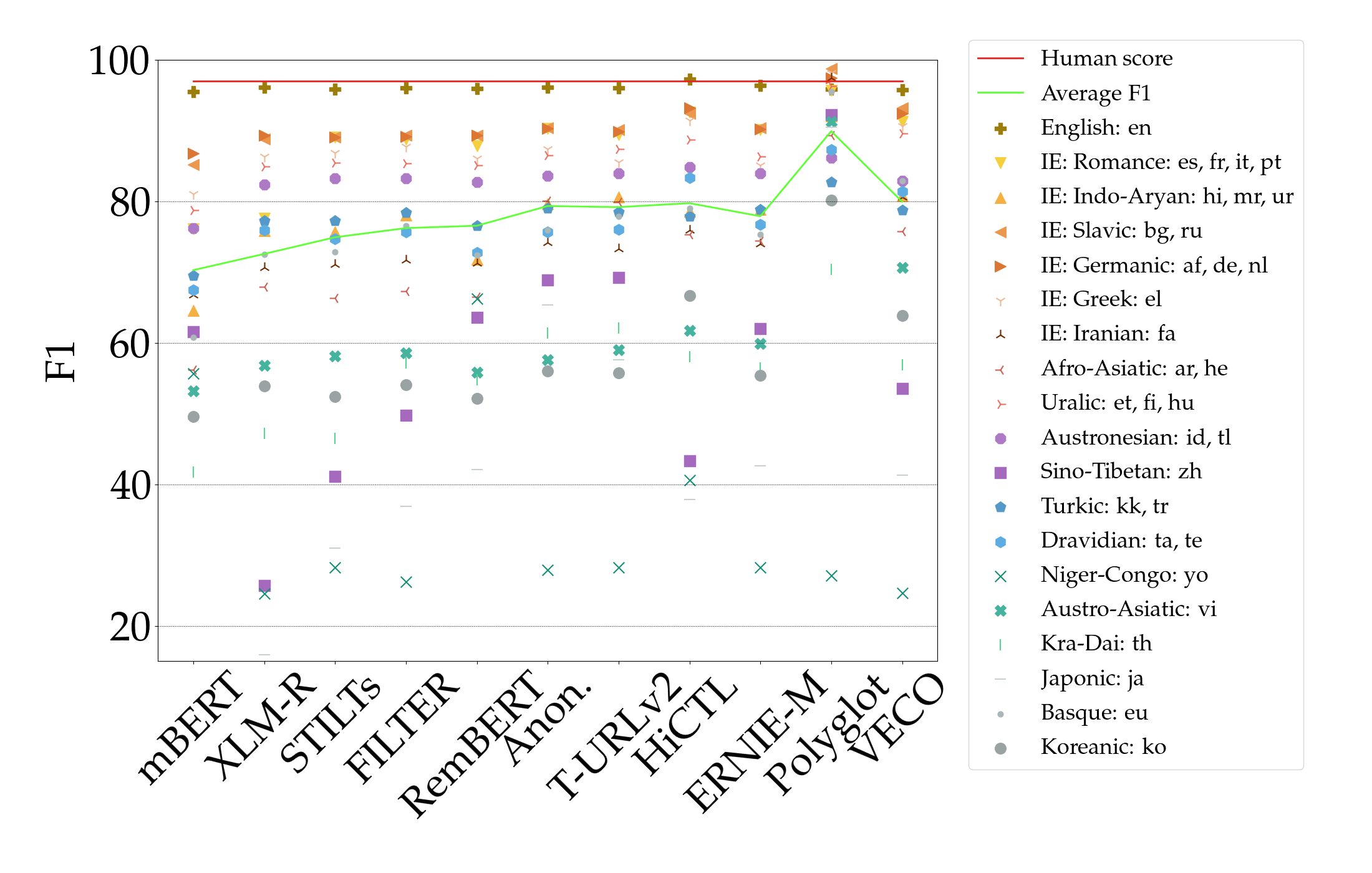}
  \caption{Performance on UD-POS}
  \label{fig:xtreme_pos_performance}
\end{subfigure}
% \vspace{-0.8cm}
\caption{Performance of models (a) on the \xtreme leaderboard across all nine \xtreme tasks, (b) on the MLQA question answering dataset, (c) on the Tatoeba retrieval task, and (d) on Universal Dependencies POS tagging across language families. Models are ordered based on their XTREME score (a). Results of models that do not evaluate on a task category are omitted, i.e. RemBERT for retrieval and mT5 for retrieval and tagging. 
% \dhg{How was the left-to-right ordering determined? Date the model was released?} \sr{Based on average score on XTREME. I'll make this clear.}
}
\label{fig:xtreme_leaderboard_analysis}
\end{figure*}

% Progress in cross-lingual transfer and representation learning and on \xtreme has largely mirrored its English counterpart. 
As of April 15, 2021, all submissions to the \xtreme leaderboard are large-scale Transformers \citep{Vaswani2017} trained with masked language modeling (MLM; see Appendix \ref{app:xtreme-model-details} for further details).
We analyze the performance of these models on the \xtreme leaderboard in Figure \ref{fig:xtreme_leaderboard_analysis}.\footnote{This evaluation compares ``models + data'' as models employ different types of data during training. We document this information in Appendix \ref{app:metadata}.} Overall, multilingual models have improved the average performance on \xtreme from 55.8 to 81.4. Much of this improvement is concentrated in the retrieval-based tasks where performance increased from 47.7 (mBERT) to 92.7 (\mbox{VECO}). In contrast, performance on question answering and structured prediction tasks has improved only slightly.

\begin{table*}[]
\centering
\resizebox{\textwidth}{!}{%
\begin{tabular}{l l r r r r r l l l}
\toprule
Task category & Task & $|$Train$|$ & $|$Dev$|$ & $|$Test$|$ & Test sets & $|$Lang.$|$ & Task & Metric & Domain \\ \midrule
% \multicolumn{8}{c}{Single sentence classification} \\ \midrule
% \multirow{2}{*}{\rotatebox[origin=b]{90}{\parbox{1.0}{Single seq. clas.}}} & \multirow{2}{*}{CliC} & \multirow{2}{*}{30,521} & \multirow{2}{*}{4,181} & \multirow{2}{*}{1,692--8,621} & \multirow{2}{*}{3} & \multirow{2}{*}{intent} & \multirow{2}{*}{ac.} & \multirow{2}{*}{assistant} \\
% \\ \midrule
% & MLDoc & &  &  & 8 & topic & ac. & news \\ \midrule
% \multicolumn{8}{c}{Sentence pair classification} \\ \midrule
% \multirow{2}{*}{\rotatebox[origin=b]{90}{\parbox{1.0}{Sent. pair clas.}}} 
\multirow{2}{*}{Classification} & XNLI & 392,702 & 2,490 & 5,010 & translations & 15 & NLI & Accuracy & Misc.  \\
& XCOPA & 33,410+400 & 100 & 500 & translations & 11 & Reasoning & Accuracy & Misc. \\ \midrule
% & PAWS-X & 49,401 & 2,000 & 2,000 & translations & 7 & Paraphrase & Acc. & Wiki / Quora \\ \midrule
% \multicolumn{8}{c}{Structured prediction} \\ \midrule
\multirow{2}{*}{Struct. prediction} & UD-POS & 21,253 & 3,974 & 47-20,436 & ind. annot. & 37 (104) & POS & F1 & Misc. \\
% & NER-CoNLL &  &  &  & 4 & NER \\ 
& WikiANN-NER & 20,000 & 10,000 & 1,000-10,000 & ind. annot. & 47 (176) & NER & F1 & Wikipedia \\
\midrule
% \multicolumn{8}{c}{Question answering} \\ \midrule
\multirow{3}{*}{QA} & XQuAD & \multirow{2}{*}{87,599} & \multirow{2}{*}{10,570} & 1,190 & translations & 11 & Span extraction & F1 / EM & Wikipedia \\
& MLQA &  &  & 4,517--11,590 & translations & 7 & Span extraction & F1 / EM & Wikipedia \\ 
& TyDiQA-GoldP & 3,696 & 634 & 323--2,719 & ind. annot. & 9 & Span extraction & F1 / EM & Wikipedia \\ \midrule
% \multicolumn{8}{c}{Sentence retrieval} \\ \midrule
\multirow{3}{*}{Retrieval} & Tatoeba & 87,599 & 10,570 & 1,000 & translations & 38 (122) & Sentence retrieval & Accuracy & Misc. \\
& Mewsli-X & 116,903 & 10,252 & 428--1,482 & ind. annot. & 11 (50) & Lang. agn. retrieval & mAP@20 & News \\
& LAReQA XQuAD-R & 87,599 & 10,570 & 1,190 & translations & 11 & Lang. agn. retrieval & mAP@20 & Wikipedia \\
% BUCC & - & - & 1,896--14,330 & - & 5 & Sent. retrieval & F1 & Wiki / news \\
% & UN &  &  &  & 6 & & MRR \\
\bottomrule
\end{tabular}%
}
\caption{The tasks in \name. For tasks that have training and dev sets in other languages, we only report the English numbers. We report the number of test examples per target language, the nature of the test sets (whether they are translations of English data or independently annotated), and the size of the intersection with our selected languages (the total number of languages is in brackets). For WikiANN-NER and UD-POS, sizes are in sentences. XCOPA includes training data of SIQa \citep{Sap2019socialiqa} and Tatoeba uses SQuAD v1.1 data for fine-tuning.}
\label{tab:tasks}
\end{table*}

Breaking down performance by language family, on Tatoeba (Figure \ref{fig:xtreme_tatoeba_performance}) recent models still struggle with a few low-resource languages.
% , i.e.~Swahili, Javanese, and Tagalog. 
Models perform well for most other languages and their scores are concentrated in a relatively small range. On MLQA (Figure \ref{fig:xtreme_mlqa_performance}), scores have increased slightly but remain well below performance on English. On POS tagging (Figure \ref{fig:xtreme_pos_performance}), scores remain largely the same; performance is lower for some languages with non-Latin scripts 
% (Japanese, Chinese, Korean, Thai) 
and low-resource languages. 
% (Yoruba).
% Similar trends hold for named entity recognition. 
We show the scores for the remaining tasks in Appendix \ref{app:xtreme_scores_language_family}. The remaining gap to English performance on these tasks is partially an artefact of the evaluation setup: zero-shot cross-lingual transfer from English favors English representations whereas models fine-tuned on in-language monolingual data perform more similarly across languages \cite{Clark2020tydiqa,Hu2020xtreme}. 
% \lr{I don't find this last sentence very clear and I think this will be difficult for readers who are less familiar with XTREME. What is it about the evaluation setup that favours English? Are all evaluation tasks in XTREME set up as zero-shot transfer from English? Maybe some comments about the XTREME setup earlier in the paper will frame this better. What does in-language performance mean? Is this saying that certain language families are more difficult across tasks?}
% Nevertheless, models that have learned strongly aligned multilingual representations should be able to demonstrate similar performance in different languages regardless of the transfer setting. 

Overall, representations from \emph{token-level} MLM pre-training are of limited use for cross-lingual \emph{sentence} retrieval, as evidenced by the comparatively poor performance of the mBERT and XLM-R models. Fine-tuning on sentence-level tasks \cite{phang2020english,Fang2020filter} can mitigate this.
The strong performance of recent models such as VECO and ERNIE-M on the retrieval tasks can be attributed to a combination of parallel data and new pre-training objectives that make use of it. Pre-training on parallel data improves performance on retrieval by making the pre-training task more similar to the downstream setting but does not significantly improve performance on other tasks.
\emph{Fine-tuning} on automatically translated task-specific data yields strong gains and is used by most recent models to achieve the best performance \cite{Hu2020xtreme,Ouyang2020erniem,luo2020veco}. Nevertheless, key challenges such as how to learn robust cross-lingual syntactic and semantic processing capabilities during \emph{pre-training} remain.

% , does not entail that more recent models are vastly more powerful multilingual encoders, but highlights that they effectively leverage additional data sources.\dhg{This sentence is a little vague. "Does not entail" does not entail "are not", but the second half of the sentence seems to imply that the gains are due to better use of data instead. Is it the case that gains are not due to better encoder models? Or is the answer that "it depends". This should be made more clear.} 
% Intermediate tasks that require high-level inference and reasoning abilities generally work best \cite{phang2020english}. 
% \gn{I'm not sure if the QA tasks here really require reasoning? Maybe ``syntactic or semantic processing capabilities'' would be sufficient?}.
% In light of the number of different objectives proposed it is not clear, however, what is the best way to make use of parallel data during pre-training.

% \lr{This paragraph is also a bit impenetrable especially without explaining the retrieval setup. Could you state more plainly whether pre-training with parallel data is a good solution to the retrieval problem? Is it only available for high resource languages for example? There are several contrasts presented here in quick succession (note `but', `in contrast', `however', `finally') so that in the end I don't come away with an understanding of the narrative on developments in the retrieval task, or whether you see them as positive for multilingual NLU or superficial.}

\section{XTREME-R}

% \gn{It felt a little weird that the previous section is ``Revisiting XTREME'' and then this one is ``XTREME-R'' which also stands for ``XTREME Revisited''. Maybe the previous section could be something more like ``Examining the State of Multilingual Benchmarking'' or ``Examining the State of XTREME''?}

In order to encourage the NLP community to tackle challenging research directions in pursuit of better cross-lingual model generalization, we propose \name (\textsc{xtreme} \textsc{r}evisited).
% , an improved version of the \xtreme benchmark. 
\name shares its predecessor's core design principles for creating an accessible benchmark to evaluate cross-lingual transfer
% ---that tasks should have an appropriate degree of difficulty to incentivize progress, be diverse, be efficient to train, 
% be inherently multilingual, and have a permissive license---
but makes some key changes.
% refocuses on the tasks that have proven to be hardest for current multilingual models.
% \name takes some of \xtreme's principles more seriously. 
% We benchmark the runtimes of two standard models on the tasks in \name to show which tasks are most efficient to train. \name also improves on language diversity.

First, \name focuses on the tasks that have proven to be hardest for current multilingual models. To this end, it drops \xtreme's \mbox{PAWS-X} and BUCC tasks since recent advances have left less room for further improvement, and they cover only a small number of less diverse languages. They are replaced instead by three new, more challenging tasks: one focusing on causal commonsense reasoning (\S\ref{sec:multilingual-causal-reasoning}) and two focusing on harder retrieval scenarios (\S\ref{sec:retrieval-from-multilingual-pool}), as this has been the task category where gains have been easiest to achieve.
We retain \xtreme's seven other tasks as each still presents substantial challenges for state-of-the-art cross-lingual models (\S\ref{sec:retained}). Overall, \name includes 10 diverse tasks, summarized in Table \ref{tab:tasks}.

We also make changes to the structured prediction tasks, NER and POS. Instead of only providing examples as lists of tokens, \name always provides the full text of an input sentence, thus ensuring that the entire benchmark now supports research on models that operate directly from the raw input string \cite{Clark2021canine}. Furthermore, \name adopts a more realistic version of the NER task in which no gold tokenization is provided at all, meaning that systems will either have to use model-predicted tokens or embrace tokenization-free approaches.
Finally, \name provides a multilingual diagnostic and evaluation suite (\textsection \ref{sec:diagnostic_suite}).

% \vspace{-0.5mm}
\subsection{Retained Tasks}
\label{sec:retained}

% \dhg{This is a comment that applies to the whole paper, but is particular apparent in this section. The paper seems to use the word ``task'' in a somewhat imprecise way. Sometimes it is used for a general problem type (as in traditional NLP ``tasks'' such as POS, NER, QA, etc), but sometimes it's used to mean ``one of the 10 things that make up the benchmark''. For example, XQuAD, MLQA, and TyDiQA-GoldP are all instantiations of the same basic task (cross-lingual QA), differing only in the source of the data, but POS and NER are both general tasks, whose instantiations happen to be based on UD UPOS data and WikiANN data, repectively. I think, at a minimum, it would be good to make sure that the conceptual granularity is consistent when you talk about these things, so in places where the paper talks about, e.g., XQuAD, MLQA, and TyDiQA-GoldP, I've changed POS and NER to UD-POS and WikiANN-NER.}

We retain the XNLI \citep{Conneau2018xnli}, UD-POS \citep{nivre2018universal}, WikiANN-NER \citep{Pan2017}, XQuAD \citep{artetxe2020cross}, MLQA \citep{Lewis2019mlqa}, TyDiQA-GoldP \citep{Clark2020tydiqa}, and Tatoeba \citep{Artetxe2019massively} tasks from \xtreme (see Appendix \ref{app:retained-tasks}).

\subsection{New Tasks}

% \gn{Might be best to have a sub-section here such as ``Multilingual Causal Reasoning'', just so we don't have the asymmetry with ``Retrieval from a Multilingual Pool'' having its own section and this task not.}
\subsubsection{Multilingual Causal Reasoning}
\label{sec:multilingual-causal-reasoning}

\noindent \textbf{XCOPA} $\:$ The Cross-lingual Choice of Plausible Alternatives \citep{Ponti2020xcopa} dataset asks models to decide which of two sentences causally follows a premise sentence. The XCOPA authors translated and re-annotated the validation and test sets of the English COPA \citep{roemmele2011choice} dataset into 11 languages, which we use for evaluation. The English COPA training set together with the Social IQa \citep{Sap2019socialiqa} training data are used for training. While accuracy on the English COPA recently reached 94.8\% \citep{Raffel2020t5}, the state-of-the-art on XCOPA is only around 70\%.
% (with human-level performance on average at 97.6\%), 
% leaving ample room for improvement.

\subsubsection{Retrieval from a Multilingual Pool}
\label{sec:retrieval-from-multilingual-pool}

% \gn{This would benefit from being explained after we've already explained the tasks in the existing benchmark.}\dhg{I agree: I would flip the order of ``Retained Tasks'' and ``New Tasks''.}
Many previous retrieval benchmarks assume that the entire candidate pool is in a single language. For instance, a French query will be used to search over only English candidates. 
% This setting is easy in the sense that a model is not penalized for assigning a high rank to a less relevant candidate in the same or a similar language to the query.
% a high rank to irrelevant candidates in other languages.\dhg{I would assume that the main risk for a model would be ranking a candidate written in the same language as the query higher than a more relevant candidate written in a different language, i.e. that candidates tend to look ``more relevant'' just because they're in the query's language. As such, I would change this sentence to something like: ``...a model is not penalized for assigning a higher rank to a less relevant candidate just because it happens to be written in the same language as the query''.} 
However, practical settings often violate this assumption, e.g.~the answer to a question may be available in any number of languages, possibly different from the query language. Models that cannot compare the appropriateness of retrieval results across languages are thus ineffective in such real-world scenarios.

% To shed new light on the 
\name includes two new related cross-lingual retrieval tasks.
The first seeks to measure the extent to which cross-lingual representations are ``strongly aligned'' \citep{roy-etal-2020-lareqa}, i.e.~they place the semantically most related text pairs (e.g.~a question and its answer) closest together in representation space, \emph{regardless} of their language identities.
The second analogously frames entity linking as retrieving from a multilingual pool of entity descriptions, given an entity mention in context \citep{botha-etal-2020-entity}.
For both, we report performance as mean average precision at 20 (mAP@20).

%%% Previous shorter version:
% We incorporate both tasks into \name, and report performance as mean average precision at 20 (mAP@20) for a dual encoder architecture trained with in-batch sampled softmax loss \citep{gillick2018end}.

\begin{table*}[]
\centering
\includegraphics[width=\textwidth]{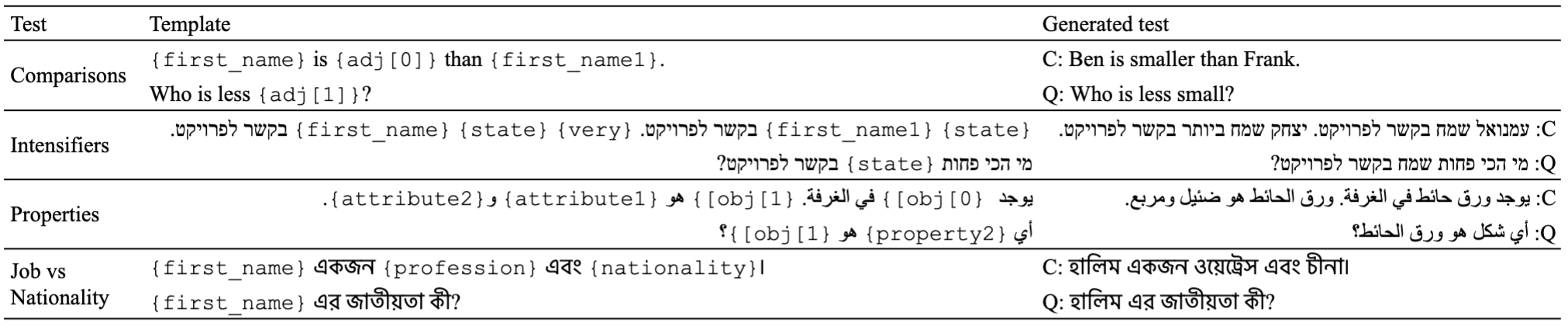}
\caption{\checklist templates and generated tests for different capabilities in English, Hebrew, Arabic, and Bengali. Words in curly brackets \texttt{$\{\ldots\}$} are placeholders; see \citet{Ribeiro2020checklist} for more information.}
\label{tab:checklist_tests}
\end{table*}

\noindent \textbf{LAReQA} $\:$
%%% Noah's original isolated LAReQA description %%
%%Language Agnostic Retrieval Question Answering \citep{roy-etal-2020-lareqa} is a sentence retrieval task that requires identifying answers to a posed question from within a large multilingual candidate pool. Unlike previous cross-lingual retrieval benchmarks, each query has target answers in multiple languages, and models are expected to rank all correct answers above all incorrect answers, regardless of language. This problem is challenging for cross-lingual representation learning in that it requires ``strongly aligned'' representations that are agnostic to the text language. We use the LAReQA XQuAD-R dataset which contains 13,090 questions each of which has 11 target answers (in 11 distinct languages) within the pool of 13,014 candidate answer sentences. We evaluate in the zero-shot setting. Following \citet{roy-etal-2020-lareqa}, we fine-tune models on the SQuAD v1.1 train set using a dual encoder architecture with an in-batch sampled softmax loss. We select checkpoints based on the fine-tuning loss on the SQuAD v1.1 dev set. The fine-tuned model is used to rank all 170M question-answer pairs in XQuAD-R, and we report mean average precision (mAP).
%%%
Language Agnostic Retrieval Question Answering \citep{roy-etal-2020-lareqa} is a sentence retrieval task. Each query has target answers in multiple languages, and models are expected to rank all correct answers above all incorrect answers, regardless of language. We use the LAReQA XQuAD-R dataset which contains 13,090 questions each of which has 11 target answers (in 11 distinct languages) within the pool of 13,014 candidate answer sentences. 
Following \citet{roy-etal-2020-lareqa}, we fine-tune models on the SQuAD v1.1 train set.
%using a dual encoder architecture with an in-batch sampled softmax loss.
% We select checkpoints based on the fine-tuning loss on the SQuAD v1.1 dev set. 
The fine-tuned model is used to rank the 13K candidates for each question.
%, and we report mean average precision at 20 (mAP@20).
% Time is 14m on GCP instance (Intel Skylake) using one V100 GPU

\noindent \textbf{Mewsli-X} $\:$ Mewsli (\textbf{M}ultilingual Entities in N\textbf{ews}, \textbf{li}nked) is an automatically extracted dataset that requires linking a contextual entity mention to its entry in a language-agnostic knowledge base by retrieving the entity's description from a multilingual candidate pool \citep{botha-etal-2020-entity}.
For \name, we derive Mewsli-X as a new variant of \mbox{Mewsli-9}, still linking against WikiData \citep{vrandevcic2014wikidata}.
Mewsli-X features 15K mentions in 11 languages: given a mention in context, the task is to retrieve the single correct target entity description from a candidate pool ranging over 1M candidates across all 50 languages of \name.
Fine-tuning is done on a predefined set of English-only mention-entity pairs randomly sampled from English Wikipedia hyperlinks (see Appendix \ref{app:mewsli-x} for further details).

For our baseline systems on both tasks, we follow previous work \cite{roy-etal-2020-lareqa, botha-etal-2020-entity} and train a dual encoder initialized from the pre-trained model weights, optimizing for an in-batch sampled softmax loss \citep{gillick2018end}.

%Compared to existing entity linking benchmarks \citep{petroni2020kilt,roder2018gerbil}, Mewsli-X is better suited to \name for reasons of accessibility, multilingualism and entity diversity.

\subsection{Languages}

\name adds the following ten languages to \xtreme: Haitian Creole, Cusco Quechuan,
% (members of two new language families), 
Wolof,
% (spoken in West Africa),
Lithuanian,
% (Baltic branch of the Indo-European language family),
Punjabi, Gujarati,
% (two major world languages widely spoken in India)
Polish, Ukrainian, Azerbaijani, and Romanian.
% all languages with large speaker populations.
In total, \name covers the following 50 languages (shown using their ISO 639-1 codes for brevity; new languages are bolded) belonging to 14 language families and two isolates: af, ar, \textbf{az}, bg, bn, de, el, en, es, et, eu, fa, fi, fr, \textbf{gu}, he, hi, \textbf{ht}, hu, id, it, ja, jv, ka, kk, ko, \textbf{lt}, ml, mr, ms, my, nl, \textbf{pa}, \textbf{pl}, pt, \textbf{qu}, \textbf{ro}, ru, sw, ta, te, th, tl, tr, \textbf{uk}, ur, vi, \textbf{wo}, yo, zh.\footnote{The new languages are covered in both the new tasks as well as in UD-POS, WikiANN-NER, and Tatoeba.}
\name is similarly typologically and genealogically diverse as \xtreme while covering a larger number of languages (see Appendix \ref{app:languages}).

% A downside of \name and similar benchmarks is that even though it covers a number of low-resource languages, many of them occur only in one or two tasks. It is thus difficult to obtain a more fine-grained impression of a model's generalization performance. In addition, judging a model's performance based on a single aggregate number obfuscates many nuances of model behaviour.

\subsection{Diagnostic and evaluation suite} \label{sec:diagnostic_suite}

To increase the language coverage of low-resource languages in \name and to enable us to systematically evaluate a model's cross-lingual generalization ability, we augment \name with a massively multilingual diagnostic and evaluation suite. Challenge sets and diagnostic suites in NLP \citep{Wang2019superglue,Wang2019glue,Belinkov2019analysis} are mostly limited to English, with a few exceptions \citep{Gulordava2018colorless}. As challenge sets are generally created with a human in the loop, the main challenge for creating a large multilingual diagnostic suite is to scale the annotation or translation effort to many languages and to deal with each language's idiosyncrasies. 

% Challenge sets and diagnostic suites have become more popular in NLP recently \citep{Wang2019superglue,Wang2019glue,Belinkov2019analysis}. Except for a few exceptions \citep{Gulordava2018colorless} all challenge sets that we are aware of are in English. As challenge sets are generally created with a human in the loop, the main challenge for creating a large multilingual diagnostic suite is to scale the annotation or translation effort to many languages and to deal with each language's idiosyncrasies.

\noindent \textbf{\multichecklist} $\:$ To address this, we build on the \checklist \citep{Ribeiro2020checklist} framework, which facilitates creating parameterized tests for models. \checklist enables the creation of test cases using templates, which test for specific behavioral capabilities of a model with regard to a downstream task. Importantly, by relying on template-based tests, we can efficiently generate a large number of diverse multilingual test cases by creating a relatively small number of templates in 50 languages.\footnote{In contrast, translating an existing test suite or dataset or annotating individual examples for 50 languages would be prohibitively expensive} We focus on translating English tests, which consist of templates and their fill-in values.\footnote{Templates could alternatively be created by native speakers in each language.} To study the feasibility of creating multilingual test cases at scale, we translate the minimum functionality tests (MFT) of \checklist, which probe for general vocabulary and taxonomic knowledge in question answering. We instruct translators to create separate variants of a template to disambiguate linguistic phenomena, such as gender of fill-in values, question type, semantics of properties, etc. We automatically fill names in each language based on data from Wikidata and programmatically consolidate different templates in each language.
% We restrict selection to male names to remove one source of gender variation in templates. 
We show examples of templates and the tests that they generate in different languages in Table \ref{tab:checklist_tests}.

We highlight statistics of the dataset and translation process, instructions to translators, and general challenges of template translation in Appendix \ref{app:multichecklist}. We believe that parameterized tests are a powerful tool to obtain diverse diagnostics data for otherwise resource-starved languages. We view participatory research \cite{nekoto_etal_2020_participatory} with native speakers to create template-based test cases testing for language-specific behaviour as particularly promising. 
% We make our test cases in 50 languages available to facilitate future extension.

% We tried to address most of these challenges by instructing translators to create additional templates to disambiguate linguistic phenomena and by consolidating different templates programmatically.

% While the translation of template-based tests enables expansion to many languages, additional effort was often necessary to unify language-specific variants of a template. In addition, translations of English-centric templates are sometimes stilted. Finally, by only translating tests we are not able to directly probe idiosyncrasies of a target language. We thus recommend future effort to focus on working with native speakers to create template-based test cases in their own languages. We make our test cases in 50 languages available to facilitate future extension and provide a broader overview of the translation process in the Appendix.

\begin{table*}[]
\centering
\resizebox{\textwidth}{!}{%
\begin{tabular}{l c cc cc ccc cc c}
\toprule
\multirow{2}{*}{Model} & \multirow{2}{*}{Avg} & \multicolumn{2}{c}{Classification} & \multicolumn{2}{c}{Structured prediction} & \multicolumn{3}{c}{Question answering} & \multicolumn{2}{c}{Lang.-agnostic retrieval} & \multicolumn{1}{c}{Retrieval} \\
 & & XNLI & XCOPA & UD-POS & WikiANN-NER & XQuAD & MLQA & TyDiQA-GoldP & Mewsli-X & LAReQA & Tatoeba \\
\midrule
Metrics & & Acc. & Acc. & F1 & F1 & F1 / EM & F1 / EM & F1 / EM & mAP@20 & mAP@20 & Acc. \\ \midrule
\multicolumn{11}{l}{\emph{Cross-lingual zero-shot transfer (models are trained on English data)}} \\ \midrule
mBERT & 54.1 & 66.5 & 56.1 & 70.9 & 62.7 & 65.1 / 50.4 & 61.3 / 44.1 & 58.4 / 43.7 & 38.6 & 21.6 & 43.3\\
XLM-R Large & 65.3 & 79.2 & 69.2 & 75.0 & 64.4 & 77.2 / 61.6 & 72.7 / 54.2 & 64.3 / 45.8 & 45.7 & 40.7 & 77.3\\
mT5-XXL & 64.6 & 84.8 & 74.9 & 71.5 & 68.8 & 81.5 / 66.6 & 75.6 / 57.3 & 80.1 / 65.7 & 41.7* & 24.7* & 45.7 \\ \midrule
\multicolumn{11}{l}{\emph{Translate-train (models are trained on English training data translated to the target language)}} \\ \midrule
mBERT & - & 74.6 & 57.3  & - & - & 70.0 / 56.0 & 65.6 / 48.0 & 52.4 / 39.5 & - & - & - \\
mBERT, multilingual & - & 75.1 & 57.9 & - & - & 72.4 / 58.3 & 67.6 / 49.8 & 59.5 / 45.8 & - & - & - \\
mT5-XXL, multilingual & - & 87.5 & 77.8 & - & - & 84.7 / 71.8 & 76.4 / 58.4 & 81.9 / 68.6 & - & - & - \\
% \\\midrule
% \multicolumn{11}{l}{\emph{Translate-test (models are trained on English data and evaluated on target language data translated to English)}} \\ \midrule
% BERT-large & & 76.8 & & & & 76.3 / 62.1 & 72.9 / 55.3 & 67.8 / 51.2 & - & - & - \\
% \midrule
% \multicolumn{11}{l}{\emph{In-language models (models are trained on the target language training data)}} \\ \midrule
% XLM-R, 1000 examples & \\
% XLM-R & \\
% XLM-R, multi-task & \\
\midrule
Human & - & 92.8 & 97.6 & 97.0 & - & 91.2 / 82.3 & 91.2 / 82.3 & 90.1 / - & - & - & -\\
\bottomrule
\end{tabular}%
}
\caption{Overall results of baselines across all \name tasks. *: Due to compute limitations, mT5-XXL language-agnostic retrieval results are obtained with a frozen rather than a fine-tuned model.
% *: Proximity to estimated human performance does not entail that a model actually exhibits human-level cross-lingual generalization ability.
}
\label{tab:main-results}
\end{table*}

\noindent \textbf{Multilingual \explainaboard} $\:$ The standard practice in leaderboards is to average performance across different settings \cite{Wang2019glue,Wang2019superglue}. While this provides discriminative power,
% and yields a clear ranking of systems
it has limited utility for examining the relative advantages of systems, the characteristics of different datasets and languages, and how these factors relate to each other. To provide more granular evaluation capabilities, we extend \citet{Fu2020explainaboard,liu2021explainaboard}'s \explainaboard to the task categories and languages in \name.
% introduce a fine-grained multilingual evaluation framework. 
\explainaboard provides a more nuanced impression of a model's performance on a task by defining task-specific attributes (e.g.~\emph{entity length} for NER). The test set is partitioned into different buckets based on the defined attributes and performance is broken down over different attribute values. We define new task-specific attributes for the four task types as well as task-independent attributes (see Appendix \ref{app:nuanced-evaluation}). 

\noindent \textbf{Metadata} $\:$ 
% Beyond breaking down performance by attributes, 
We additionally would like to enable practitioners to rank submissions based on other information. To this end, we ask each submission to \name for relevant metadata such as the number of parameters, the amount of pre-training data, etc. We will show this information in an interactive leaderboard (see Appendix \ref{app:metadata} for the metadata of current \xtreme submissions).

\section{Experiments}

\paragraph{Training and evaluation setup} \name focuses on \emph{zero-shot cross-lingual transfer} from English. While recent work \cite{Hu2020xtreme,Lauscher2020fromzerotohero,Hedderich2020transfer} demonstrates the benefits of fine-tuning on in-language data, we believe the zero-shot scenario remains the most effective way to evaluate the amount of \emph{a priori} multilingual knowledge a pre-trained model captures. Due to variation in cross-lingual evaluation \cite{keung-etal-2020-dont}, we recommend researchers to use the validation set of a \emph{single} target language for development \cite{artetxe-etal-2020-call}.
% and only use the source language validation set if no target language validation set is available.

\subsection{Baselines}

We employ established pre-trained multilingual and models using translations as baselines.

\noindent \textbf{mBERT} $\:$ Multilingual BERT \cite{Devlin2019bert} has been pretrained on the Wikipedias of 104 languages using MLM.

\noindent \textbf{XLM-R} $\:$ XLM-R Large \cite{Conneau2020xlmr} uses the same MLM objective with a larger model, and was trained on a magnitude more web data from 100 languages.

\noindent \textbf{mT5} $\:$ Multilingual T5 \cite{Xue2020mt5} is an encoder-decoder transformer that frames NLP tasks in a ``text-to-text'' format. It was pre-trained with MLM on a large multilingual web corpus covering 101 languages. We employ the largest mT5-XXL variant with 13B parameters.
% \footnote{We do not provide results for retrieval as it is unclear how to use encoder-decoder models most effectively.}
% in order to assess the impact of scale.

\noindent \textbf{Translate-train} $\:$ To evaluate the impact of MT, we fine-tune mBERT on translations of English training data from \citet{Hu2020xtreme}. We create new translations for the XCOPA and SIQa data using an in-house MT system.\footnote{We are unable to produce translations for Quechua.} 

\noindent \textbf{Translate-train multilingual} $\:$ In addition, we fine-tune both mBERT and mT5 on the combined translated training data of all languages (including the original English data) jointly.

% \noindent \textbf{Translate-test} $\:$ Alternatively, we train the English BERT-Large \cite{Devlin2019bert} model on the English training data and evaluate it on test data that was translated from the target language to English. Analogous to above, we use translations from  \citet{Hu2020xtreme} and create new translations for XCOPA and the structured prediction tasks.
 
\noindent \textbf{Human performance} $\:$ We use the human performance estimates from \xtreme for the retained tasks. For XCOPA we average the proportion of annotated labels disagreeing with the majority label across all languages \cite{Ponti2020xcopa}. We are not able to obtain human performance estimates for the new retrieval tasks as identifying a translation among a large number of candidates is too time-consuming for a human to perform.

\subsection{Results} 

\begin{table}[t!]
\centering
\resizebox{\columnwidth}{!}{%
\begin{tabular}{ccccccc|c}
\toprule
\multirow{2}{*}{Lang.} & \multirow{2}{*}{Comparisons} & \multirow{2}{*}{Intensifiers} & \multirow{2}{*}{Properties} & Job vs & Animal vs & Animal vs & \multirow{2}{*}{Avg} \\
& & & & Nationality & Vehicles & Vehicles 2\\ \midrule

fi & \cellcolor[HTML]{7AC9A3}10.7 & \cellcolor[HTML]{F1B1AC}79.8 & \cellcolor[HTML]{C9E9D9}34.1 & \cellcolor[HTML]{64C093}4.0 & \cellcolor[HTML]{76C79F}9.2 & \cellcolor[HTML]{5CBD8D}1.5 & \cellcolor[HTML]{A5DAC0}23.2 \\
fr & \cellcolor[HTML]{64C093}4.0 & \cellcolor[HTML]{E78279}98.0 & \cellcolor[HTML]{8ACFAE}15.4 & \cellcolor[HTML]{59BC8B}0.8 & \cellcolor[HTML]{78C8A1}10.0 & \cellcolor[HTML]{84CDA9}13.5 & \cellcolor[HTML]{A6DBC1}23.6 \\
en & \cellcolor[HTML]{6EC49A}7.0 & \cellcolor[HTML]{EB968E}90.4 & \cellcolor[HTML]{D8EFE4}38.5 & \cellcolor[HTML]{57BB8A}0.0 & \cellcolor[HTML]{6BC398}6.0 & \cellcolor[HTML]{58BB8B}0.5 & \cellcolor[HTML]{A6DBC1}23.7 \\
pl & \cellcolor[HTML]{9DD7BA}20.9 & \cellcolor[HTML]{E67C73}100.0 & \cellcolor[HTML]{9CD7BA}20.8 & \cellcolor[HTML]{57BB8A}0.0 & \cellcolor[HTML]{7BC9A3}11.0 & \cellcolor[HTML]{62BF92}3.5 & \cellcolor[HTML]{AEDEC6}26.0 \\
bg & \cellcolor[HTML]{FDFEFD}49.5 & \cellcolor[HTML]{F5C8C5}71.1 & \cellcolor[HTML]{86CEAB}14.3 & \cellcolor[HTML]{57BB8A}0.0 & \cellcolor[HTML]{8ED1B0}16.5 & \cellcolor[HTML]{6BC398}6.1 & \cellcolor[HTML]{AFDEC7}26.2 \\
ka & \cellcolor[HTML]{57BB8A}0.0 & \cellcolor[HTML]{E98A82}95.0 & \cellcolor[HTML]{C9E9D9}34.0 & \cellcolor[HTML]{7FCBA6}12.0 & \cellcolor[HTML]{82CCA8}13.0 & \cellcolor[HTML]{7FCBA6}12.0 & \cellcolor[HTML]{B3E0CA}27.7 \\
nl & \cellcolor[HTML]{86CEAA}14.1 & \cellcolor[HTML]{EB928A}91.9 & \cellcolor[HTML]{ACDDC5}25.4 & \cellcolor[HTML]{5ABC8C}1.0 & \cellcolor[HTML]{A7DBC2}24.1 & \cellcolor[HTML]{7DCAA4}11.3 & \cellcolor[HTML]{B4E1CB}28.0 \\
de & \cellcolor[HTML]{90D2B1}17.1 & \cellcolor[HTML]{EB958E}90.6 & \cellcolor[HTML]{D7EEE3}38.2 & \cellcolor[HTML]{5FBE8F}2.5 & \cellcolor[HTML]{90D2B1}17.0 & \cellcolor[HTML]{75C79F}9.1 & \cellcolor[HTML]{B8E2CE}29.1 \\
it & \cellcolor[HTML]{62BF92}3.5 & \cellcolor[HTML]{E78078}98.5 & \cellcolor[HTML]{FFFFFF}50.0 & \cellcolor[HTML]{6DC499}6.7 & \cellcolor[HTML]{7AC9A2}10.5 & \cellcolor[HTML]{69C296}5.6 & \cellcolor[HTML]{B8E2CE}29.1 \\
hu & \cellcolor[HTML]{6BC398}6.1 & \cellcolor[HTML]{E78078}98.5 & \cellcolor[HTML]{C7E8D8}33.5 & \cellcolor[HTML]{86CEAA}14.0 & \cellcolor[HTML]{B5E1CB}28.0 & \cellcolor[HTML]{7AC9A2}10.6 & \cellcolor[HTML]{C1E6D4}31.8 \\
hi & \cellcolor[HTML]{A1D9BD}22.2 & \cellcolor[HTML]{F9DBD9}63.8 & \cellcolor[HTML]{F8DAD8}64.3 & \cellcolor[HTML]{71C59C}8.0 & \cellcolor[HTML]{B5E1CB}28.0 & \cellcolor[HTML]{73C69E}8.6 & \cellcolor[HTML]{C4E7D5}32.5 \\
ru & \cellcolor[HTML]{C2E6D4}32.0 & \cellcolor[HTML]{E98A82}95.0 & \cellcolor[HTML]{C8E9D9}33.8 & \cellcolor[HTML]{6EC49A}7.1 & \cellcolor[HTML]{A7DBC2}24.0 & \cellcolor[HTML]{61BF91}3.0 & \cellcolor[HTML]{C4E7D6}32.5 \\
fa & \cellcolor[HTML]{7AC9A2}10.6 & \cellcolor[HTML]{EEA6A0}84.2 & \cellcolor[HTML]{FFFBFB}51.7 & \cellcolor[HTML]{67C195}5.0 & \cellcolor[HTML]{D8EFE4}38.5 & \cellcolor[HTML]{7DCAA5}11.6 & \cellcolor[HTML]{C7E8D8}33.6 \\
et & \cellcolor[HTML]{7CCAA3}11.1 & \cellcolor[HTML]{EB938C}91.4 & \cellcolor[HTML]{D0ECDE}36.1 & \cellcolor[HTML]{87CEAB}14.5 & \cellcolor[HTML]{FBFDFC}49.0 & \cellcolor[HTML]{57BB8A}0.0 & \cellcolor[HTML]{C8E8D8}33.7 \\
es & \cellcolor[HTML]{77C8A0}9.6 & \cellcolor[HTML]{E77E75}99.5 & \cellcolor[HTML]{FAE0DE}62.0 & \cellcolor[HTML]{57BB8A}0.0 & \cellcolor[HTML]{C4E7D6}32.5 & \cellcolor[HTML]{69C296}5.5 & \cellcolor[HTML]{CCEADB}34.8 \\
ms & \cellcolor[HTML]{64C093}4.0 & \cellcolor[HTML]{E88279}97.9 & \cellcolor[HTML]{EEA6A0}84.0 & \cellcolor[HTML]{69C296}5.5 & \cellcolor[HTML]{95D4B5}18.5 & \cellcolor[HTML]{58BB8B}0.5 & \cellcolor[HTML]{CCEADC}35.1 \\
ml & \cellcolor[HTML]{6EC49A}7.1 & \cellcolor[HTML]{F4C1BD}73.8 & \cellcolor[HTML]{F3C1BC}74.0 & \cellcolor[HTML]{6CC399}6.5 & \cellcolor[HTML]{C4E7D6}32.5 & \cellcolor[HTML]{96D4B6}19.0 & \cellcolor[HTML]{CEEBDD}35.5 \\
lt & \cellcolor[HTML]{81CCA7}12.7 & \cellcolor[HTML]{EEA6A0}84.2 & \cellcolor[HTML]{F7D1CE}67.6 & \cellcolor[HTML]{ABDDC4}25.2 & \cellcolor[HTML]{A2D9BE}22.5 & \cellcolor[HTML]{5FBE8F}2.5 & \cellcolor[HTML]{CFEBDD}35.8 \\
el & \cellcolor[HTML]{6BC398}6.1 & \cellcolor[HTML]{E88279}98.0 & \cellcolor[HTML]{D7EFE3}38.3 & \cellcolor[HTML]{93D3B4}18.1 & \cellcolor[HTML]{EEF8F3}45.0 & \cellcolor[HTML]{8ED1B0}16.5 & \cellcolor[HTML]{D3EDE0}37.0 \\
af & \cellcolor[HTML]{F4FAF7}47.0 & \cellcolor[HTML]{F1B4AF}78.7 & \cellcolor[HTML]{CEEBDD}35.5 & \cellcolor[HTML]{5ABC8C}1.0 & \cellcolor[HTML]{FCF0EF}56.0 & \cellcolor[HTML]{9ED7BB}21.2 & \cellcolor[HTML]{DDF1E7}39.9 \\
ta & \cellcolor[HTML]{FBE8E6}59.0 & \cellcolor[HTML]{F1B4AF}78.7 & \cellcolor[HTML]{F8D7D4}65.5 & \cellcolor[HTML]{7DCAA4}11.5 & \cellcolor[HTML]{84CDA9}13.5 & \cellcolor[HTML]{83CCA8}13.2 & \cellcolor[HTML]{DEF1E8}40.2 \\
uk & \cellcolor[HTML]{8DD1AF}16.2 & \cellcolor[HTML]{E98A82}94.9 & \cellcolor[HTML]{DCF1E7}39.8 & \cellcolor[HTML]{B1DFC8}26.9 & \cellcolor[HTML]{FFFCFB}51.5 & \cellcolor[HTML]{86CEAB}14.1 & \cellcolor[HTML]{DFF2E8}40.6 \\
pt & \cellcolor[HTML]{FFFBFB}51.8 & \cellcolor[HTML]{E77F76}99.0 & \cellcolor[HTML]{F9DFDD}62.4 & \cellcolor[HTML]{59BC8B}0.8 & \cellcolor[HTML]{A4DABF}23.0 & \cellcolor[HTML]{6EC49A}7.1 & \cellcolor[HTML]{DFF2E9}40.7 \\
tl & \cellcolor[HTML]{57BB8A}0.0 & \cellcolor[HTML]{E67C73}100.0 & \cellcolor[HTML]{F7D4D1}66.5 & \cellcolor[HTML]{81CCA7}12.5 & \cellcolor[HTML]{FBE9E8}58.5 & \cellcolor[HTML]{84CDA9}13.6 & \cellcolor[HTML]{E3F3EB}41.8 \\
id & \cellcolor[HTML]{6CC399}6.5 & \cellcolor[HTML]{E88279}98.0 & \cellcolor[HTML]{F2B9B4}77.0 & \cellcolor[HTML]{57BB8A}0.0 & \cellcolor[HTML]{E4F4EC}42.0 & \cellcolor[HTML]{C7E8D8}33.5 & \cellcolor[HTML]{E6F5EE}42.8 \\
ko & \cellcolor[HTML]{95D4B5}18.5 & \cellcolor[HTML]{E78078}98.5 & \cellcolor[HTML]{CAE9DA}34.5 & \cellcolor[HTML]{75C79F}9.0 & \cellcolor[HTML]{E4F4EC}42.0 & \cellcolor[HTML]{FAE3E1}60.9 & \cellcolor[HTML]{EAF6F0}43.9 \\
tr & \cellcolor[HTML]{E67C73}100.0 & \cellcolor[HTML]{EEA59F}84.4 & \cellcolor[HTML]{F4C5C1}72.5 & \cellcolor[HTML]{58BB8B}0.5 & \cellcolor[HTML]{9DD7BB}21.0 & \cellcolor[HTML]{86CEAB}14.1 & \cellcolor[HTML]{FAFDFC}48.8 \\
pa & \cellcolor[HTML]{E77E75}99.5 & \cellcolor[HTML]{FCEFEE}56.3 & \cellcolor[HTML]{E67C73}100.0 & \cellcolor[HTML]{57BB8A}0.0 & \cellcolor[HTML]{D6EEE2}38.0 & \cellcolor[HTML]{7AC9A2}10.6 & \cellcolor[HTML]{FFFEFD}50.7 \\
vi & \cellcolor[HTML]{84CDA9}13.6 & \cellcolor[HTML]{E77F76}99.0 & \cellcolor[HTML]{F0B1AB}80.0 & \cellcolor[HTML]{78C8A1}10.0 & \cellcolor[HTML]{E67C73}100.0 & \cellcolor[HTML]{5FBE8F}2.6 & \cellcolor[HTML]{FFFDFD}50.9 \\
te & \cellcolor[HTML]{CDEADC}35.2 & \cellcolor[HTML]{EB928A}92.0 & \cellcolor[HTML]{F7D0CD}68.0 & \cellcolor[HTML]{B5E1CB}28.0 & \cellcolor[HTML]{CFEBDE}36.0 & \cellcolor[HTML]{FDF2F1}55.0 & \cellcolor[HTML]{FEF9F9}52.4 \\
ar & \cellcolor[HTML]{A7DBC1}23.9 & \cellcolor[HTML]{E8837B}97.5 & \cellcolor[HTML]{E67C73}100.0 & \cellcolor[HTML]{57BB8A}0.0 & \cellcolor[HTML]{E67C73}100.0 & \cellcolor[HTML]{A2D9BE}22.6 & \cellcolor[HTML]{FCECEB}57.3 \\
eu & \cellcolor[HTML]{E67C73}100.0 & \cellcolor[HTML]{E78078}98.5 & \cellcolor[HTML]{F8D6D3}66.0 & \cellcolor[HTML]{91D2B2}17.5 & \cellcolor[HTML]{ACDDC5}25.5 & \cellcolor[HTML]{D9EFE4}38.7 & \cellcolor[HTML]{FCEBEA}57.7 \\
bn & \cellcolor[HTML]{EC9790}89.9 & \cellcolor[HTML]{EA8C84}94.0 & \cellcolor[HTML]{EB948D}91.0 & \cellcolor[HTML]{75C79F}9.0 & \cellcolor[HTML]{F9FCFB}48.5 & \cellcolor[HTML]{AADCC3}24.7 & \cellcolor[HTML]{FBE7E5}59.5 \\
ur & \cellcolor[HTML]{EB948D}90.9 & \cellcolor[HTML]{FCECEB}57.5 & \cellcolor[HTML]{F2BBB6}76.1 & \cellcolor[HTML]{8ED1B0}16.5 & \cellcolor[HTML]{E67C73}100.0 & \cellcolor[HTML]{95D4B5}18.7 & \cellcolor[HTML]{FBE5E4}59.9 \\
my & \cellcolor[HTML]{E77F76}99.0 & \cellcolor[HTML]{EDA19B}86.0 & \cellcolor[HTML]{EFA9A3}83.0 & \cellcolor[HTML]{76C7A0}9.5 & \cellcolor[HTML]{EA8E86}93.5 & \cellcolor[HTML]{57BB8A}0.0 & \cellcolor[HTML]{FAE1DE}61.8 \\
kk & \cellcolor[HTML]{EC9992}88.9 & \cellcolor[HTML]{E77F76}99.0 & \cellcolor[HTML]{EFAAA5}82.5 & \cellcolor[HTML]{58BB8B}0.5 & \cellcolor[HTML]{E67C73}100.0 & \cellcolor[HTML]{90D2B1}17.0 & \cellcolor[HTML]{F8D9D6}64.7 \\
az & \cellcolor[HTML]{E88279}98.0 & \cellcolor[HTML]{F3BEB9}75.1 & \cellcolor[HTML]{F4C2BE}73.5 & \cellcolor[HTML]{66C194}4.5 & \cellcolor[HTML]{DFF2E8}40.5 & \cellcolor[HTML]{E77E75}99.5 & \cellcolor[HTML]{F8D8D5}65.2 \\
jv & \cellcolor[HTML]{75C79F}9.0 & \cellcolor[HTML]{E67C73}100.0 & \cellcolor[HTML]{EFACA6}82.0 & \cellcolor[HTML]{62BF92}3.5 & \cellcolor[HTML]{E67C73}100.0 & \cellcolor[HTML]{E67C73}100.0 & \cellcolor[HTML]{F8D6D3}65.8 \\
mr & \cellcolor[HTML]{57BB8A}0.0 & \cellcolor[HTML]{EFA9A3}83.0 & \cellcolor[HTML]{EFAAA4}82.8 & \cellcolor[HTML]{E67C73}100.0 & \cellcolor[HTML]{EEF8F3}45.0 & \cellcolor[HTML]{EDA099}86.4 & \cellcolor[HTML]{F7D5D2}66.2 \\
gu & \cellcolor[HTML]{E67C73}100.0 & \cellcolor[HTML]{E67C73}100.0 & \cellcolor[HTML]{E67C73}100.0 & \cellcolor[HTML]{DAF0E5}39.0 & \cellcolor[HTML]{EB978F}90.0 & \cellcolor[HTML]{C1E6D4}31.8 & \cellcolor[HTML]{F2B9B4}76.8 \\
ja & \cellcolor[HTML]{E9877F}96.0 & \cellcolor[HTML]{E67C73}100.0 & \cellcolor[HTML]{FAE4E2}60.5 & \cellcolor[HTML]{E77F76}99.0 & \cellcolor[HTML]{BBE3D0}30.0 & \cellcolor[HTML]{E98880}95.5 & \cellcolor[HTML]{F0B0AB}80.2 \\
zh & \cellcolor[HTML]{E98B83}94.4 & \cellcolor[HTML]{E67C73}100.0 & \cellcolor[HTML]{C5E7D7}33.0 & \cellcolor[HTML]{E67C73}100.0 & \cellcolor[HTML]{E98B83}94.5 & \cellcolor[HTML]{F5C8C4}71.2 & \cellcolor[HTML]{EFABA5}82.2 \\
sw & \cellcolor[HTML]{E67C73}100.0 & \cellcolor[HTML]{E67C73}100.0 & \cellcolor[HTML]{EA8C84}94.0 & \cellcolor[HTML]{6BC398}6.0 & \cellcolor[HTML]{E67C73}100.0 & \cellcolor[HTML]{E78279}98.0 & \cellcolor[HTML]{EFA9A3}83.0 \\
th & \cellcolor[HTML]{EB938C}91.4 & \cellcolor[HTML]{F1B5B0}78.4 & \cellcolor[HTML]{E67C73}100.0 & \cellcolor[HTML]{E67C73}100.0 & \cellcolor[HTML]{E67C73}100.0 & \cellcolor[HTML]{E0F2E9}41.0 & \cellcolor[HTML]{EEA39D}85.1 \\
he & \cellcolor[HTML]{E67C73}100.0 & \cellcolor[HTML]{E8837B}97.5 & \cellcolor[HTML]{E67C73}100.0 & \cellcolor[HTML]{E67C73}100.0 & \cellcolor[HTML]{E67C73}100.0 & \cellcolor[HTML]{B4E0CB}27.9 & \cellcolor[HTML]{ED9D96}87.6 \\
qu & \cellcolor[HTML]{EB928A}91.9 & \cellcolor[HTML]{E67C73}100.0 & \cellcolor[HTML]{E67C73}100.0 & \cellcolor[HTML]{E78279}98.0 & \cellcolor[HTML]{E98880}95.5 & \cellcolor[HTML]{E8847C}97.0 & \cellcolor[HTML]{E8847C}97.1 \\
ht & \cellcolor[HTML]{E98880}95.5 & \cellcolor[HTML]{E67C73}100.0 & \cellcolor[HTML]{E67C73}100.0 & \cellcolor[HTML]{E67C73}100.0 & \cellcolor[HTML]{E67C73}100.0 & \cellcolor[HTML]{EB948D}90.9 & \cellcolor[HTML]{E8827A}97.7 \\
ha & \cellcolor[HTML]{E67C73}100.0 & \cellcolor[HTML]{E67C73}100.0 & \cellcolor[HTML]{E77E75}99.5 & \cellcolor[HTML]{E67C73}100.0 & \cellcolor[HTML]{E67C73}100.0 & \cellcolor[HTML]{EB938B}91.5 & \cellcolor[HTML]{E78078}98.5 \\
yo & \cellcolor[HTML]{E67C73}100.0 & \cellcolor[HTML]{E67C73}100.0 & \cellcolor[HTML]{E67C73}100.0 & \cellcolor[HTML]{E67C73}100.0 & \cellcolor[HTML]{E67C73}100.0 & \cellcolor[HTML]{E77E75}99.5 & \cellcolor[HTML]{E77D74}99.9 \\
wo & \cellcolor[HTML]{E67C73}100.0 & \cellcolor[HTML]{E67C73}100.0 & \cellcolor[HTML]{E67C73}100.0 & \cellcolor[HTML]{E67C73}100.0 & \cellcolor[HTML]{E67C73}100.0 & \cellcolor[HTML]{E67C73}100.0 & \cellcolor[HTML]{E67C73}100.0 \\
Avg & \cellcolor[HTML]{F5FBF8}47.3 & \cellcolor[HTML]{EB948D}90.9 & \cellcolor[HTML]{F8D9D6}64.8 & \cellcolor[HTML]{B0DFC8}26.7 & \cellcolor[HTML]{FFFBFB}51.6 & \cellcolor[HTML]{C5E7D6}32.8 & \cellcolor[HTML]{FEF9F9}52.4 \\
\bottomrule
\end{tabular}%
 }
\caption{Error rate of XLM-R fine-tuned on English SQuAD v1.1 on 6 \textsc{CheckList} QA tests.}
\label{tab:checklist_xmlr}
\end{table}

\begin{table*}[!htb]
  \centering \tiny
    % \addtolength{\tabcolsep}{-1pt} 
    % \setlength{\tabcolsep}{-1pt}
    \renewcommand\tabcolsep{0.72pt}
    \renewcommand\arraystretch{0.77}  
    \begin{tabular}{cccccccc cccccccc cccccccc cccccccc cccccccc cccccccc cccccccc cccccccc cccccc cc cccccccc cccccccc ccccccc cccccccccccc} 
    \toprule
          & \multicolumn{17}{c}{\texttt{aLen}}                           & \multicolumn{17}{c}{\texttt{qLen}}                            & \multicolumn{17}{c}{\texttt{cLen}}                             &
          \multicolumn{17}{c}{\texttt{BLEU-AQ}}  &
          \multicolumn{17}{c}{\texttt{BLEu-QC}}  & \multicolumn{17}{c}{\texttt{qType}}\\
    \midrule
    % \cmidrule(lr){2-107}
    % \cmidrule(lr){1-1}
    & &&&&&&& 
    en & zh & hi & el & ru &  tr & ar &  vi &
    % &&&&&&&
     &&&&&&& && 
      en & zh & hi & el & ru &  tr & ar &  vi &
    %  \multicolumn{1}{l}{\rotatebox{90}{en}} & 
    % \multicolumn{1}{l}{\rotatebox{90}{es}} & 
    % \multicolumn{1}{l}{\rotatebox{90}{de}} & 
    % \multicolumn{1}{l}{\rotatebox{90}{el}} & 
    % \multicolumn{1}{l}{\rotatebox{90}{ru}} & 
    % \multicolumn{1}{l}{\rotatebox{90}{tr}} & 
    % \multicolumn{1}{l}{\rotatebox{90}{ar}} & 
    % \multicolumn{1}{l}{\rotatebox{90}{vi}} &
    % &&&&&&&
    &&&&&&& &&
   en & zh & hi & el & ru &  tr & ar &  vi &
    &&&&&&& &&
     en & zh & hi & el & ru &  tr & ar &  vi &
    &&&&&&& &&
     en & zh & hi & el & ru &  tr & ar &  vi &
    &&&&&&& &&
    en & zh & hi & el & ru &  tr & ar &  vi &
    \\
    \midrule
    % &&&&&&&
% \midrule % bert self-diagnose
%     \multicolumn{1}{l}{Overall F1} & 
%      \multicolumn{15}{c}{M1: 91.11} &
%      \multicolumn{15}{c}{M1: 47.77} &
%      \multicolumn{15}{c}{M1: 86.90}  &
%      \multicolumn{15}{c}{M1: 81.03} &
%      \multicolumn{15}{c}{M1: 89.64}&
%      \multicolumn{15}{c}{M1: 66.35} \\
    % \cmidrule(r){1-1}\cmidrule(lr){2-16}\cmidrule(lr){17-33}\cmidrule(lr){34-49}\cmidrule(lr){49-63}\cmidrule(lr){64-79}\cmidrule(lr){80-95} %\cmidrule(lr){86-99}
    
     & \multicolumn{17}{l}{\multirow{12}[2]{*}{\includegraphics[scale=0.28]{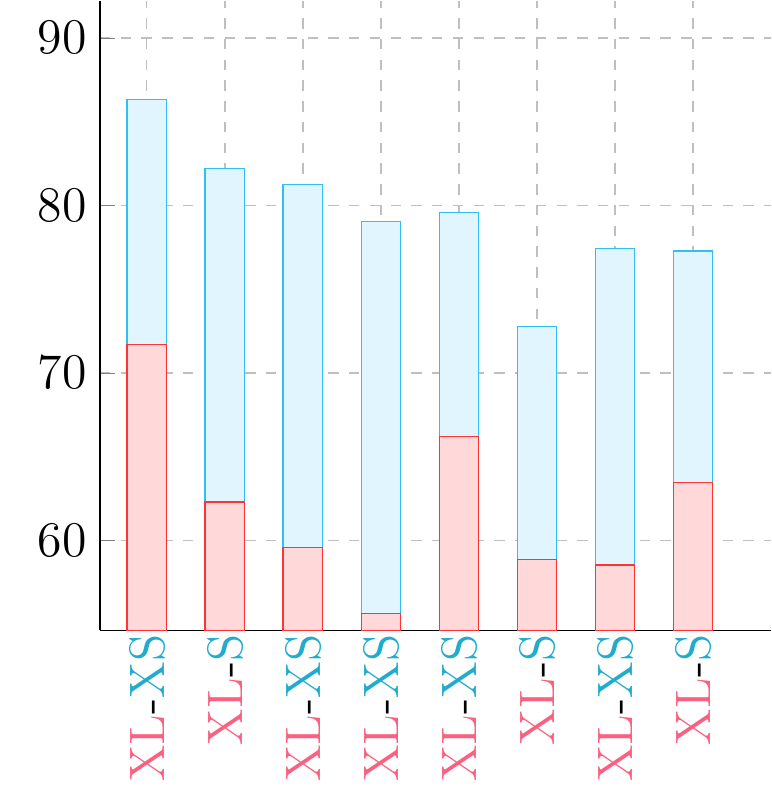}}}              
    & \multicolumn{17}{l}{\multirow{12}[2]{*}{\includegraphics[scale=0.28]{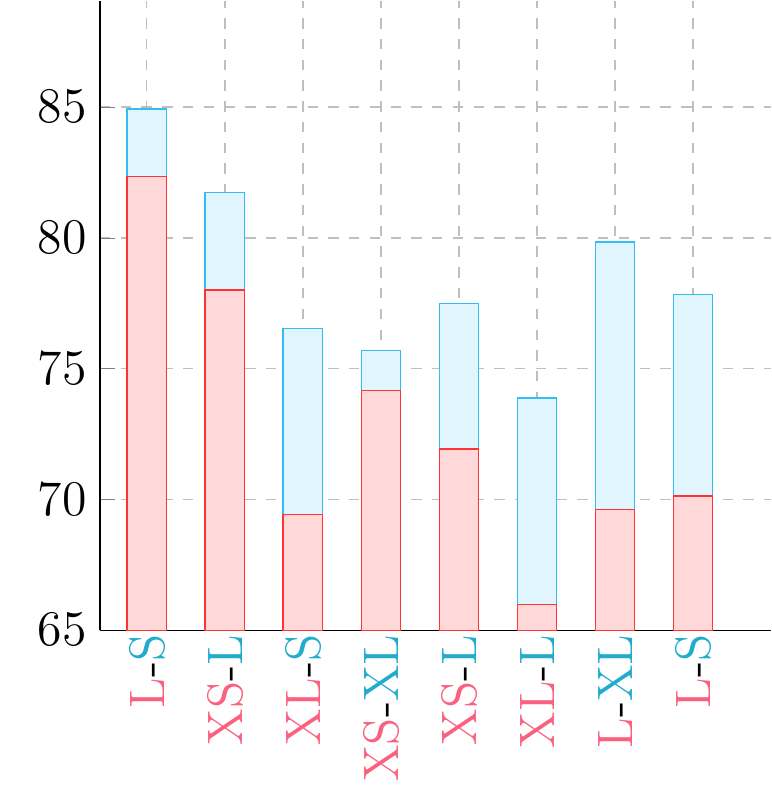}}}              
    & \multicolumn{17}{l}{\multirow{12}[2]{*}{\includegraphics[scale=0.28]{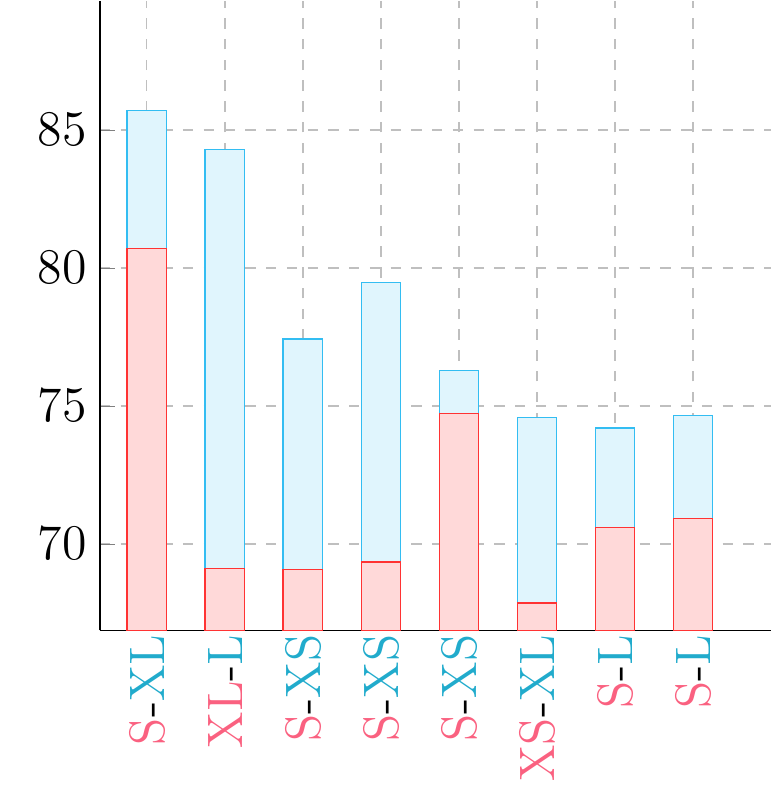}}}              
    &\multicolumn{17}{l}{\multirow{12}[2]{*}{\includegraphics[scale=0.28]{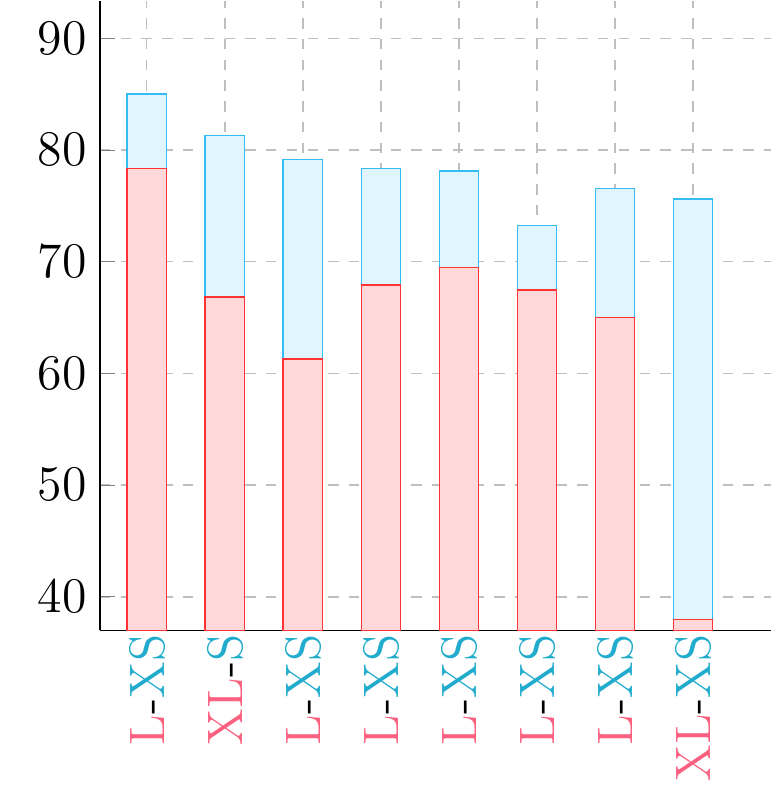}}} 
    &\multicolumn{17}{l}{\multirow{12}[2]{*}{\includegraphics[scale=0.28]{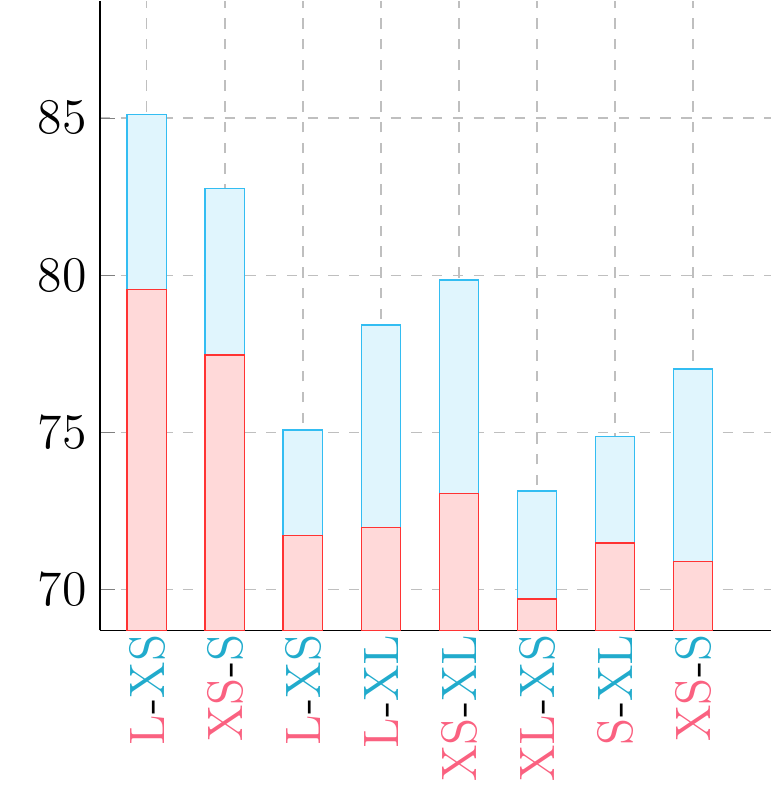}}} 
    & \multicolumn{17}{l}{\multirow{12}[2]{*}{\includegraphics[scale=0.28]{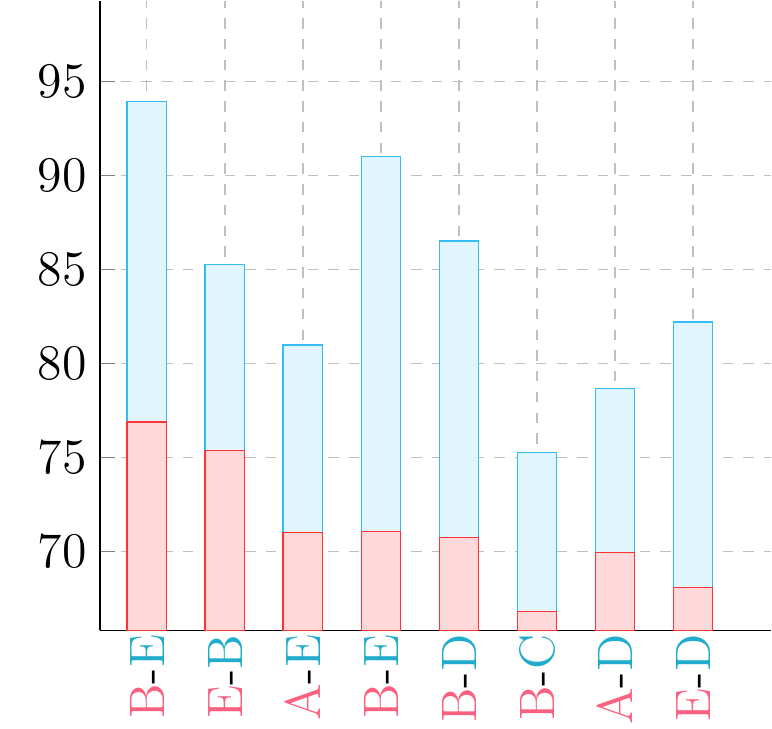}}} 
    \\ \\ \\ 
    \multicolumn{1}{l}{M1: \textit{ERNIE-M}} & \\ 
    \multicolumn{1}{l}{(Overall: 75.46)} &  \\
    \\  
    \multicolumn{1}{c}{\textbf{Single Sys.}} & 
    \\ \\ \\ \\  \\ \\
    
    \midrule
    % & \multicolumn{15}{l}{\multirow{12}[2]{*}{\includegraphics[scale=0.28]{figures/analysis/single/XLMR/aLen.pdf}}}              
    % & \multicolumn{15}{l}{\multirow{12}[2]{*}{\includegraphics[scale=0.28]{figures/analysis/single/XLMR/qLen.pdf}}}              
    % & \multicolumn{15}{l}{\multirow{12}[2]{*}{\includegraphics[scale=0.28]{figures/analysis/single/XLMR/cLen.pdf}}}              
    % & \multicolumn{15}{l}{\multirow{12}[2]{*}{\includegraphics[scale=0.28]{figures/analysis/single/XLMR/aQRatio.pdf}}} 
    % &\multicolumn{15}{l}{\multirow{12}[2]{*}{\includegraphics[scale=0.28]{figures/analysis/single/XLMR/bleu_AQ.pdf}}} 
    % &\multicolumn{15}{l}{\multirow{12}[2]{*}{\includegraphics[scale=0.28]{figures/analysis/single/XLMR/bleu_QC.pdf}}} 
    % \\ \\ \\ \\ \\ \\
    % \multicolumn{1}{l}{M1: \textit{XLM-R}} & 
    % \\  \\
    % \multicolumn{1}{c}{\textbf{Self-diagnosis}} & 
    % \\ \\ \\ \\  
    
    % \midrule
    
    & \multicolumn{17}{l}{\multirow{12}[2]{*}{\includegraphics[scale=0.28]{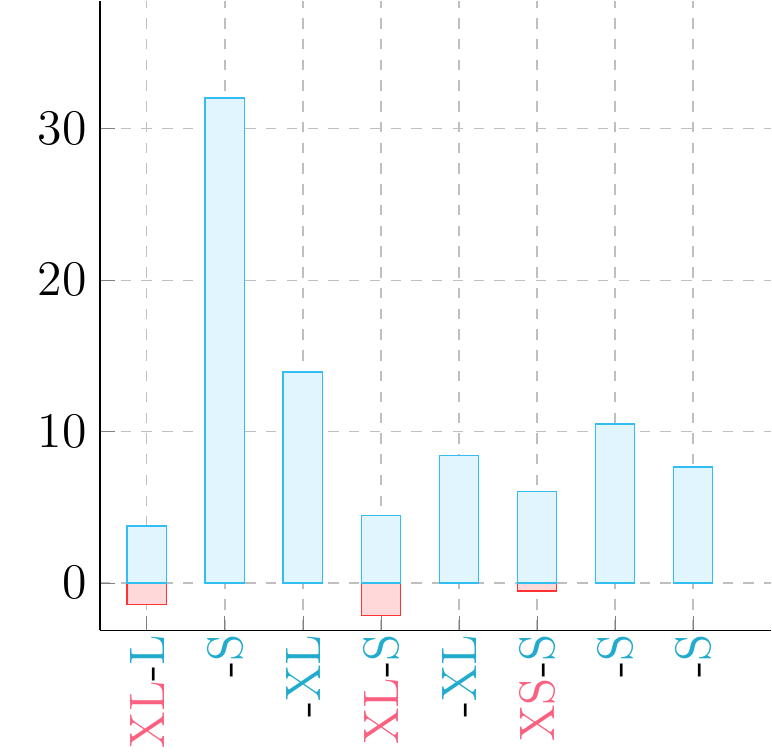}}}              
    & \multicolumn{17}{l}{\multirow{12}[2]{*}{\includegraphics[scale=0.28]{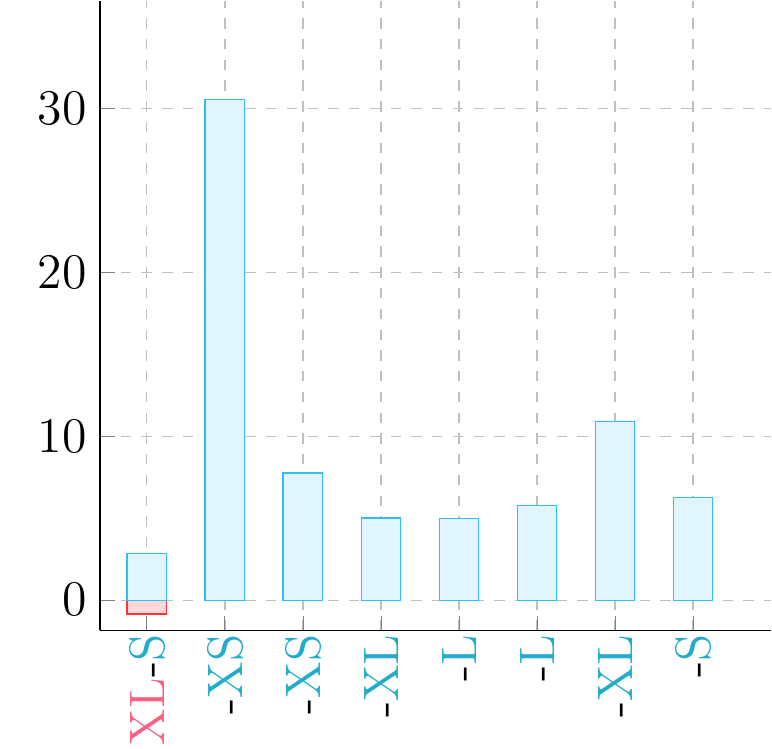}}}              
    & \multicolumn{17}{l}{\multirow{12}[2]{*}{\includegraphics[scale=0.28]{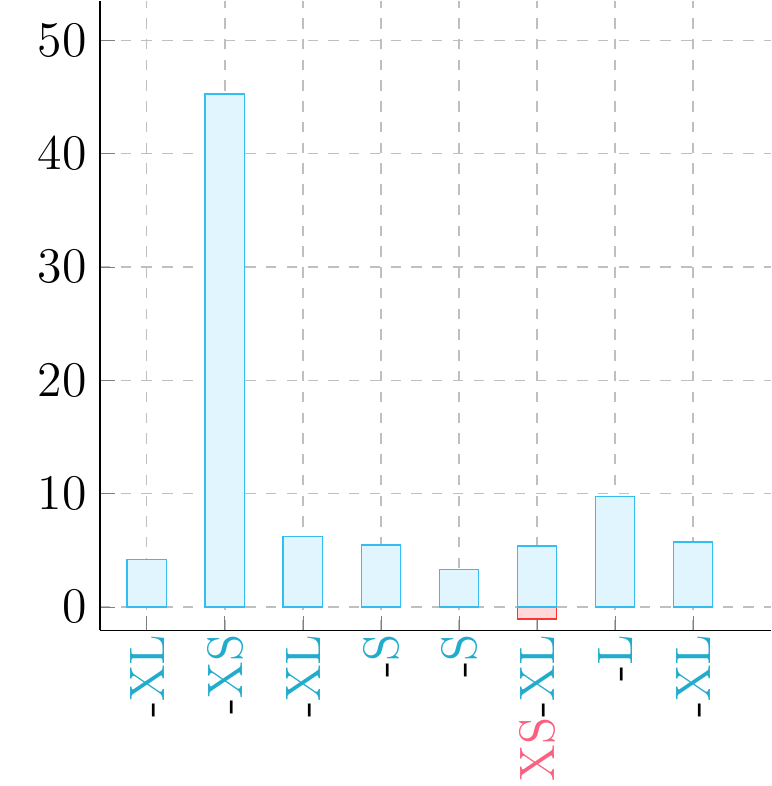}}}              
    &\multicolumn{17}{l}{\multirow{12}[2]{*}{\includegraphics[scale=0.28]{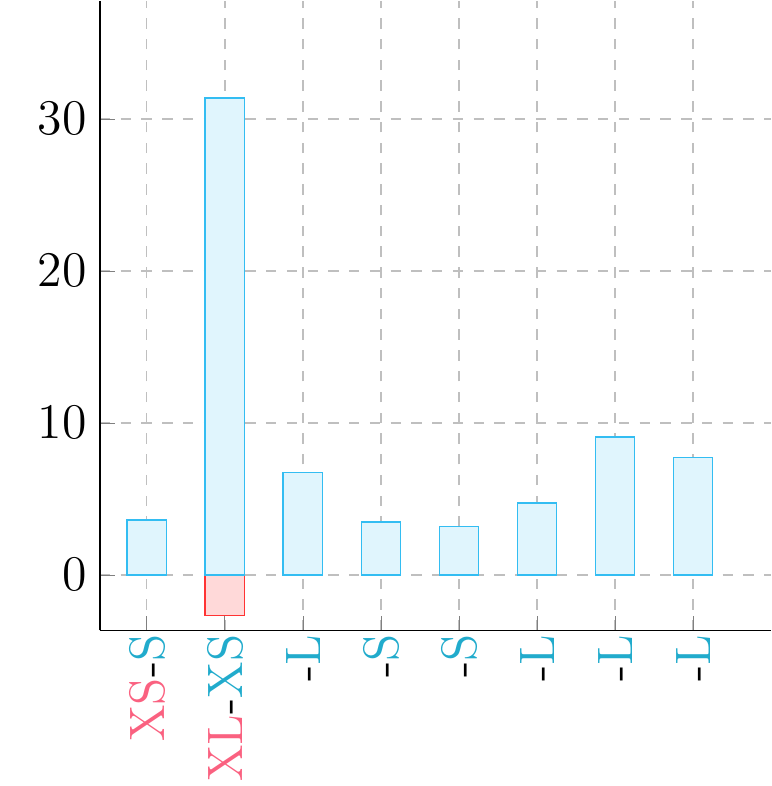}}} 
    &\multicolumn{17}{l}{\multirow{12}[2]{*}{\includegraphics[scale=0.28]{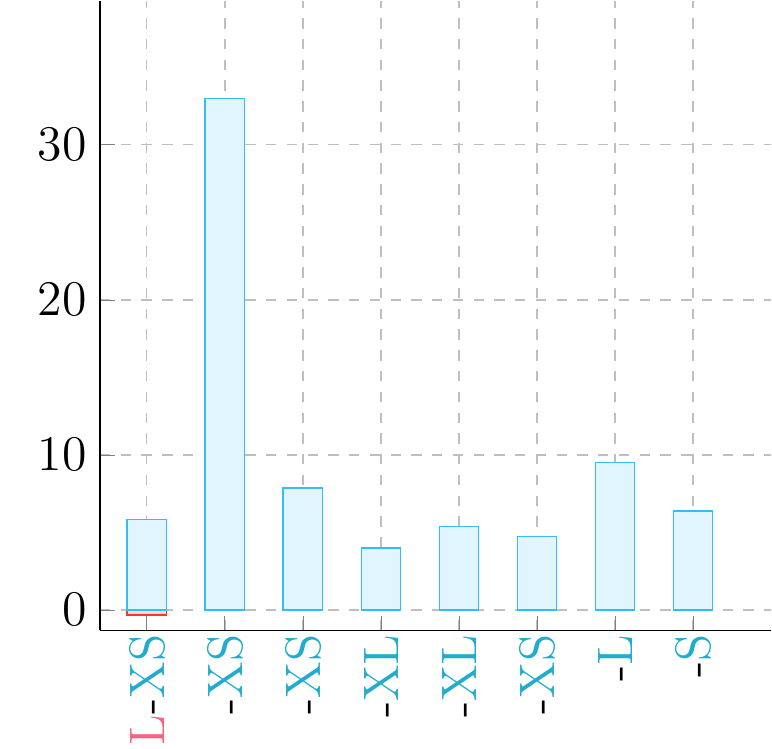}}} 
    & \multicolumn{17}{l}{\multirow{12}[2]{*}{\includegraphics[scale=0.28]{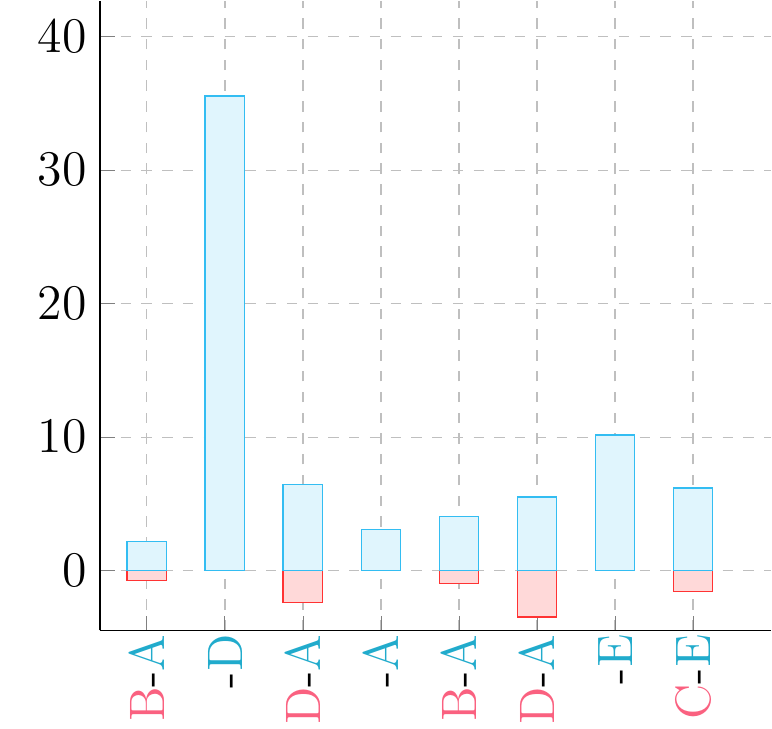}}} 
    \\ \\ 
    \multicolumn{1}{l}{M1: \textit{ERNIE-M}} &  \\
    \multicolumn{1}{l}{(Overall: 75.46)} &  \\ \\
    \multicolumn{1}{l}{M2: \textit{XLM-R}} &  \\
    \multicolumn{1}{l}{(Overall: 68.45)} &  \\ \\
    \multicolumn{1}{c}{\textbf{Pairwise Sys.}} & 
    \\ 
    \multicolumn{1}{c}{\textbf{(M1$-$M2)}} & 
    \\  \\ \\   
    
    % \midrule
    % & \multicolumn{17}{l}{\multirow{12}[2]{*}{\includegraphics[scale=0.28]{figures/analysis/pair/erniem_tulrms/aLen.pdf}}}              
    % & \multicolumn{17}{l}{\multirow{12}[2]{*}{\includegraphics[scale=0.28]{figures/analysis/pair/erniem_tulrms/qLen.pdf}}}              
    % & \multicolumn{17}{l}{\multirow{12}[2]{*}{\includegraphics[scale=0.28]{figures/analysis/pair/erniem_tulrms/cLen.pdf}}}              
    % &\multicolumn{17}{l}{\multirow{12}[2]{*}{\includegraphics[scale=0.28]{figures/analysis/pair/erniem_tulrms/bleu_AQ.pdf}}} 
    % &\multicolumn{17}{l}{\multirow{12}[2]{*}{\includegraphics[scale=0.28]{figures/analysis/pair/erniem_tulrms/bleu_QC.pdf}}} 
    % & \multicolumn{17}{l}{\multirow{12}[2]{*}{\includegraphics[scale=0.28]{figures/analysis/pair/erniem_tulrms/qtype.pdf}}} 
    % \\ \\ 
    % \multicolumn{1}{l}{M1: \textit{ERNIE-M}} &  \\
    % \multicolumn{1}{l}{(Overall: 75.46)} &  \\ \\
    % \multicolumn{1}{l}{M2: \textit{T-URLv2}} &  \\
    % \multicolumn{1}{l}{(Overall: 76.68)} &  \\ \\
    % \multicolumn{1}{c}{\textbf{Pairwise Sys.}} & 
    % \\ 
    % \multicolumn{1}{c}{\textbf{(M1$-$M2)}} & 
    % \\  \\ \\   

    \bottomrule
    \end{tabular}%
    \vspace{-5pt}
\caption{Single and pairwise system diagnosis  of {ERNIE-M} (M1) and {XLM-R} (M2) on XQuAD.  ``M1$-$M2'' represents the performance difference between M1 and M2. We average ``F1'' and ``Exact Match'' of QA systems.
We classify the attribute values into four \textit{categories}: extra-small (XS), small (S), large (L) and extra-large (XL) values (see the Appendix for detailed interval information).
The top 5 most frequent question types (\texttt{qType}) are: ``what'' (A), ``how'' (B), ``who'' (C), ``when'' (D) and ``which'' (E).
% ranked by their frequencies in the training set. 
In the \textit{single system diagnosis} histogram, \textcolor{ballblue}{blue} (\textcolor{brinkpink}{red}) $x$ ticklabels represent the bucket category (e.g., XS) of a specific attribute on which a system achieved the best (worst) performance.
% Red bins represent worst performance while  blue bins denote the gap between best and worst performance.
In the \textit{pairwise system diagnosis} histogram, \textcolor{ballblue}{blue} (\textcolor{brinkpink}{red}) $x$ ticklabels represent the bucket value of a specific attribute where system M1 surpasses (under-performs) M2 by the largest margin illustrated by a \textcolor{ballblue}{blue} (\textcolor{brinkpink}{red}) bin. \textcolor{ballblue}{Blue}-only $x$ ticklabels (e.g., \textcolor{ballblue}{-D}) indicate that M1 outperforms M2 in all categories of an attribute. 
 }
  \label{tab:bucket-wise}%
\end{table*}%

% \noindent \textbf{Overall results} $\:$ 
We show the main results in Table \ref{tab:main-results}. As in prior work, XLM-R Large generally outperforms mBERT. Fine-tuning helps significantly on Tatoeba compared to the zero-shot setting \cite{Hu2020xtreme}. The new tasks are challenging for current models, which show relatively lower performance compared to other tasks. 
XCOPA presents a challenging classification task that requires cross-lingual common sense reasoning while the language-agnostic nature of Mewsli-X and LAReQA puts the cross-lingual alignment of multilingual representations to the test. 
Analysis of the language-agnostic retrieval results show a large gap remains between cross-lingual and same-language test cases. XLM-R Large improves significantly over mBERT on the cross-lingual case 
% (+38\% on Mewsli-X and +137\% for LAReQA) 
in exchange for a slight drop for the same-language case. This points to XLM-R Large inducing more ``strongly-aligned'' representations
% , consistent with its cross-lingual pretraining loss 
(see Appendix \ref{app:retrieval-results} for details).
The state-of-the-art mT5 improves performance on classification and QA tasks but performs less well on structured prediction and retrieval, highlighting settings where advances beyond scale are needed.\footnote{Due to compute limitations, we extract mT5 embeddings by averaging the encoder outputs of a frozen mT5 model fine-tuned on SQuAD v1.1, as opposed to fine-tuning a dual encoder. For this reason, the mT5 language-agnostic retrieval scores are not directly comparable to those of mBERT and XLM-R.} 
Training on task-specific translations is beneficial in all cases and generally performs best, although improvements on QA tasks are diminishing. To obtain a more fine-grained understanding of the performance of current models, we conduct several analyses using our multilingual diagnostic suite.

\section{Analyses}

\subsection{\multichecklist}

We show the results of XLM-R fine-tuned on English SQuAD v1.1 on the 6 tests of \multichecklist in Table \ref{tab:checklist_xmlr} (see Appendix \ref{app:multichecklist} for the full results, example failure cases, and mBERT results). While mBERT's average error rate is greater than 85\% on 4/6 test categories, XLM-R demonstrates a substantially more robust cross-lingual understanding ability. 
XLM-R performs worst on tests in low-resource languages with limited or no pre-training data such as \texttt{gu}, \texttt{ha}, \texttt{ht}, \texttt{qu}, \texttt{sw}, \texttt{wo}, and \texttt{yo} and in languages with non-Latin scripts such as \texttt{he}, \texttt{ja}, \texttt{th}, and \texttt{zh}. 
% XLM-R consistently performs worse on tests in low-resource languages with limited or no pre-training data such as Gujarati, Hausa, Haitian Creole, Quechua, Swahili, Wolof, and Yoruba and in languages with non-Latin scripts such as Hebrew, Japanese, Thai, and Chinese. 
% Scores on the Intensifiers test are comparatively low for both models, as the 
% as each test consists of 12 variations and models need to succeed on all of them to pass.
In addition, XLM-R displays interesting variation across languages, for instance failing in modeling comparisons in some languages, like Basque (\texttt{eu}), where it otherwise succeeds. We release the tests and test outputs to encourage deeper analysis and extension to other tasks and languages.

%\subsection{Multilingual \explainaboard}
\subsection{Nuanced Multilingual Evaluation}

% \pfliu{to-do: (1) sections should be re-organized (2) even three languages seems too many?  (3) where should we describe our attributes?  Suggestions: we could only focus on QA task in our main text and move the NER to appendix.}

% \jh{Re (2) It is better to have one closed language and one distant language w.r.t. English. Re (3)We could just sample a few attributes here and leave the complete list to the Appendix. Some attributes are correlated.}\pfliu{Thanks!}
% \pfliu{Quick question1: Here the subsection title is \explainaboard, but I start the description with the concept of ``fine-grained evaluation''. Is there a good idea to make a nice transition from \explainaboard to ``fine-grained evaluation''.  Or we could mention \explainaboard late after we finish the description of fine-grained evaluation analysis?}

% \pfliu{Quick question2: Also, should we post current demo url somewhere? Or we don't need to do it now. So far, my understanding of it is, basically (1) we submit the main draft with url link (2) we have one week after the deadline to refine our appendix (3) we have more weeks to refine our demo. I'm sorry if I have some mis-understanding and feel free to correct me.}

% \pfliu{Quick question3: Knowing the architectural/training differences between ERNIE-M and XLM-R would be much helpful for further analysis. So, (1) is there a pointer to the XLM-R paper?  (2) It would be super nice if who (lol) can highlight their major differences briefly.}

We showcase how nuanced multilingual evaluation enables us to perform \textit{single} and \textit{pairwise} system diagnosis on XQuAD in Table~\ref{tab:bucket-wise} (see Appendix \ref{app:nuanced-evaluation} for analyses of the other tasks).
% The basic idea of nuanced evaluations is to break down the holistic performance of models on \name into different interpretable groups based on diverse task-dependent attributes.
% In the following, we elaborate on six attributes defined for the QA task (XQuAD) and report the analyses of the other tasks in Appendix \ref{app:nuanced-evaluation}. 
% Specifically, we showcase how nuanced multilingual evaluation enables us to perform \textit{single} and \textit{pairwise} system diagnosis in Table~\ref{tab:bucket-wise}. 
We choose two {systems}: {ERNIE-M}, one of the top systems on \xtreme,  and {XLM-R} in eight {languages}: English, Chinese, Hindi, Greek, Russian, Turkish, Arabic, and Vietnamese (\texttt{en}, \texttt{zh}, \texttt{hi}, \texttt{el}, \texttt{ru}, \texttt{tr}, \texttt{ar}, \texttt{vi}).

% \pfliu{Due to limited pages, I condensed the description of attributes. Feel free to let me know if this is not clear.}
\noindent \textbf{Attributes} $\:$ We denote $(\mathbf{X}_c, \mathbf{X}_q, \mathbf{X}_a)$ as a tuple of a context, question and answer, and refer to \texttt{cLen}, \texttt{qLen}, \texttt{aLen} as their lengths (i.e., the number of tokens).
We use BLEU \cite{papineni-etal-2002-bleu} to measure lexical overlap between $(\mathbf{X}_a, \mathbf{X}_q)$ and  $(\mathbf{X}_q, \mathbf{X}_c)$ as \texttt{BLEU-AQ} and \texttt{BLEU-QC}. We 
% classify questions based on their first tokens and 
report the top 5 most frequent question types (\texttt{qType}), which cover 85\% of questions in the training set. 

% \noindent \textbf{Setup}$\:$ We showcase how the fine-grained evaluation enables us to perform \textit{single} and \textit{pair-wise} system diagnosis.
% Specifically, we focus on the QA {task} (XQuAD) and choose two {systems}: {ERNIE-M}, which has achieved the SOTA holistic performance on \xtreme,  and {XLM-R} in eight {languages}: \texttt{en, zh, hi, el, ru, tr, ar, vi}.

\noindent \textbf{Single System Analysis} $\:$ 
% The first row in Table~\ref{tab:bucket-wise} shows the single system diagnosis of {ERNIE-M}. 
% We observe that:
\noindent 
% (1) 
For almost all languages, {ERNIE-M} achieves the highest performance on shorter answers (\texttt{XS}), but the worst performance on longer answers (\texttt{XL}). Especially in \texttt{el}, the performance difference between long and short answers is more than $40$ absolute points.
% (2) 
The influence of question and context length is language-dependent. For example, in \texttt{zh} the system favors long questions and contexts while in \texttt{hi},  it is the opposite.
% {ERNIE-M} favors longer questions (except in \texttt{en} and \texttt{vi} two languages), 
% and longer context (except in  \texttt{el} and \texttt{ru}).
% (3) 
If the answer is lexically similar to the question (larger \texttt{BLEU-AQ}), the system tends to make more mistakes in all eight languages. However, a higher lexical overlap between questions and contexts (\texttt{BLEU-QC}) is helpful for some languages: \texttt{el}, \texttt{ru}, \texttt{ar}.
% (4) 
Surprisingly, {ERNIE-M} struggles to answer relatively frequent question types (i.e., \texttt{what}, and \texttt{how}), while it performs better on less frequent questions, 
% (i.e., \texttt{who}, \texttt{where}, and \texttt{which}), 
indicating that although questions about person, place and choice are less frequent, they are easier than abstract questions.   
% but not in all cases (e.g., \texttt{tr} and \texttt{vi}).

% \noindent (1) Answer length is an attribute that consistently influences the system's performance, which would be decreased in all languages with the increase of answer's length, even for the state-of-the-art system.

% \noindent (2) if the answer is similar to the question in two aspects: (i) their lengths (ii) their lexical overlaps, then the system would make more wrong predictions.

% \noindent (3) the influence of other attributes is relatively language-dependent. For example, in English, shorter questions would be easier to answer, while in \texttt{es}, the system favors longer questions.

% \paragraph{XLM-R}
% The effect of attributes \texttt{aLen}, \texttt{aQRatio}, \texttt{bleu(A,Q)} on system \texttt{XLM-R} is consistent with what we have observed for system  \texttt{ERNIE-M}.
% Interesting findings here are:
% \texttt{XLM-R} longer questions and in languages \texttt{en} and \texttt{es}, questions with larger lexical overlap with context texts would be easier to answer. 

\noindent \textbf{Pairwise System Analysis} $\:$
% \pfliu{The observations will be updated once Jinlan finishes our new figure.}
% From the second row of Table~\ref{tab:bucket-wise} we find that 
Although {ERNIE-M} outperforms {XLM-R} by a large margin, it is surpassed by {XLM-R} on a few buckets.
In \texttt{en}, {XLM-R} is better at dealing with longer answers and questions.
% short answers and longer questions.
In \texttt{tr}, {XLM-R} surpasses {ERNIE-M} on samples with shorter answers and contexts.
In \texttt{zh}, 
% although ERNIE-M outperforms  XLM-R by a large margin over multiple attributes, 
XLM-R performs better when dealing with questions that are lexically similar to the answers.

\section{Conclusions}

Our analyses and experiments have shed light on important directions where scale alone is not sufficient such as ``strong'' alignment, syntactic transfer, fine-grained natural language understanding, and answering of abstract questions. We encourage the development of better inductive biases, pre-training objectives, and evaluation resources.
We make our data, translations, evaluation resources, and interactive leaderboard supporting detailed comparative analyses available to help the community gain a better understanding of multilingual models.
% The interactive leaderboard supports fine-grained analyses to enable practicioners and researchers to obtain a better understanding of multilingual models.

\section{Ethical Considerations}

\subsection{Language representation}

\name seeks to improve language representation and language diversity in NLP research, which has been identified as a large challenge \cite{Joshi2020}. We tried to cover a set of languages that is as diverse as possible, while still providing access to evaluation data in multiple tasks for each language. Despite this, \name has little representation of languages of the Americas and Africa due to a lack of labeled datasets for these languages. In addition, some languages included in \name with few data available online are only covered in a small number of datasets (see Table \ref{tab:languages}). To ameliorate this, we release training data of tasks translated into other languages, as well as the new \mbox{\multichecklist}. We reiterate the on-going need for creating labeled datasets for diverse tasks in under-represented languages, to facilitate the development and evaluation of NLP models for such languages. We emphasize the importance of participatory research \cite{nekoto_etal_2020_participatory} as a modus operandi for such work in order to involve marginalized communities in the research process.

\subsection{Leaderboard chasing}

New benchmarks incentivize researchers to hill-climb on aggregate metrics \cite{ethayarajh2020utility}. In addition, new benchmarks create new opportunities for models to reach ``superhuman'' performance, which may lead people outside the field to erroneously conclude that some model has ``solved language''. We hope that our inclusion of \explainaboard and \multichecklist help to prevent such a fallacy, by enabling more fine-grained evaluation that goes beyond a single aggregate metric.

\subsection{Biases in multilingual models}

Multilingual models have been shown to reflect biases similar to their monolingual counterparts \cite{Zhao2020bias}. In addition, multilingual models are biased towards languages with more pre-training data \cite{Hu2020xtreme}. Zero-shot cross-lingual transfer additionally introduces a bias towards the source language \cite{Sogaard2018unsupervised,Anastasopoulos2020}. Due to the paucity of training data in other languages, we nevertheless focus on English-centric transfer and encourage future dataset creation efforts to include training data in multiple languages.

\subsection{Environmental concerns}

\name aims to enable efficient evaluation of multilingual models. To this end, we created a new dataset, Mewsli-X, that captures the essence of multilingual entity linking against a diverse knowledge base but is computationally cheaper to evaluate than the large-scale Mewsli-9 \cite{botha-etal-2020-entity}. Nevertheless, the models that perform best on benchmarks like \name are generally large-scale Transformer models pre-trained on large amounts of data, which comes at a high cost \cite{Strubell2019energy}. We thus particularly encourage the development of efficient methods to adapt existing models to new languages \cite{Pfeiffer2020mad-x} rather than training multilingual models entirely from scratch.

\section*{Acknowledgements}

We thank Marco Tulio Ribeiro for advice on \checklist. We are grateful to Laura Rimell and Jon Clark for valuable feedback on drafts of this paper, and to
Dan Gillick for feedback on the \mbox{Mewsli-X} dataset design.
We thank Hila Gonen, Bidisha Samantha, and Partha Talukdar for advice on Arabic, Bengali, and Hebrew \checklist examples.

\bibliography{main_emnlp}

\begin{thebibliography}{63}
\expandafter\ifx\csname natexlab\endcsname\relax\def\natexlab#1{#1}\fi

\bibitem[{Anastasopoulos and Neubig(2020)}]{Anastasopoulos2020}
Antonios Anastasopoulos and Graham Neubig. 2020.
\newblock \href {https://doi.org/10.18653/v1/2020.acl-main.766} {{Should all
  cross-lingual embeddings speak english?}}
\newblock In \emph{Proceedings of ACL 2020}, pages 8658--8679.

\bibitem[{Artetxe et~al.(2020{\natexlab{a}})Artetxe, Ruder, and
  Yogatama}]{artetxe2020cross}
Mikel Artetxe, Sebastian Ruder, and Dani Yogatama. 2020{\natexlab{a}}.
\newblock {On the Cross-lingual Transferability of Monolingual
  Representations}.
\newblock In \emph{Proceedings of ACL 2020}.

\bibitem[{Artetxe et~al.(2020{\natexlab{b}})Artetxe, Ruder, Yogatama, Labaka,
  and Agirre}]{artetxe-etal-2020-call}
Mikel Artetxe, Sebastian Ruder, Dani Yogatama, Gorka Labaka, and Eneko Agirre.
  2020{\natexlab{b}}.
\newblock \href {https://doi.org/10.18653/v1/2020.acl-main.658} {A call for
  more rigor in unsupervised cross-lingual learning}.
\newblock In \emph{Proceedings of the 58th Annual Meeting of the Association
  for Computational Linguistics}, pages 7375--7388, Online. Association for
  Computational Linguistics.

\bibitem[{Artetxe and Schwenk(2019)}]{Artetxe2019massively}
Mikel Artetxe and Holger Schwenk. 2019.
\newblock {Massively Multilingual Sentence Embeddings for Zero-Shot
  Cross-Lingual Transfer and Beyond}.
\newblock \emph{Transactions of the ACL 2019}.

\bibitem[{Bakker et~al.(2011)Bakker, Daval-Markussen, Parkvall, and
  Plag}]{bakker2011creoles}
Peter Bakker, Aymeric Daval-Markussen, Mikael Parkvall, and Ingo Plag. 2011.
\newblock Creoles are typologically distinct from non-creoles.
\newblock \emph{Journal of Pidgin and Creole Languages}, 26(1):5--42.

\bibitem[{Belinkov and Glass(2019)}]{Belinkov2019analysis}
Yonatan Belinkov and James Glass. 2019.
\newblock {Analysis Methods in Neural Language Processing: A Survey}.
\newblock \emph{Transactions of the ACL}.

\bibitem[{Botha et~al.(2020)Botha, Shan, and Gillick}]{botha-etal-2020-entity}
Jan~A. Botha, Zifei Shan, and Daniel Gillick. 2020.
\newblock {E}ntity {L}inking in 100 {L}anguages.
\newblock In \emph{Proceedings of EMNLP 2020}, pages 7833--7845.

\bibitem[{Chi et~al.(2021)Chi, Dong, Wei, Yang, Singhal, Wang, Song, Mao,
  Huang, and Zhou}]{Chi2020infoxml}
Zewen Chi, Li~Dong, Furu Wei, Nan Yang, Saksham Singhal, Wenhui Wang, Xia Song,
  Xian-Ling Mao, Heyan Huang, and Ming Zhou. 2021.
\newblock \href {https://doi.org/10.18653/v1/2021.naacl-main.280} {{I}nfo{XLM}:
  An information-theoretic framework for cross-lingual language model
  pre-training}.
\newblock In \emph{Proceedings of the 2021 Conference of the North American
  Chapter of the Association for Computational Linguistics: Human Language
  Technologies}, pages 3576--3588, Online. Association for Computational
  Linguistics.

\bibitem[{Chung et~al.(2021)Chung, F{\'{e}}vry, Tsai, Johnson, and
  Ruder}]{Chung2021rembert}
Hyung~Won Chung, Thibault F{\'{e}}vry, Henry Tsai, Melvin Johnson, and
  Sebastian Ruder. 2021.
\newblock {Rethinking Embedding Coupling in Pre-trained Language Models}.
\newblock In \emph{Proceedings of ICLR 2021}.

\bibitem[{Clark et~al.(2020)Clark, Choi, Collins, Garrette, Kwiatkowski,
  Nikolaev, and Palomaki}]{Clark2020tydiqa}
Jonathan~H. Clark, Eunsol Choi, Michael Collins, Dan Garrette, Tom Kwiatkowski,
  Vitaly Nikolaev, and Jennimaria Palomaki. 2020.
\newblock {TyDi QA: A Benchmark for Information-Seeking Question Answering in
  Typologically Diverse Languages}.
\newblock In \emph{Transactions of the Association of Computational
  Linguistics}.

\bibitem[{Clark et~al.(2021)Clark, Garrette, Turc, and
  Wieting}]{Clark2021canine}
Jonathan~H Clark, Dan Garrette, Iulia Turc, and John Wieting. 2021.
\newblock \href {http://arxiv.org/abs/arXiv:2103.06874v2} {{CANINE:
  Pre-training an Efficient Tokenization-Free Encoder for Language
  Representation}}.
\newblock \emph{arXiv preprint arXiv:2103.06874}.

\bibitem[{Conneau et~al.(2020)Conneau, Khandelwal, Goyal, Chaudhary, Wenzek,
  Guzm{\'{a}}n, Grave, Ott, Zettlemoyer, and Stoyanov}]{Conneau2020xlmr}
Alexis Conneau, Kartikay Khandelwal, Naman Goyal, Vishrav Chaudhary, Guillaume
  Wenzek, Francisco Guzm{\'{a}}n, Edouard Grave, Myle Ott, Luke Zettlemoyer,
  and Veselin Stoyanov. 2020.
\newblock {Unsupervised Cross-lingual Representation Learning at Scale}.
\newblock In \emph{Proceedings of ACL 2020}.

\bibitem[{Conneau et~al.(2018)Conneau, Rinott, Lample, Williams, Bowman,
  Schwenk, and Stoyanov}]{Conneau2018xnli}
Alexis Conneau, Ruty Rinott, Guillaume Lample, Adina Williams, Samuel Bowman,
  Holger Schwenk, and Veselin Stoyanov. 2018.
\newblock {XNLI}: Evaluating cross-lingual sentence representations.
\newblock In \emph{Proceedings of EMNLP 2018}, pages 2475--2485.

\bibitem[{Devlin et~al.(2019)Devlin, Chang, Lee, and
  Toutanova}]{Devlin2019bert}
Jacob Devlin, Ming-Wei Chang, Kenton Lee, and Kristina Toutanova. 2019.
\newblock \href {https://doi.org/10.18653/v1/N19-1423} {{BERT}: Pre-training of
  deep bidirectional transformers for language understanding}.
\newblock In \emph{Proceedings of NAACL 2019}, pages 4171--4186, Minneapolis,
  Minnesota. Association for Computational Linguistics.

\bibitem[{Ethayarajh and Jurafsky(2020)}]{ethayarajh2020utility}
Kawin Ethayarajh and Dan Jurafsky. 2020.
\newblock \href {https://doi.org/10.18653/v1/2020.emnlp-main.393} {Utility is
  in the eye of the user: A critique of {NLP} leaderboards}.
\newblock In \emph{Proceedings of the 2020 Conference on Empirical Methods in
  Natural Language Processing (EMNLP)}, pages 4846--4853, Online. Association
  for Computational Linguistics.

\bibitem[{Fang et~al.(2021)Fang, Wang, Gan, Sun, and Liu}]{Fang2020filter}
Yuwei Fang, Shuohang Wang, Zhe Gan, Siqi Sun, and Jingjing Liu. 2021.
\newblock \href {http://arxiv.org/abs/2009.05166} {{FILTER: An Enhanced Fusion
  Method for Cross-lingual Language Understanding}}.
\newblock In \emph{Proceedings of AAAI 2021}.

\bibitem[{$\forall$ et~al.(2020)$\forall$, Nekoto, Marivate, Matsila, Fasubaa,
  Fagbohungbe, Akinola, Muhammad, Kabongo~Kabenamualu, Osei, Sackey, Niyongabo,
  Macharm, Ogayo, Ahia, Berhe, Adeyemi, Mokgesi-Selinga, Okegbemi, Martinus,
  Tajudeen, Degila, Ogueji, Siminyu, Kreutzer, Webster, Ali, Abbott, Orife,
  Ezeani, Dangana, Kamper, Elsahar, Duru, Kioko, Espoir, van Biljon, Whitenack,
  Onyefuluchi, Emezue, Dossou, Sibanda, Bassey, Olabiyi, Ramkilowan, {\"O}ktem,
  Akinfaderin, and Bashir}]{nekoto_etal_2020_participatory}
{}~$\forall$, Wilhelmina Nekoto, Vukosi Marivate, Tshinondiwa Matsila, Timi
  Fasubaa, Taiwo Fagbohungbe, Solomon~Oluwole Akinola, Shamsuddeen Muhammad,
  Salomon Kabongo~Kabenamualu, Salomey Osei, Freshia Sackey, Rubungo~Andre
  Niyongabo, Ricky Macharm, Perez Ogayo, Orevaoghene Ahia, Musie~Meressa Berhe,
  Mofetoluwa Adeyemi, Masabata Mokgesi-Selinga, Lawrence Okegbemi, Laura
  Martinus, Kolawole Tajudeen, Kevin Degila, Kelechi Ogueji, Kathleen Siminyu,
  Julia Kreutzer, Jason Webster, Jamiil~Toure Ali, Jade Abbott, Iroro Orife,
  Ignatius Ezeani, Idris~Abdulkadir Dangana, Herman Kamper, Hady Elsahar,
  Goodness Duru, Ghollah Kioko, Murhabazi Espoir, Elan van Biljon, Daniel
  Whitenack, Christopher Onyefuluchi, Chris~Chinenye Emezue, Bonaventure F.~P.
  Dossou, Blessing Sibanda, Blessing Bassey, Ayodele Olabiyi, Arshath
  Ramkilowan, Alp {\"O}ktem, Adewale Akinfaderin, and Abdallah Bashir. 2020.
\newblock \href {https://www.aclweb.org/anthology/2020.findings-emnlp.195}
  {Participatory research for low-resourced machine translation: A case study
  in {A}frican languages}.
\newblock In \emph{Findings of the Association for Computational Linguistics:
  EMNLP 2020}, Online.

\bibitem[{Fu et~al.(2020)Fu, Liu, Neubig, and Tab}]{Fu2020explainaboard}
Jinlan Fu, Pengfei Liu, Graham Neubig, and Mo~Tab. 2020.
\newblock {Interpretable Multi-dataset Evaluation for Named Entity
  Recognition}.
\newblock In \emph{Proceedings of EMNLP 2020}, pages 6058--6069.

\bibitem[{Gillick et~al.(2018)Gillick, Presta, and Tomar}]{gillick2018end}
Daniel Gillick, Alessandro Presta, and Gaurav~Singh Tomar. 2018.
\newblock End-to-end retrieval in continuous space.
\newblock \emph{arXiv preprint arXiv:1811.08008}.

\bibitem[{Gulordava et~al.(2018)Gulordava, Bojanowski, Grave, Linzen, and
  Baroni}]{Gulordava2018colorless}
Kristina Gulordava, Piotr Bojanowski, Edouard Grave, Tal Linzen, and Marco
  Baroni. 2018.
\newblock {Colorless green recurrent networks dream hierarchically}.
\newblock In \emph{Proceedings of NAACL-HLT 2018}.

\bibitem[{He et~al.(2021)He, Liu, Gao, and Chen}]{He2021deberta}
Pengcheng He, Xiaodong Liu, Jianfeng Gao, and Weizhu Chen. 2021.
\newblock {DeBERTa: Decoding-enhanced BERT with Disentangled Attention}.
\newblock In \emph{Proceedings of ICLR 2021}.

\bibitem[{Hedderich et~al.(2020)Hedderich, Adelani, Zhu, Alabi, Markus, and
  Klakow}]{Hedderich2020transfer}
Michael~A. Hedderich, David Adelani, Dawei Zhu, Jesujoba Alabi, Udia Markus,
  and Dietrich Klakow. 2020.
\newblock {Transfer Learning and Distant Supervision for Multilingual
  Transformer Models: A Study on African Languages}.
\newblock In \emph{Proceedings of EMNLP 2020}.

\bibitem[{Hu et~al.(2020)Hu, Ruder, Siddhant, Neubig, Firat, and
  Johnson}]{Hu2020xtreme}
Junjie Hu, Sebastian Ruder, Aditya Siddhant, Graham Neubig, Orhan Firat, and
  Melvin Johnson. 2020.
\newblock {XTREME: A Massively Multilingual Multi-task Benchmark for Evaluating
  Cross-lingual Generalization}.
\newblock In \emph{Proceedings of ICML 2020}.

\bibitem[{Jiang et~al.(2020)Jiang, Anastasopoulos, Araki, Ding, and
  Neubig}]{jiang2020x-factr}
Zhengbao Jiang, Antonios Anastasopoulos, Jun Araki, Haibo Ding, and Graham
  Neubig. 2020.
\newblock \href {https://doi.org/10.18653/v1/2020.emnlp-main.479} {{X}-{FACTR}:
  Multilingual factual knowledge retrieval from pretrained language models}.
\newblock In \emph{Proceedings of the 2020 Conference on Empirical Methods in
  Natural Language Processing (EMNLP)}, pages 5943--5959, Online. Association
  for Computational Linguistics.

\bibitem[{Joshi et~al.(2020)Joshi, Santy, Budhiraja, Bali, and
  Choudhury}]{Joshi2020}
Pratik Joshi, Sebastin Santy, Amar Budhiraja, Kalika Bali, and Monojit
  Choudhury. 2020.
\newblock {The State and Fate of Linguistic Diversity and Inclusion in the NLP
  World}.
\newblock In \emph{Proceedings of ACL 2020}.

\bibitem[{Kakwani et~al.(2020)Kakwani, Kunchukuttan, Golla, N.C.,
  Bhattacharyya, Khapra, and Kumar}]{kakwani-etal-2020-indicnlpsuite}
Divyanshu Kakwani, Anoop Kunchukuttan, Satish Golla, Gokul N.C., Avik
  Bhattacharyya, Mitesh~M. Khapra, and Pratyush Kumar. 2020.
\newblock \href {https://doi.org/10.18653/v1/2020.findings-emnlp.445}
  {{I}ndic{NLPS}uite: Monolingual corpora, evaluation benchmarks and
  pre-trained multilingual language models for {I}ndian languages}.
\newblock In \emph{Findings of the Association for Computational Linguistics:
  EMNLP 2020}, pages 4948--4961, Online. Association for Computational
  Linguistics.

\bibitem[{Keung et~al.(2020)Keung, Lu, Salazar, and
  Bhardwaj}]{keung-etal-2020-dont}
Phillip Keung, Yichao Lu, Julian Salazar, and Vikas Bhardwaj. 2020.
\newblock \href {https://doi.org/10.18653/v1/2020.emnlp-main.40} {Don{'}t use
  {E}nglish dev: On the zero-shot cross-lingual evaluation of contextual
  embeddings}.
\newblock In \emph{Proceedings of the 2020 Conference on Empirical Methods in
  Natural Language Processing (EMNLP)}, pages 549--554, Online. Association for
  Computational Linguistics.

\bibitem[{Khanuja et~al.(2020)Khanuja, Dandapat, and
  Srinivasan}]{Khanuja2020gluecos}
Simran Khanuja, Sandipan Dandapat, and Anirudh Srinivasan. 2020.
\newblock {GLUECoS : An Evaluation Benchmark for Code-Switched NLP}.
\newblock In \emph{Proceedings of ACL 2020}, pages 3575--3585.

\bibitem[{Khashabi et~al.(2020)Khashabi, Cohan, Shakeri, Hosseini, Pezeshkpour,
  Alikhani, Aminnaseri, Bitaab, Brahman, Ghazarian, Gheini, Kabiri, Mahabadi,
  Memarrast, Mosallanezhad, Noury, Raji, Rasooli, Sadeghi, Azer, Samghabadi,
  Shafaei, Sheybani, Tazarv, and Yaghoobzadeh}]{Khashabi2020parsinlu}
Daniel Khashabi, Arman Cohan, Siamak Shakeri, Pedram Hosseini, Pouya
  Pezeshkpour, Malihe Alikhani, Moin Aminnaseri, Marzieh Bitaab, Faeze Brahman,
  Sarik Ghazarian, Mozhdeh Gheini, Arman Kabiri, Rabeeh~Karimi Mahabadi, Omid
  Memarrast, Ahmadreza Mosallanezhad, Erfan Noury, Shahab Raji, Mohammad~Sadegh
  Rasooli, Sepideh Sadeghi, Erfan~Sadeqi Azer, Niloofar~Safi Samghabadi, Mahsa
  Shafaei, Saber Sheybani, Ali Tazarv, and Yadollah Yaghoobzadeh. 2020.
\newblock {ParsiNLU: A Suite of Language Understanding Challenges for Persian}.
\newblock \emph{arXiv preprint arXiv:2012.06154}.

\bibitem[{Lauscher et~al.(2020)Lauscher, Ravishankar, Vuli{\'{c}}, and
  Glava{\v{s}}}]{Lauscher2020fromzerotohero}
Anne Lauscher, Vinit Ravishankar, Ivan Vuli{\'{c}}, and Goran Glava{\v{s}}.
  2020.
\newblock {From Zero to Hero: On the Limitations of Zero-Shot Cross-Lingual
  Transfer with Multilingual Transformers}.
\newblock In \emph{Proceedings of EMNLP 2020}.

\bibitem[{Lewis et~al.(2020)Lewis, Oguz, Rinott, Riedel, and
  Schwenk}]{Lewis2019mlqa}
Patrick Lewis, Barlas Oguz, Ruty Rinott, Sebastian Riedel, and Holger Schwenk.
  2020.
\newblock \href {https://doi.org/10.18653/v1/2020.acl-main.653} {{MLQA}:
  Evaluating cross-lingual extractive question answering}.
\newblock In \emph{Proceedings of the 58th Annual Meeting of the Association
  for Computational Linguistics}, pages 7315--7330, Online. Association for
  Computational Linguistics.

\bibitem[{Liang et~al.(2020)Liang, Duan, Gong, Wu, Guo, Qi, Gong, Shou, Jiang,
  Cao, Fan, Zhang, Agrawal, Cui, Wei, Bharti, Qiao, Chen, Wu, Liu, Yang,
  Campos, Majumder, and Zhou}]{liang2020xglue}
Yaobo Liang, Nan Duan, Yeyun Gong, Ning Wu, Fenfei Guo, Weizhen Qi, Ming Gong,
  Linjun Shou, Daxin Jiang, Guihong Cao, Xiaodong Fan, Ruofei Zhang, Rahul
  Agrawal, Edward Cui, Sining Wei, Taroon Bharti, Ying Qiao, Jiun-Hung Chen,
  Winnie Wu, Shuguang Liu, Fan Yang, Daniel Campos, Rangan Majumder, and Ming
  Zhou. 2020.
\newblock \href {https://doi.org/10.18653/v1/2020.emnlp-main.484} {{XGLUE}: A
  new benchmark datasetfor cross-lingual pre-training, understanding and
  generation}.
\newblock In \emph{Proceedings of the 2020 Conference on Empirical Methods in
  Natural Language Processing (EMNLP)}, pages 6008--6018, Online. Association
  for Computational Linguistics.

\bibitem[{Linzen(2020)}]{linzen2020can}
Tal Linzen. 2020.
\newblock \href {https://doi.org/10.18653/v1/2020.acl-main.465} {How can we
  accelerate progress towards human-like linguistic generalization?}
\newblock In \emph{Proceedings of the 58th Annual Meeting of the Association
  for Computational Linguistics}, pages 5210--5217, Online. Association for
  Computational Linguistics.

\bibitem[{Littell et~al.(2017)Littell, Mortensen, Lin, Kairis, Turner, and
  Levin}]{littell2017uriel}
Patrick Littell, David~R Mortensen, Ke~Lin, Katherine Kairis, Carlisle Turner,
  and Lori Levin. 2017.
\newblock Uriel and lang2vec: Representing languages as typological,
  geographical, and phylogenetic vectors.
\newblock In \emph{Proceedings of the 15th Conference of the European Chapter
  of the Association for Computational Linguistics: Volume 2, Short Papers},
  pages 8--14.

\bibitem[{Liu et~al.(2021)Liu, Fu, Xiao, Yuan, Chang, Dai, Liu, Ye, and
  Neubig}]{liu2021explainaboard}
Pengfei Liu, Jinlan Fu, Yang Xiao, Weizhe Yuan, Shuaichen Chang, Junqi Dai,
  Yixin Liu, Zihuiwen Ye, and Graham Neubig. 2021.
\newblock \href {https://doi.org/10.18653/v1/2021.acl-demo.34}
  {{E}xplaina{B}oard: An explainable leaderboard for {NLP}}.
\newblock In \emph{Proceedings of the 59th Annual Meeting of the Association
  for Computational Linguistics and the 11th International Joint Conference on
  Natural Language Processing: System Demonstrations}, pages 280--289, Online.
  Association for Computational Linguistics.

\bibitem[{Liu et~al.(2019)Liu, Ott, Goyal, Du, Joshi, Chen, Levy, Lewis,
  Zettlemoyer, and Stoyanov}]{liu2019roberta}
Yinhan Liu, Myle Ott, Naman Goyal, Jingfei Du, Mandar Joshi, Danqi Chen, Omer
  Levy, Mike Lewis, Luke Zettlemoyer, and Veselin Stoyanov. 2019.
\newblock Roberta: A robustly optimized bert pretraining approach.
\newblock \emph{arXiv preprint arXiv:1907.11692}.

\bibitem[{Luo et~al.(2020)Luo, Wang, Liu, Liu, Bi, Huang, Huang, and
  Si}]{luo2020veco}
Fuli Luo, Wei Wang, Jiahao Liu, Yijia Liu, Bin Bi, Songfang Huang, Fei Huang,
  and Luo Si. 2020.
\newblock Veco: Variable encoder-decoder pre-training for cross-lingual
  understanding and generation.
\newblock \emph{arXiv preprint arXiv:2010.16046}.

\bibitem[{Nivre et~al.(2018)Nivre, Abrams, Agi{\'c}, Ahrenberg, Antonsen,
  Aranzabe, Arutie, Asahara, Ateyah, Attia et~al.}]{nivre2018universal}
Joakim Nivre, Mitchell Abrams, {\v{Z}}eljko Agi{\'c}, Lars Ahrenberg, Lene
  Antonsen, Maria~Jesus Aranzabe, Gashaw Arutie, Masayuki Asahara, Luma Ateyah,
  Mohammed Attia, et~al. 2018.
\newblock Universal dependencies 2.2.

\bibitem[{Ouyang et~al.(2020)Ouyang, Wang, Pang, Sun, Tian, Wu, and
  Wang}]{Ouyang2020erniem}
Xuan Ouyang, Shuohuan Wang, Chao Pang, Yu~Sun, Hao Tian, Hua Wu, and Haifeng
  Wang. 2020.
\newblock {ERNIE-M: Enhanced Multilingual Representation by Aligning
  Cross-lingual Semantics with Monolingual Corpora}.
\newblock \emph{arXiv preprint arXiv:2012.15674}.

\bibitem[{Pan et~al.(2017)Pan, Zhang, May, Nothman, Knight, and Ji}]{Pan2017}
Xiaoman Pan, Boliang Zhang, Jonathan May, Joel Nothman, Kevin Knight, and Heng
  Ji. 2017.
\newblock {Cross-lingual name tagging and linking for 282 languages}.
\newblock In \emph{Proceedings of ACL 2017}, pages 1946--1958.

\bibitem[{Papineni et~al.(2002)Papineni, Roukos, Ward, and
  Zhu}]{papineni-etal-2002-bleu}
Kishore Papineni, Salim Roukos, Todd Ward, and Wei-Jing Zhu. 2002.
\newblock \href {https://doi.org/10.3115/1073083.1073135} {{B}leu: a method for
  automatic evaluation of machine translation}.
\newblock In \emph{Proceedings of the 40th Annual Meeting of the Association
  for Computational Linguistics}, pages 311--318, Philadelphia, Pennsylvania,
  USA. Association for Computational Linguistics.

\bibitem[{Pfeiffer et~al.(2020)Pfeiffer, Vuli{\'{c}}, Gurevych, and
  Ruder}]{Pfeiffer2020mad-x}
Jonas Pfeiffer, Ivan Vuli{\'{c}}, Iryna Gurevych, and Sebastian Ruder. 2020.
\newblock {MAD-X: An Adapter-based Framework for Multi-task Cross-lingual
  Transfer}.
\newblock In \emph{Proceedings of EMNLP 2020}.

\bibitem[{Phang et~al.(2020)Phang, Calixto, Htut, Pruksachatkun, Liu, Vania,
  Kann, and Bowman}]{phang2020english}
Jason Phang, Iacer Calixto, Phu~Mon Htut, Yada Pruksachatkun, Haokun Liu, Clara
  Vania, Katharina Kann, and Samuel~R. Bowman. 2020.
\newblock \href {https://aclanthology.org/2020.aacl-main.56} {{E}nglish
  intermediate-task training improves zero-shot cross-lingual transfer too}.
\newblock In \emph{Proceedings of the 1st Conference of the Asia-Pacific
  Chapter of the Association for Computational Linguistics and the 10th
  International Joint Conference on Natural Language Processing}, pages
  557--575, Suzhou, China. Association for Computational Linguistics.

\bibitem[{Ponti et~al.(2020)Ponti, Glava{\v{s}}, Majewska, Liu, Vuli{\'c}, and
  Korhonen}]{Ponti2020xcopa}
Edoardo~Maria Ponti, Goran Glava{\v{s}}, Olga Majewska, Qianchu Liu, Ivan
  Vuli{\'c}, and Anna Korhonen. 2020.
\newblock \href {https://doi.org/10.18653/v1/2020.emnlp-main.185} {{XCOPA}: A
  multilingual dataset for causal commonsense reasoning}.
\newblock In \emph{Proceedings of the 2020 Conference on Empirical Methods in
  Natural Language Processing (EMNLP)}, pages 2362--2376, Online. Association
  for Computational Linguistics.

\bibitem[{Raffel et~al.(2020)Raffel, Shazeer, Roberts, Lee, Narang, Matena,
  Zhou, Li, and Liu}]{Raffel2020t5}
Colin Raffel, Noam Shazeer, Adam Roberts, Katherine Lee, Sharan Narang, Michael
  Matena, Yanqi Zhou, Wei Li, and Peter~J. Liu. 2020.
\newblock {Exploring the Limits of Transfer Learning with a Unified
  Text-to-Text Transformer}.
\newblock \emph{Journal of Machine Learning Research}, 21:1--67.

\bibitem[{Rahimi et~al.(2019)Rahimi, Li, and Cohn}]{Rahimi2019}
Afshin Rahimi, Yuan Li, and Trevor Cohn. 2019.
\newblock {Massively Multilingual Transfer for NER}.
\newblock In \emph{Proceedings of ACL 2019}.

\bibitem[{Rajpurkar et~al.(2016)Rajpurkar, Zhang, Lopyrev, and
  Liang}]{Rajpurkar2016squad}
Pranav Rajpurkar, Jian Zhang, Konstantin Lopyrev, and Percy Liang. 2016.
\newblock {SQuAD: 100,000+ Questions for Machine Comprehension of Text}.
\newblock In \emph{Proceedings of EMNLP 2016}.

\bibitem[{Ribeiro et~al.(2020)Ribeiro, Wu, Guestrin, and
  Singh}]{Ribeiro2020checklist}
Marco~Tulio Ribeiro, Tongshuang Wu, Carlos Guestrin, and Sameer Singh. 2020.
\newblock {Beyond Accuracy: Behavioral Testing of NLP Models with CheckList}.
\newblock In \emph{Proceedings of ACL 2020}, pages 4902--4912.

\bibitem[{Roemmele et~al.(2011)Roemmele, Bejan, and
  Gordon}]{roemmele2011choice}
Melissa Roemmele, Cosmin~Adrian Bejan, and Andrew~S Gordon. 2011.
\newblock Choice of plausible alternatives: An evaluation of commonsense causal
  reasoning.
\newblock In \emph{AAAI spring symposium: logical formalizations of commonsense
  reasoning}, pages 90--95.

\bibitem[{Roy et~al.(2020)Roy, Constant, Al-Rfou, Barua, Phillips, and
  Yang}]{roy-etal-2020-lareqa}
Uma Roy, Noah Constant, Rami Al-Rfou, Aditya Barua, Aaron Phillips, and Yinfei
  Yang. 2020.
\newblock \href {https://doi.org/10.18653/v1/2020.emnlp-main.477} {{LAR}e{QA}:
  Language-agnostic answer retrieval from a multilingual pool}.
\newblock In \emph{Proceedings of the 2020 Conference on Empirical Methods in
  Natural Language Processing (EMNLP)}, pages 5919--5930, Online. Association
  for Computational Linguistics.

\bibitem[{Sap et~al.(2019)Sap, Rashkin, Chen, Bras, Choi, Intelligence, and
  Science}]{Sap2019socialiqa}
Maarten Sap, Hannah Rashkin, Derek Chen, Ronan~Le Bras, Yejin Choi, Artificial
  Intelligence, and Computer Science. 2019.
\newblock {Social IQa: Commonsense Reasoning about Social Interactions Maarten
  Sap}.
\newblock In \emph{Proceedings of EMNLP 2019}, pages 4453--4463.

\bibitem[{S{\o}gaard et~al.(2018)S{\o}gaard, Ruder, and
  Vuli{\'{c}}}]{Sogaard2018unsupervised}
Anders S{\o}gaard, Sebastian Ruder, and Ivan Vuli{\'{c}}. 2018.
\newblock \href {http://arxiv.org/abs/arXiv:1805.03620v1} {{On the Limitations
  of Unsupervised Bilingual Dictionary Induction}}.
\newblock In \emph{Proceedings of ACL 2018}.

\bibitem[{Strubell et~al.(2019)Strubell, Ganesh, and
  McCallum}]{Strubell2019energy}
Emma Strubell, Ananya Ganesh, and Andrew McCallum. 2019.
\newblock \href {http://arxiv.org/abs/1906.02243} {{Energy and Policy
  Considerations for Deep Learning in NLP}}.
\newblock In \emph{Proceedings of ACL 2019}.

\bibitem[{Vaswani et~al.(2017)Vaswani, Shazeer, Parmar, Uszkoreit, Jones,
  Gomez, Kaiser, and Polosukhin}]{Vaswani2017}
Ashish Vaswani, Noam Shazeer, Niki Parmar, Jakob Uszkoreit, Llion Jones,
  Aidan~N. Gomez, {\L}ukasz Kaiser, and Illia Polosukhin. 2017.
\newblock {Attention Is All You Need}.
\newblock In \emph{Proceedings of NIPS 2017}.

\bibitem[{Vrande{\v{c}}i{\'c} and Kr{\"o}tzsch(2014)}]{vrandevcic2014wikidata}
Denny Vrande{\v{c}}i{\'c} and Markus Kr{\"o}tzsch. 2014.
\newblock Wikidata: a free collaborative knowledgebase.
\newblock \emph{Communications of the ACM}, 57(10):78--85.

\bibitem[{Wang et~al.(2019{\natexlab{a}})Wang, Michael, Hill, Levy, and
  Bowman}]{Wang2019superglue}
Alex Wang, Julian Michael, Felix Hill, Omer Levy, and Samuel~R Bowman.
  2019{\natexlab{a}}.
\newblock {SuperGLUE: A Stickier Benchmark for General-Purpose Language
  Understanding Systems}.
\newblock In \emph{Proceedings of NeurIPS 2019}.

\bibitem[{Wang et~al.(2019{\natexlab{b}})Wang, Singh, Michael, Hill, Levy, and
  Bowman}]{Wang2019glue}
Alex Wang, Amanpreet Singh, Julian Michael, Felix Hill, Omer Levy, and Samuel~R
  Bowman. 2019{\natexlab{b}}.
\newblock {GLUE: A Multi-Task Benchmark and Analysis Platform for Natural
  Language Understanding}.
\newblock In \emph{Proceedings of ICLR 2019}.

\bibitem[{Wei et~al.(2021)Wei, Hu, Weng, Xing, Yu, and Luo}]{Wei2021hictl}
Xiangpeng Wei, Yue Hu, Rongxiang Weng, Luxi Xing, Heng Yu, and Weihua Luo.
  2021.
\newblock {On Learning Universal Representations Across Languages}.
\newblock In \emph{Proceedings of ICLR 2021}.

\bibitem[{Williams et~al.(2018)Williams, Nangia, and
  Bowman}]{Williams2018multinli}
Adina Williams, Nikita Nangia, and Samuel~R. Bowman. 2018.
\newblock {A Broad-Coverage Challenge Corpus for Sentence Understanding through
  Inference}.
\newblock In \emph{Proceedings of NAACL-HLT 2018}.

\bibitem[{Willie et~al.(2020)Willie, Vincentio, Winata, Cahyawijaya, Li, Lim,
  Soleman, Mahendra, Fung, Bahar, and Purwarianti}]{Willie2020indonlu}
Bryan Willie, Karissa Vincentio, Genta~Indra Winata, Samuel Cahyawijaya,
  Xiaohong Li, Zhi~Yuan Lim, Sidik Soleman, Rahmad Mahendra, Pascale Fung,
  Syafri Bahar, and Ayu Purwarianti. 2020.
\newblock {IndoNLU: Benchmark and Resources for Evaluating Indonesian Natural
  Language Understanding}.
\newblock In \emph{Proceedings of AACL-IJCNLP 2020}.

\bibitem[{Wolf et~al.(2019)Wolf, Debut, Sanh, Chaumond, Delangue, Moi, Cistac,
  Rault, Louf, Funtowicz et~al.}]{wolf2019huggingface}
Thomas Wolf, L~Debut, V~Sanh, J~Chaumond, C~Delangue, A~Moi, P~Cistac, T~Rault,
  R~Louf, M~Funtowicz, et~al. 2019.
\newblock Huggingface's transformers: State-of-the-art natural language
  processing.
\newblock \emph{arXiv preprint arXiv:1910.03771}.

\bibitem[{Xue et~al.(2021)Xue, Constant, Roberts, Kale, Al-Rfou, Siddhant,
  Barua, and Raffel}]{Xue2020mt5}
Linting Xue, Noah Constant, Adam Roberts, Mihir Kale, Rami Al-Rfou, Aditya
  Siddhant, Aditya Barua, and Colin Raffel. 2021.
\newblock \href {https://doi.org/10.18653/v1/2021.naacl-main.41} {m{T}5: A
  massively multilingual pre-trained text-to-text transformer}.
\newblock In \emph{Proceedings of the 2021 Conference of the North American
  Chapter of the Association for Computational Linguistics: Human Language
  Technologies}, pages 483--498, Online. Association for Computational
  Linguistics.

\bibitem[{Zhao et~al.(2020)Zhao, Mukherjee, Hosseini, Chang, {Hassan
  Awadallah}, Hassan, Associates, Zhao, Mukherjee, Hosseini, Awadallah,
  Najafabadipour, Tu{\~{n}}as, Rodr{\'{i}}guez-Gonz{\'{a}}lez, and
  Menasalvas}]{Zhao2020bias}
Jieyu Zhao, Subhabrata~(Subho) Mukherjee, Saghar Hosseini, Kai-Wei Chang, Ahmed
  {Hassan Awadallah}, Ahmed Hassan, Rdcp Associates, Jieyu Zhao,
  Subhabrata~(Subho) Mukherjee, Saghar Hosseini, Ahmed Awadallah,
  M~Najafabadipour, J~M Tu{\~{n}}as, A~Rodr{\'{i}}guez-Gonz{\'{a}}lez, and
  E~Menasalvas. 2020.
\newblock \href {https://doi.org/10.1007/978-3-030-06016-9_15} {{Gender Bias in
  Multilingual Embeddings and Cross-lingual Transfer}}.
\newblock In \emph{Proceedings of ACL 2020}.

\end{thebibliography}
\bibliographystyle{acl_natbib}

\clearpage

\appendix

\section*{Appendix}

\section{Details of \xtreme models} \label{app:xtreme-model-details}

All submissions to the \xtreme leaderboard are large-scale Transformers \citep{Vaswani2017} trained with masked language modeling (MLM), many of which extend monolingual models. Multilingual BERT (mBERT), XLM-RoBERTa \citep[XLM-R;][]{Conneau2020xlmr} and multilingual T5 \citep[mT5;][]{Xue2020mt5} extend BERT \citep{Devlin2019bert}, RoBERTa \citep{liu2019roberta}, and T5 \citep{Raffel2020t5} respectively. Rebalanced mBERT \citep[RemBERT;][]{Chung2021rembert} is a more efficient, scaled-up reparameterization of mBERT.
% Multilingual BERT (mBERT) is a 12-layer Transformer model that extends BERT \citep{Devlin2019bert}. XLM-RoBERTa \citep[XLM-R;][]{Conneau2020xlmr} Large is a 24-layer Transformer model that similarly extends RoBERTa \citep{liu2019roberta}. Multilingual T5 \citep[mT5;][]{Xue2020mt5} extends the T5 model \citep{Raffel2020t5}. Rebalanced mBERT \citep[RemBERT;][]{Chung2021rembert} is a scaled up version of mBERT trained with a more efficient parameter allocation.
These models have been pre-trained on unlabeled data in around 100 languages from Wikipedia (mBERT) and CommonCrawl. XLM-R was the strongest baseline in \xtreme \cite{Hu2020xtreme} and is the foundation for some subsequent work. It has been fine-tuned on English data of a related task prior to task-specific fine-tuning \citep[STILTs;][]{phang2020english}.
% Supplementary Training on Intermediate Labeled-data Tasks \citep[STILTs;][]{phang2020english} has been used to additionally fine-tune a model on English data of a related task prior to task-specific fine-tuning. 
The following models furthermore propose new methods to leverage parallel data during pre-training or fine-tuning. FILTER \citep{Fang2020filter}, based on XLM-R, fuses representations in different languages. VECO \cite{luo2020veco} is a 24-layer encoder-decoder model that uses additional MLM variants during pre-training. \mbox{T-URLv2} and \mbox{HiCTL} \cite{Wei2021hictl}, based on InfoXLM \cite{Chi2020infoxml} and XLM-R respectively, employ contrastive losses. \mbox{ERNIE-M} \cite{Ouyang2020erniem} incorporates back-translation into language modeling.\footnote{We are not aware of the technical details of Polyglot and the anonymous submission.}

\section{Task scores on \xtreme} \label{app:xtreme_scores_language_family}

\begin{figure*}[!h]
\centering
\begin{subfigure}{0.5\textwidth}
  \centering
  \includegraphics[width=\linewidth]{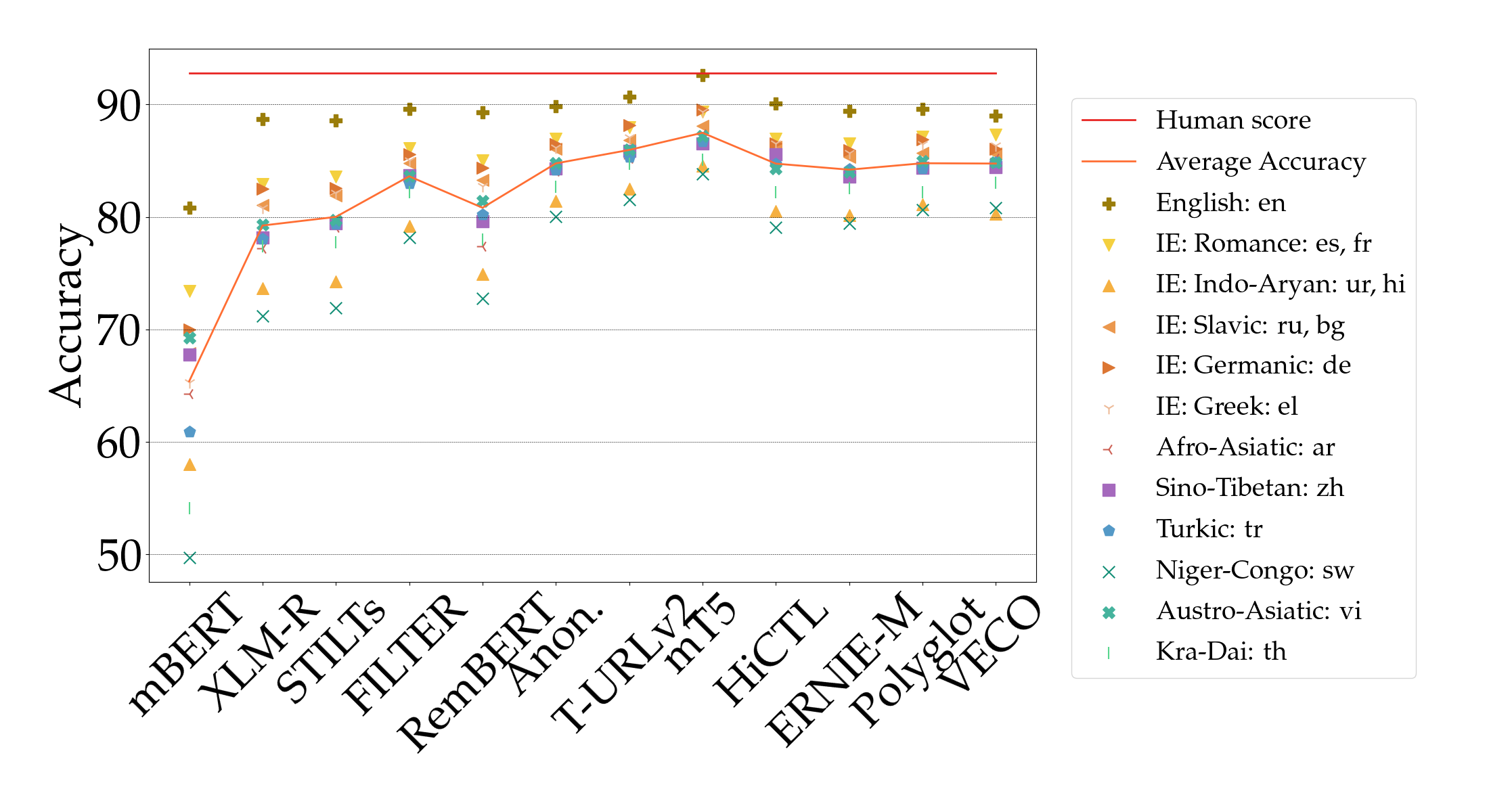}
  \caption{Performance on XNLI}
\end{subfigure}%
\begin{subfigure}{.5\textwidth}
  \centering
  \includegraphics[width=\linewidth]{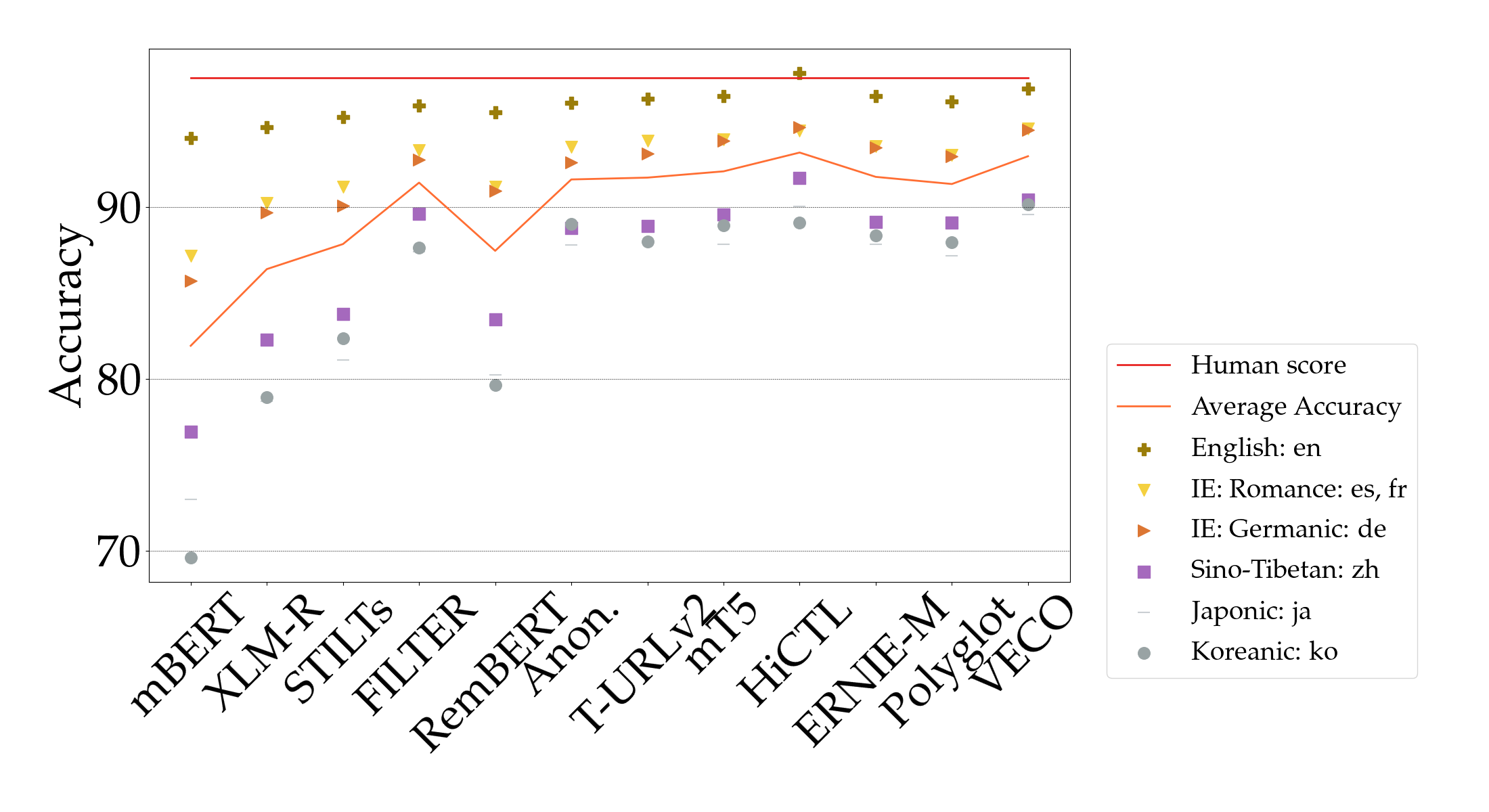}
  \caption{Performance on PAWS-X}
\end{subfigure}
\begin{subfigure}{.5\textwidth}
  \centering
  \includegraphics[width=\linewidth]{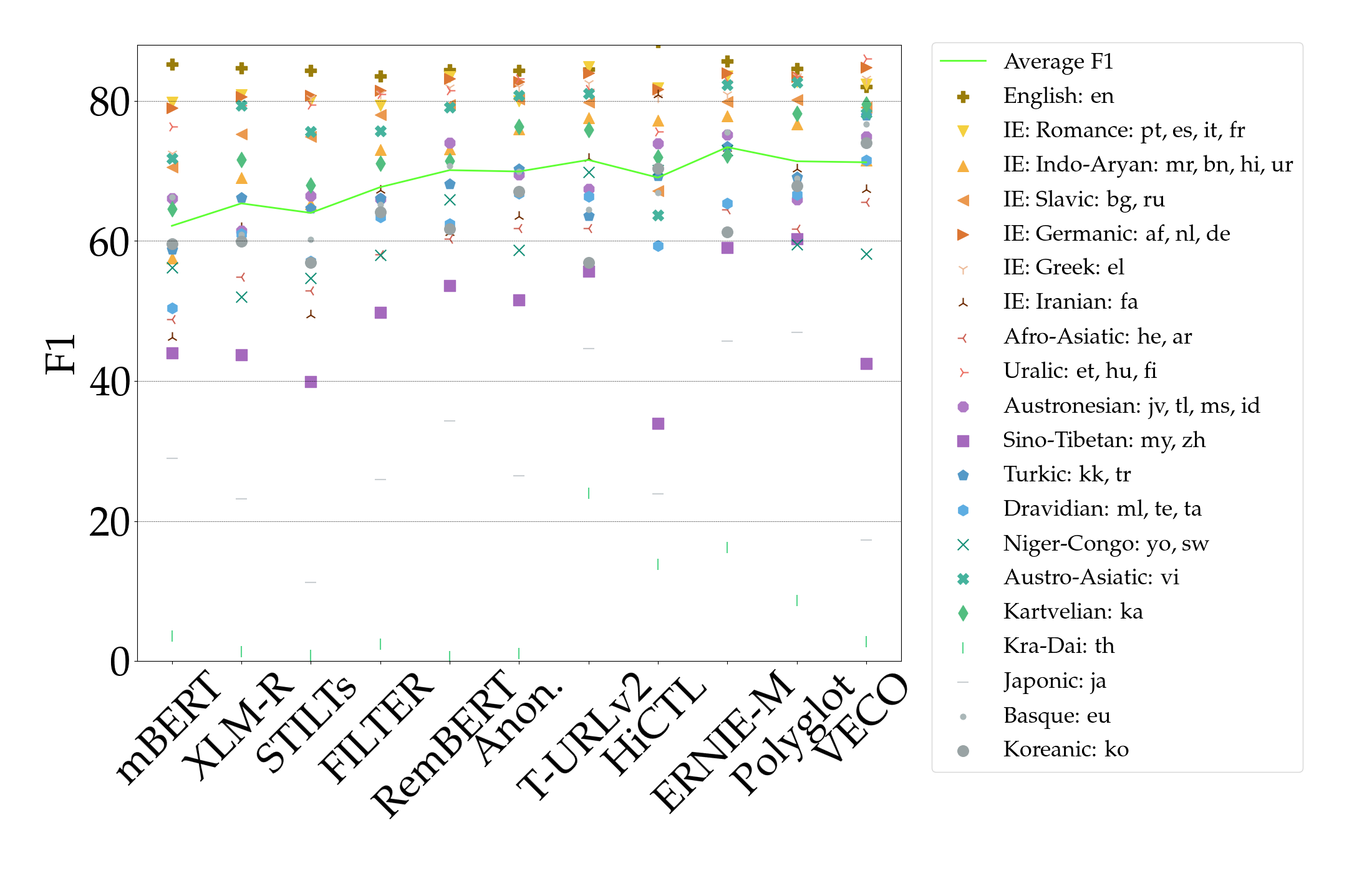}
  \caption{Performance on WikiANN-NER}
\end{subfigure}%
\begin{subfigure}{.5\textwidth}
  \centering
  \includegraphics[width=\linewidth]{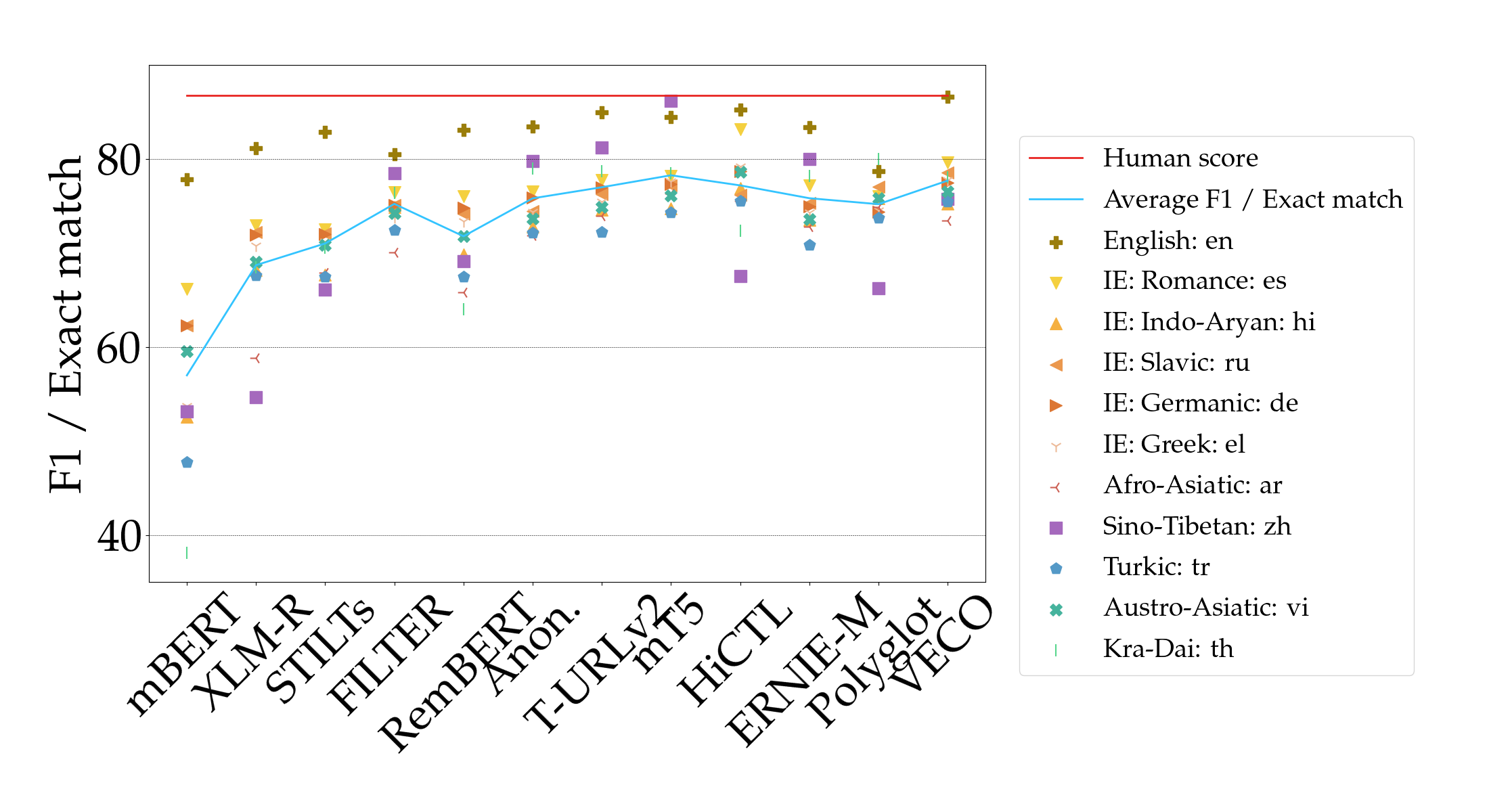}
  \caption{Performance on XQuAD}
\end{subfigure}
\begin{subfigure}{.5\textwidth}
  \centering
  \includegraphics[width=\linewidth]{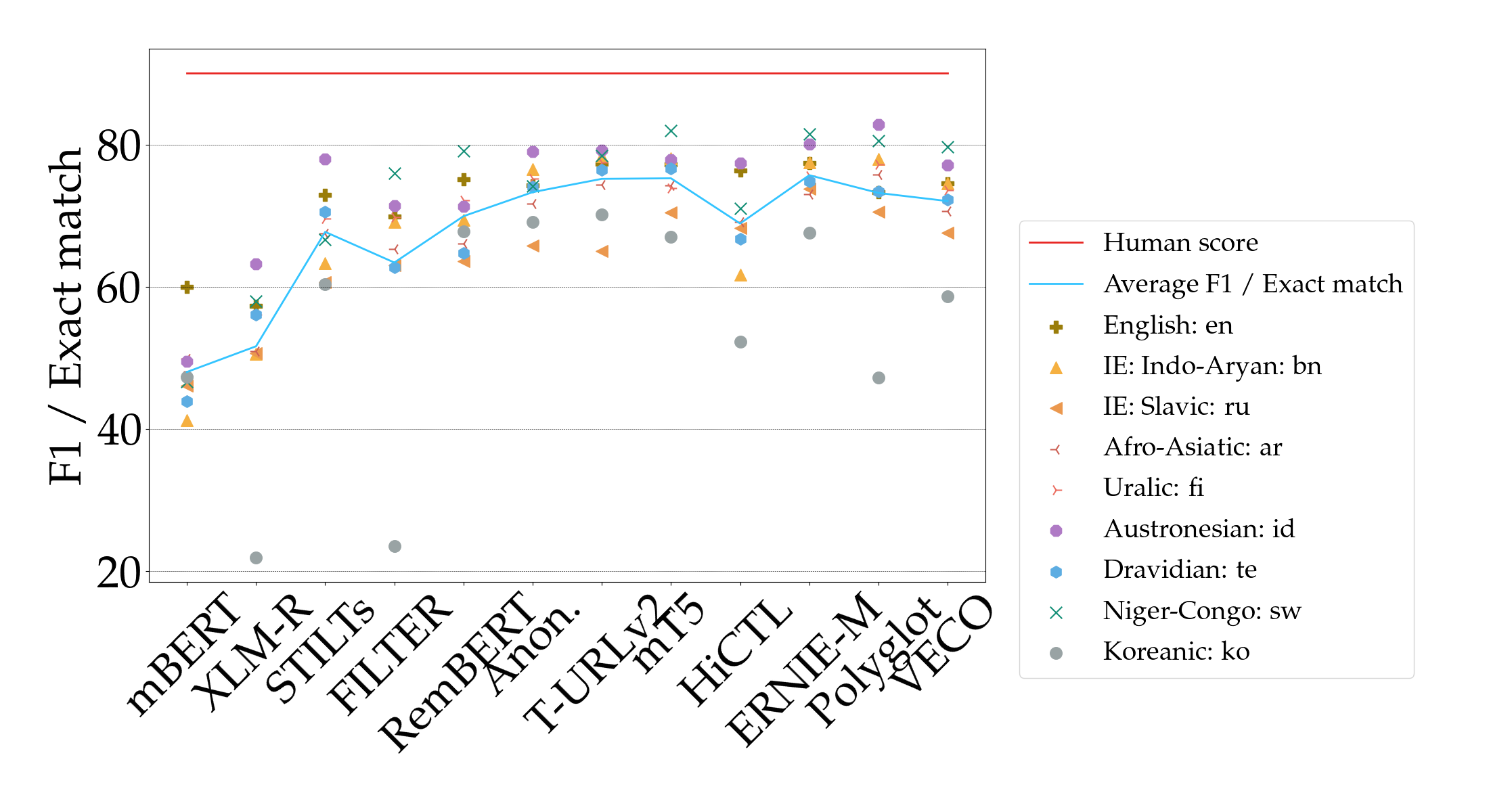}
  \caption{Performance on TyDiQA-GoldP}
\end{subfigure}%
\begin{subfigure}{.5\textwidth}
  \centering
  \includegraphics[width=\linewidth]{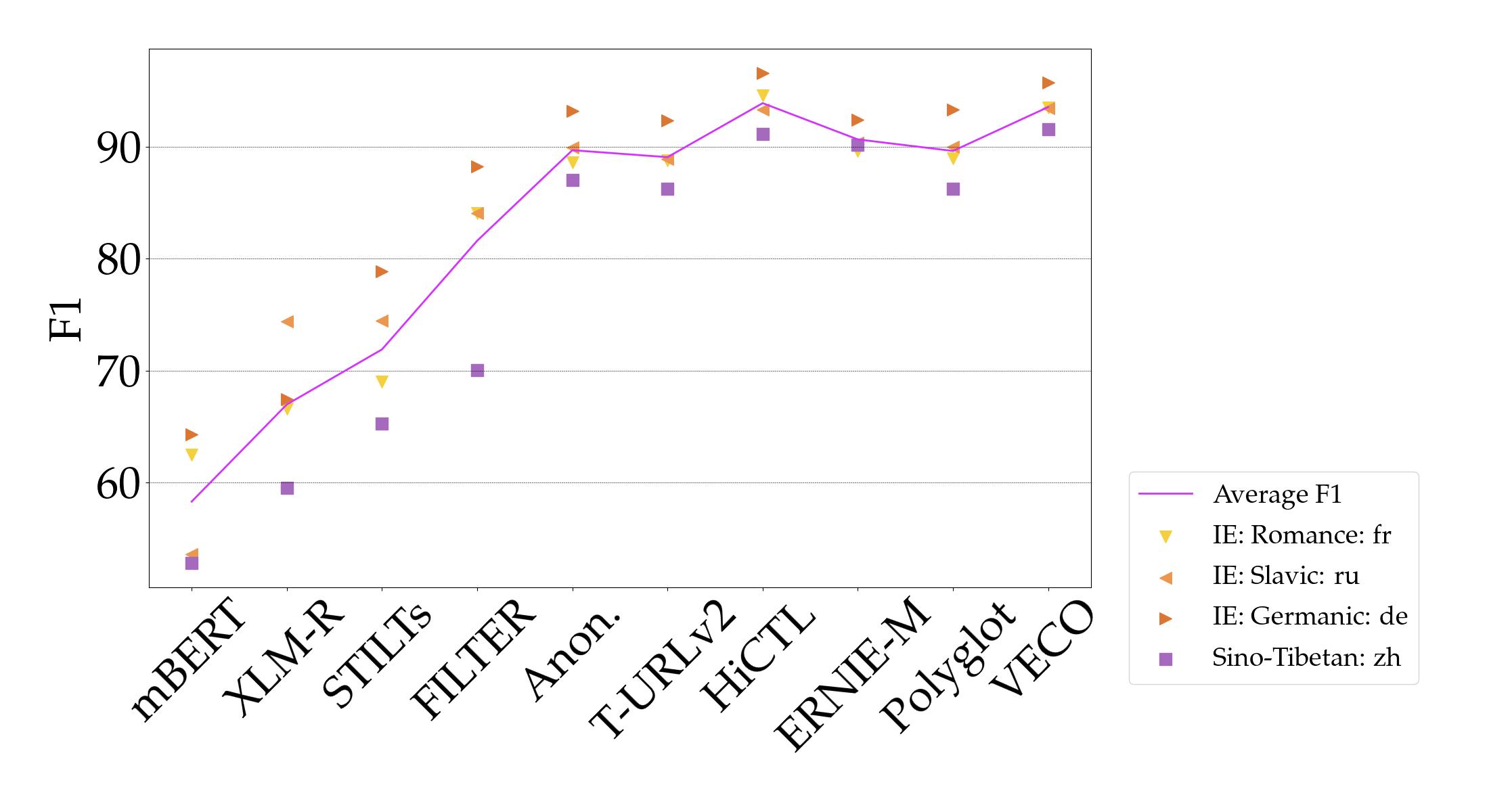}
  \caption{Performance on BUCC}
\end{subfigure}
% \vspace{-0.8cm}
\caption{Performance of all models on the \xtreme leaderboard on (a) XNLI, (b) PAWS-X, (c) WikiANN-NER, (d) XQuAD, (e) TyDiQA-GoldP, and (f) BUCC across language families.}
\label{fig:xtreme_leaderboard_analysis_appendix}
\end{figure*}

We show the performance of the models on the \xtreme leaderboard broken down by language family on the remaining \xtreme tasks in Figure \ref{fig:xtreme_leaderboard_analysis_appendix}.

\section{\xtreme tasks retained in \name} \label{app:retained-tasks}

% \mj{If we're short on space, we can cut these descriptions to two lines each.}
\noindent \textbf{XNLI} $\:$ The Cross-lingual Natural Language Inference corpus \citep{Conneau2018xnli} requires a model to determine whether a premise sentence entails, contradicts, or is neutral with respect to a hypothesis sentence. We use the crowd-sourced English data that was professionally translated to 14 other languages for evaluation and the MultiNLI \citep{Williams2018multinli} train set for training. 
% While average accuracy on XNLI has recently exceeded 80\%, there is a substantial gap to human performance. 
% In addition, automatically translated data enables analyzing performance in 40 languages \citep{Hu2020xtreme}.

\noindent \textbf{UD-POS} $\:$ We employ the part-of-speech tagging data from the Universal Dependencies v2.7 treebanks \citep{nivre2018universal} covering 104 languages. 
% Each word is assigned one of 17 universal POS tags. 
We use the English training data for training and evaluate on the test sets of the target languages. 
% While monolingual models generally reach high performance on this task, cross-lingual transfer of syntactic information is still challenging for current methods, as indicated by the limited progress of recent methods.

\noindent \textbf{WikiANN-NER} $\:$ For named entity recognition, we use the WikiANN dataset \citep{Pan2017}, in which proper noun spans in Wikipedia text have been automatically annotated as either \textit{location}, \textit{person}, or \textit{organization}. 
% using a combination of knowledge base properties, cross-lingual and anchor links, self-training, and data selection. 
We use the balanced train, dev, and test splits from \citet{Rahimi2019}. 
% Similar to POS tagging, cross-lingual progress on this task has been slow.

\noindent \textbf{XQuAD} $\:$ The Cross-lingual Question Answering Dataset \citep{artetxe2020cross} requires identifying the answer to a question as a span in the corresponding paragraph. A subset of the English SQuAD v1.1 \citep{Rajpurkar2016squad} dev set was professionally translated into ten other languages for XQuAD.

\noindent \textbf{MLQA} $\:$ Similarly to XQuAD, the Multilingual Question Answering dataset \citep{Lewis2019mlqa} is another cross-lingual question answering task. The evaluation data in seven languages was automatically mined from Wikipedia, annotations were crowd-sourced, and answer spans aligned. 
% The evaluation data for English and six other languages was obtained by automatically mining parallel English and target language sentences from Wikipedia, crowd-sourcing annotations in English, translating the question, and aligning the answer spans in the target languages. 
For both XQuAD and MLQA, we use their respective data for evaluation and train on SQuAD v1.1.
% For both XQuAD and MLQA, we use SQuAD v1.1 for training and evaluate on the test data of the corresponding task. 

\noindent \textbf{TyDiQA-GoldP} $\:$ We use the gold passage (GoldP) version of TyDiQA \citep{Clark2020tydiqa}, a benchmark for information-seeking question answering, which covers nine typologically diverse languages. The GoldP version is a simplification of the primary task, using only the gold passage as context and excluding unanswerable questions. We use the English training data for training and evaluate on the test sets of the target languages. 
% TyDiQA-GoldP overall is more challenging than XQuAD and MLQA as questions have been written without seeing the answers and so have less lexical overlap. 
% While performance on the QA tasks has improved recently, there is still a substantial gap to human-level performance.

\noindent \textbf{Tatoeba} $\:$ We evaluate on the Tatoeba dataset \citep{Artetxe2019massively}, which consists of up to 1,000 English-aligned sentence pairs covering 122 languages. We find the nearest neighbor using cosine similarity.
% Despite strong recent progress on Tatoeba, performance lags behind in a number of languages. 
To make the setting more realistic, we move away from zero-shot retrieval and fine-tune models on SQuAD v1.1.

\section{Languages} \label{app:languages}

\noindent \textbf{Language characteristics} $\:$ We show a detailed overview of languages in \name including interesting typological differences in Table \ref{tab:languages}. Wikipedia information is taken from Wikipedia\footnote{\url{https://meta.wikimedia.org/wiki/List_of_Wikipedias}} and linguistic information from WALS Online\footnote{\url{https://wals.info/languoid}}. \name includes members of the Afro-Asiatic, Austro-Asiatic, Austronesian, Dravidian, Indo-European, Japonic, Kartvelian, Kra-Dai, Niger-Congo, Sino-Tibetan, Turkic, Uralic, Creole\footnote{For simplicity, we treat Creole as a distinct language family \cite{bakker2011creoles}.}, and Quechuan language families as well as of two isolates, Basque and Korean.

\begin{table*}[]
\centering
\resizebox{\textwidth}{!}{%
\begin{tabular}{lccllcccccc}
\toprule
Language & \begin{tabular}[c]{@{}l@{}}ISO\\ 639-1\\ code\end{tabular} & \begin{tabular}[c]{@{}l@{}}\# Wikipedia\\ articles (in\\ millions)\end{tabular} & Script & \begin{tabular}[c]{@{}l@{}}Language\\ family\end{tabular} & \begin{tabular}[c]{@{}l@{}}Diacritics /\\ special\\ characters\end{tabular} & \begin{tabular}[c]{@{}l@{}}Extensive\\ compound-\\ ing\end{tabular} & \begin{tabular}[c]{@{}l@{}}Bound\\ words /\\ clitics\end{tabular} & \begin{tabular}[c]{@{}l@{}}Inflec-\\ tion\end{tabular} & \begin{tabular}[c]{@{}l@{}}Deriva-\\ tion\end{tabular} & \begin{tabular}[c]{@{}l@{}}\# datasets\\ with\\ language\end{tabular} \\ \midrule
Afrikaans & af & 0.09 & Latin & IE: Germanic &  & \cmark &  &  &  & 3 \\
Arabic & ar & 1.02 & Arabic & Afro-Asiatic & \cmark &  & \cmark & \cmark &  & 9 \\
Azerbaijani* & az & 0.18 & Latin & Turkic &  &  &  &  & \cmark & 2 \\
Bulgarian & bg & 0.26 & Cyrillic & IE: Slavic & \cmark &  & \cmark & \cmark &  & 4 \\
Bengali & bn & 0.08 & Brahmic & IE: Indo-Aryan & \cmark & \cmark & \cmark & \cmark & \cmark & 3 \\
German & de & 2.37 & Latin & IE: Germanic &  & \cmark &  & \cmark &  & 8 \\
Greek & el & 0.17 & Greek & IE: Greek & \cmark & \cmark &  & \cmark &  & 6 \\
English & en & 5.98 & Latin & IE: Germanic &  &  &  &  &  & 9 \\
Spanish & es & 1.56 & Latin & IE: Romance & \cmark &  & \cmark &  &  & 8 \\
Estonian & et & 0.2 & Latin & Uralic & \cmark & \cmark &  & \cmark & \cmark & 4 \\
Basque & eu & 0.34 & Latin & Basque & \cmark &  & \cmark & \cmark & \cmark & 3 \\
Persian & fa & 0.7 & Perso-Arabic & IE: Iranian &  & \cmark &  &  &  & 3 \\
Finnish & fi & 0.47 & Latin & Uralic &  &  &  & \cmark & \cmark & 4 \\
French & fr & 2.16 & Latin & IE: Romance & \cmark &  & \cmark &  &  & 4 \\
Gujarati* & gu & 0.03 & Brahmic & IE: Indo-Aryan &  &  & \cmark &  &  & 1 \\
Hebrew & he & 0.25 & Jewish & Afro-Asiatic &  &  &  & \cmark &  & 3 \\
Hindi & hi & 0.13 & Devanagari & IE: Indo-Aryan & \cmark & \cmark & \cmark & \cmark & \cmark & 7 \\
Haitian Creole* & ht & 0.06 & Latin & Creole &  &  &  &  &  & 1 \\
Hungarian & hu & 0.46 & Latin & Uralic & \cmark & \cmark &  & \cmark & \cmark & 3 \\
Indonesian & id & 0.51 & Latin & Austronesian &  &  & \cmark & \cmark & \cmark & 5 \\
Italian & it & 1.57 & Latin & IE: Romance & \cmark &  & \cmark &  &  & 4 \\
Japanese & ja & 1.18 & Ideograms & Japonic &  &  & \cmark & \cmark &  & 4 \\
Javanese & jv & 0.06 & Brahmic & Austronesian & \cmark &  & \cmark &  &  & 1 \\
Georgian & ka & 0.13 & Georgian & Kartvelian &  &  &  & \cmark & \cmark & 2 \\
Kazakh & kk & 0.23 & Arabic & Turkic & \cmark &  &  & \cmark & \cmark & 2 \\
Korean & ko & 0.47 & Hangul & Koreanic &  & \cmark &  & \cmark & \cmark & 4 \\
Lithuanian* & lt & 0.2 & Latin & IE: Baltic & \cmark &  &  & \cmark &  & 3 \\
Malayalam & ml & 0.07 & Brahmic & Dravidian & \cmark & \cmark & \cmark & \cmark &  & 2 \\
Marathi & mr & 0.06 & Devanagari & IE: Indo-Aryan &  &  & \cmark & \cmark &  & 3 \\
Malay & ms & 0.33 & Latin & Austronesian &  &  & \cmark & \cmark &  & 2 \\
Burmese & my & 0.05 & Brahmic & Sino-Tibetan & \cmark & \cmark &  &  &  & 1 \\
Dutch & nl & 1.99 & Latin & IE: Germanic &  & \cmark &  &  &  & 3 \\
Punjabi* & pa & 0.04 & Brahmic & IE: Indo-Aryan & \cmark &  &  & \cmark &  & 1 \\
Polish* & pl & 1.44 & Latin & IE: Slavic & \cmark &  &  & \cmark &  & 4 \\
Portuguese & pt & 1.02 & Latin & IE: Romance & \cmark &  & \cmark &  &  & 3 \\
Cusco Quechua* & qu & 0.02 & Latin & Quechuan &  &  &  &  & \cmark & 2 \\
Romanian* & ro & 0.42 & Latin & IE: Romance & \cmark &  & \cmark & \cmark &  & 4 \\
Russian & ru & 1.58 & Cyrillic & IE: Slavic &  &  &  & \cmark &  & 7 \\
Swahili & sw & 0.05 & Latin & Niger-Congo &  &  & \cmark & \cmark & \cmark & 4 \\
Tamil & ta & 0.12 & Brahmic & Dravidian & \cmark & \cmark & \cmark & \cmark & \cmark & 5 \\
Telugu & te & 0.07 & Brahmic & Dravidian & \cmark & \cmark & \cmark & \cmark & \cmark & 4 \\
Thai & th & 0.13 & Brahmic & Kra-Dai & \cmark &  &  &  &  & 7 \\
Tagalog & tl & 0.08 & Brahmic & Austronesian & \cmark &  & \cmark & \cmark &  & 2 \\
Turkish & tr & 0.34 & Latin & Turkic & \cmark & \cmark &  & \cmark & \cmark & 7 \\
Ukrainian* & uk & 1.06 & Cyrillic & IE: Slavic &  &  &  & \cmark &  & 4 \\
Urdu & ur & 0.15 & Perso-Arabic & IE: Indo-Aryan & \cmark & \cmark & \cmark & \cmark & \cmark & 4 \\
Vietnamese & vi & 1.24 & Latin & Austro-Asiatic & \cmark &  &  &  &  & 8 \\
Wolof* & wo & 0.002 & Latin & Niger-Congo & \cmark &  &  &  &  & 1 \\
Yoruba & yo & 0.03 & Arabic & Niger-Congo & \cmark &  &  &  &  & 2 \\
Mandarin & zh & 1.09 & Chinese ideograms & Sino-Tibetan &  & \cmark &  &  &  & 8 \\ \bottomrule
\end{tabular}%
}
\caption{Statistics about languages in \name. Languages belong to 14 language families and two isolates, with Indo-European (IE) having the most members. * indicates newly added languages. Diacritics / special characters: Language adds diacritics (additional symbols to letters). Compounding: Language makes extensive use of word compounds. Bound words / clitics: Function words attach to other words. Inflection: Words are inflected to represent grammatical meaning (e.g.~case marking). Derivation: A single token can represent entire phrases or sentences.}
\label{tab:languages}
\end{table*}

\noindent \textbf{Language diversity indices} $\:$
We measure the language diversity of \name according to the typology and language family indices of \citet{Ponti2020xcopa}, which we show in Table \ref{tab:language-diversity-indices} for \name, \xtreme \citep{Hu2020xtreme}, and \xglue \citep{liang2020xglue}. The typology index is based on the mean entropy of the distribution over 103 typological features from URIEL 
\citep{littell2017uriel} across the languages while the family index consists of the number of distinct language families divided by the total number of languages. \name is similarly diverse while covering a larger number of languages.

\begin{table*}[]
\centering
% \resizebox{\textwidth}{!}{%
\begin{tabular}{l c c c c }
\toprule
% & Range &  XCOPA & TyDiQA & XNLI & XQuaD & MLQA & PAWS-X & \\
%     \cmidrule(lr){3-8}
% Typology & [0, 1] & {0.41} & {0.41} & 0.39 & 0.36 & 0.32 & 0.31 \\
% Family & [0, 1] & {1} & 0.9 & 0.78  & 0.6 & 0.66 & 0.66 \\
% Geography & [0, $\log_2 6$] &  {1.79} & 1.16 & 0.95 & 0.72 & 0.66 & 0 \\
& Range &  \name & \xtreme & \xglue\\
\cmidrule(lr){3-5}
Typology & [0, 1] & 0.42 & 0.43 & 0.35\\
Family & [0, 1] & 0.34 & 0.38 & 0.37\\
% Geography & [0, $\log_2 6$] & \\
\bottomrule
\end{tabular}%
% }
\caption{Diversity indices with regard to typology and language family of \name, \xtreme, and \xglue.}
\label{tab:language-diversity-indices}
\end{table*}

\section{Mewsli-X Dataset} \label{app:mewsli-x}
Mewsli-X is constructed specifically for \name and is 
a more carefully sampled variant of the Mewsli-9 dataset \cite{botha-etal-2020-entity}, derived from WikiNews in the same way.
Compared to Mewsli-9, Serbian is dropped and Polish, Romanian and Ukrainian are added to obtain 11 languages, while the entity descriptions to be retrieved range over all 50 languages in \name (\autoref{app:tab_mewslix_language_matrix}).
To broaden accessibility, the mention queries, candidates and Wikipedia-based training instances are all downsampled compared to the previous work.

\noindent \textbf{Mention Extraction} $\:$
Viable mentions were taken as hyperlinks in the WikiNews articles pointing to Wikipedia entity pages (in any language) that could be mapped successfully to items in our base WikiData entity vocabulary. 
$V_{\mathrm{base}}$ is defined as the set of entities that have a Wikipedia page in any of the 50 languages in \name ($|V_{\mathrm{base}}| \approx 12\mathrm{M}$). The latter condition ensures that entity descriptions are available.
We also extended the WikiData filter used by \citet{botha-etal-2020-entity} to additionally exclude entities that are instances of Wikipedia List Pages (Q13406463) or Wikimedia (Human) Disambiguation Pages (Q22808320, Q4167410), which are not of large interest for entity linking. 

The resolved, viable mentions were filtered to drop duplicate surface mentions of an entity in the same article, and mentions of years (e.g. 2014), which are commonly linked in WikiNews but not of great interest.
We then performed stratified sampling by both mention language and entity frequency bins, seeking uniform sizes across strata. Entity frequency is estimated as the number of times an entity is referenced on pages in the 50-language Wikipedia collection, and then binned into five intervals: $[0,1), [1,10), [10,100),[100,1000),[1000,\infty)$.

The resulting set of 15,000 test mentions covers 9,647 distinct gold entities ($V_{\mathrm{gold}}$).

\noindent \textbf{Candidate Set} $\:$
Doing vector encoding and nearest neighbor search over a full knowledge base of millions of entities is relatively time-consuming, but searching only among $V_{\mathrm{gold}}$ is also unrealistic.
We thus strike a balance by defining a candidate set $V_{\mathrm{cand}}\supset V_{\mathrm{gold}}$, by sampling additional items from $V_{\mathrm{base}} \setminus V_{\mathrm{gold}}$, this time only stratified by entity frequency, to obtain $|V_{\mathrm{cand}}|=1,000,000$.

Each entity $e \in V_{\mathrm{cand}}$ is assigned a single description: we randomly sample a language from among the $L_e \leq 50$ Wikipedia pages corresponding to $e$, and take the page's first sentence as entity description. 

\noindent \textbf{Fine-tuning Data} $\:$
The fine-tuning data constitutes 115K English-only (mention, entity)-pairs, which were sampled from English Wikipedia hyperlinks that map to
$V_{\mathrm{base}}\setminus V_{\mathrm{cand}}$.
Sampling is according to the natural distribution.
The Mewsli-X evaluation setting in \name is thus doubly zero-shot: no candidate or test entities are observed during fine-tuning, nor is non-English text.

% We constructed the candidate pool to contain entities with diverse background frequencies (estimated from the 50 monolingual Wikipedias), and randomly assigned one fixed language description for each entity.
% Unlike Mewsli-9, Mewsli-X was downsampled to meet the lower computational requirements of \name while still expanding significantly over the scale of its retrieval tasks. 
%Following \citep{botha-etal-2020-entity,Wu2019}
% The multilingual Mewsli-X evaluation is doubly zero-shot, both in languages (the fine-tuning set contains English only) and the set of entities (disjoint from those in the fine-tuning set). 

%\input{tables/mewslix_table_lang}
\begin{table*}[ht]
    \centering
    %scale=0.54
    \includegraphics[scale=0.6]{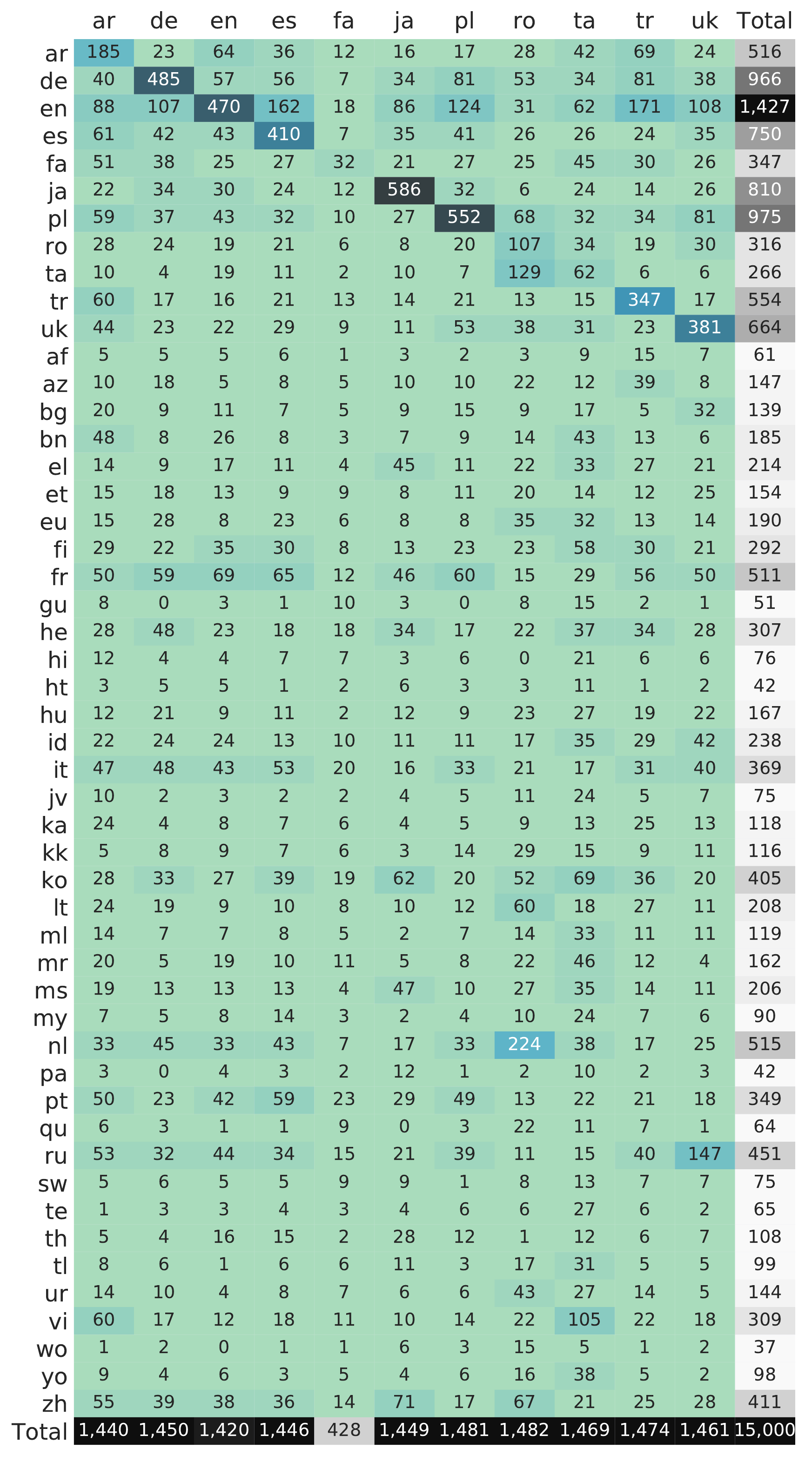}
    \caption{Composition of Mewsli-X test set by mention language (columns) and entity description language (rows).}
    \label{app:tab_mewslix_language_matrix}
\end{table*}

\section{\multichecklist} \label{app:multichecklist}

\noindent \textbf{General statistics} $\:$ Creating \multichecklist involved translating around 550 words (templates and fill-in values) into 49 languages. Some languages required a larger number of words due to the creation of additional templates (Russian required translating 1043 words). In total, the translation effort cost \$4,360. For each test category and each language, we automatically generate 200 test cases. Depending on the number of possible variations for each test, each test case can consist of 2 (Comparisons) to 12 (Intensifiers) examples.\footnote{Note that the number of variations does not correlate with the test success rate, e.g. the ``Job vs nationality'' test has 8 variations for each test case and the highest success rate overall.}

\noindent \textbf{Instructions to translators} $\:$ We sent the following guidelines to annotators for the translation:
\begin{enumerate}
    \itemsep0em 
    \item We would like you to translate the following templates and their corresponding fill-in values into other languages. Each template contains some text enclosed in curly brackets \{ \}. These are the names of the fields that will be substituted in the template. We ask you not to translate the text within such curly brackets.
    \item Have a look at the text in the other fields to get a better sense what can be substituted in the template. For instance, by referring to the lines with ``adj'', we can see that ``\{first\_name\} is \{adj[0]\} than \{first\_name1\}'' is a comparison between two people.
    \item If there is not a literal translation or the same translation has already been used for another word, feel free to use a translation that is similar in meaning.
    \item If the translation of the template differs based on the substituted words, please create multiple translations for the template and indicate which substituted words correspond to it. For instance, if one translation of the template assumes that some substituted words have male gender but others have female gender, create a separate translation of the template that conforms to the female gender.
    \item In one template, \{p.p1\} and \{p.p2\} refer to a property (either ``shape'', ``color'', ``size'', ``age'', or ``material''). \{p.v1\} and \{p.v2\} refer to an attribute of each property such as ``old'', ``new'', ``red'', ``blue'', etc.
\end{enumerate}

\begin{table}[t!]
\centering
\resizebox{\columnwidth}{!}{%
\begin{tabular}{ccccccc}
\toprule
\multirow{2}{*}{Lang.} & \multirow{2}{*}{Comparisons} & \multirow{2}{*}{Intensifiers} & \multirow{2}{*}{Properties} & Job vs & Animal vs & Animal vs \\
& & & & Nationality & Vehicles & Vehicles 2 \\ \midrule

af & \cellcolor[HTML]{F4FAF7}47.0 & \cellcolor[HTML]{F1B4AF}78.7 & \cellcolor[HTML]{CEEBDD}35.5 & \cellcolor[HTML]{5ABC8C}1.0 & \cellcolor[HTML]{FCF0EF}56.0 & \cellcolor[HTML]{9ED7BB}21.2 \\
ar & \cellcolor[HTML]{A7DBC1}23.9 & \cellcolor[HTML]{E8837B}97.5 & \cellcolor[HTML]{E67C73}100.0 & \cellcolor[HTML]{57BB8A}0.0 & \cellcolor[HTML]{E67C73}100.0 & \cellcolor[HTML]{A2D9BE}22.6 \\
az & \cellcolor[HTML]{E88279}98.0 & \cellcolor[HTML]{F3BEB9}75.1 & \cellcolor[HTML]{F4C2BE}73.5 & \cellcolor[HTML]{66C194}4.5 & \cellcolor[HTML]{DFF2E8}40.5 & \cellcolor[HTML]{E77E75}99.5 \\
bg & \cellcolor[HTML]{FDFEFD}49.5 & \cellcolor[HTML]{F5C8C5}71.1 & \cellcolor[HTML]{86CEAB}14.3 & \cellcolor[HTML]{57BB8A}0.0 & \cellcolor[HTML]{8ED1B0}16.5 & \cellcolor[HTML]{6BC398}6.1 \\
bn & \cellcolor[HTML]{EC9790}89.9 & \cellcolor[HTML]{EA8C84}94.0 & \cellcolor[HTML]{EB948D}91.0 & \cellcolor[HTML]{75C79F}9.0 & \cellcolor[HTML]{F9FCFB}48.5 & \cellcolor[HTML]{AADCC3}24.7 \\
de & \cellcolor[HTML]{90D2B1}17.1 & \cellcolor[HTML]{EB958E}90.6 & \cellcolor[HTML]{D7EEE3}38.2 & \cellcolor[HTML]{5FBE8F}2.5 & \cellcolor[HTML]{90D2B1}17.0 & \cellcolor[HTML]{75C79F}9.1 \\
el & \cellcolor[HTML]{6BC398}6.1 & \cellcolor[HTML]{E88279}98.0 & \cellcolor[HTML]{D7EFE3}38.3 & \cellcolor[HTML]{93D3B4}18.1 & \cellcolor[HTML]{EEF8F3}45.0 & \cellcolor[HTML]{8ED1B0}16.5 \\
en & \cellcolor[HTML]{6EC49A}7.0 & \cellcolor[HTML]{EB968E}90.4 & \cellcolor[HTML]{D8EFE4}38.5 & \cellcolor[HTML]{57BB8A}0.0 & \cellcolor[HTML]{6BC398}6.0 & \cellcolor[HTML]{58BB8B}0.5 \\
es & \cellcolor[HTML]{77C8A0}9.6 & \cellcolor[HTML]{E77E75}99.5 & \cellcolor[HTML]{FAE0DE}62.0 & \cellcolor[HTML]{57BB8A}0.0 & \cellcolor[HTML]{C4E7D6}32.5 & \cellcolor[HTML]{69C296}5.5 \\
et & \cellcolor[HTML]{7CCAA3}11.1 & \cellcolor[HTML]{EB938C}91.4 & \cellcolor[HTML]{D0ECDE}36.1 & \cellcolor[HTML]{87CEAB}14.5 & \cellcolor[HTML]{FBFDFC}49.0 & \cellcolor[HTML]{57BB8A}0.0 \\
eu & \cellcolor[HTML]{E67C73}100.0 & \cellcolor[HTML]{E78078}98.5 & \cellcolor[HTML]{F8D6D3}66.0 & \cellcolor[HTML]{91D2B2}17.5 & \cellcolor[HTML]{ACDDC5}25.5 & \cellcolor[HTML]{D9EFE4}38.7 \\
fa & \cellcolor[HTML]{7AC9A2}10.6 & \cellcolor[HTML]{EEA6A0}84.2 & \cellcolor[HTML]{FFFBFB}51.7 & \cellcolor[HTML]{67C195}5.0 & \cellcolor[HTML]{D8EFE4}38.5 & \cellcolor[HTML]{7DCAA5}11.6 \\
fi & \cellcolor[HTML]{7AC9A3}10.7 & \cellcolor[HTML]{F1B1AC}79.8 & \cellcolor[HTML]{C9E9D9}34.1 & \cellcolor[HTML]{64C093}4.0 & \cellcolor[HTML]{76C79F}9.2 & \cellcolor[HTML]{5CBD8D}1.5 \\
fr & \cellcolor[HTML]{64C093}4.0 & \cellcolor[HTML]{E78279}98.0 & \cellcolor[HTML]{8ACFAE}15.4 & \cellcolor[HTML]{59BC8B}0.8 & \cellcolor[HTML]{78C8A1}10.0 & \cellcolor[HTML]{84CDA9}13.5 \\
gu & \cellcolor[HTML]{E67C73}100.0 & \cellcolor[HTML]{E67C73}100.0 & \cellcolor[HTML]{E67C73}100.0 & \cellcolor[HTML]{DAF0E5}39.0 & \cellcolor[HTML]{EB978F}90.0 & \cellcolor[HTML]{C1E6D4}31.8 \\
ha & \cellcolor[HTML]{E67C73}100.0 & \cellcolor[HTML]{E67C73}100.0 & \cellcolor[HTML]{E77E75}99.5 & \cellcolor[HTML]{E67C73}100.0 & \cellcolor[HTML]{E67C73}100.0 & \cellcolor[HTML]{EB938B}91.5 \\
he & \cellcolor[HTML]{E67C73}100.0 & \cellcolor[HTML]{E8837B}97.5 & \cellcolor[HTML]{E67C73}100.0 & \cellcolor[HTML]{E67C73}100.0 & \cellcolor[HTML]{E67C73}100.0 & \cellcolor[HTML]{B4E0CB}27.9 \\
hi & \cellcolor[HTML]{A1D9BD}22.2 & \cellcolor[HTML]{F9DBD9}63.8 & \cellcolor[HTML]{F8DAD8}64.3 & \cellcolor[HTML]{71C59C}8.0 & \cellcolor[HTML]{B5E1CB}28.0 & \cellcolor[HTML]{73C69E}8.6 \\
ht & \cellcolor[HTML]{E98880}95.5 & \cellcolor[HTML]{E67C73}100.0 & \cellcolor[HTML]{E67C73}100.0 & \cellcolor[HTML]{E67C73}100.0 & \cellcolor[HTML]{E67C73}100.0 & \cellcolor[HTML]{EB948D}90.9 \\
hu & \cellcolor[HTML]{6BC398}6.1 & \cellcolor[HTML]{E78078}98.5 & \cellcolor[HTML]{C7E8D8}33.5 & \cellcolor[HTML]{86CEAA}14.0 & \cellcolor[HTML]{B5E1CB}28.0 & \cellcolor[HTML]{7AC9A2}10.6 \\
id & \cellcolor[HTML]{6CC399}6.5 & \cellcolor[HTML]{E88279}98.0 & \cellcolor[HTML]{F2B9B4}77.0 & \cellcolor[HTML]{57BB8A}0.0 & \cellcolor[HTML]{E4F4EC}42.0 & \cellcolor[HTML]{C7E8D8}33.5 \\
it & \cellcolor[HTML]{62BF92}3.5 & \cellcolor[HTML]{E78078}98.5 & \cellcolor[HTML]{FFFFFF}50.0 & \cellcolor[HTML]{6DC499}6.7 & \cellcolor[HTML]{7AC9A2}10.5 & \cellcolor[HTML]{69C296}5.6 \\
ja & \cellcolor[HTML]{E9877F}96.0 & \cellcolor[HTML]{E67C73}100.0 & \cellcolor[HTML]{FAE4E2}60.5 & \cellcolor[HTML]{E77F76}99.0 & \cellcolor[HTML]{BBE3D0}30.0 & \cellcolor[HTML]{E98880}95.5 \\
jv & \cellcolor[HTML]{75C79F}9.0 & \cellcolor[HTML]{E67C73}100.0 & \cellcolor[HTML]{EFACA6}82.0 & \cellcolor[HTML]{62BF92}3.5 & \cellcolor[HTML]{E67C73}100.0 & \cellcolor[HTML]{E67C73}100.0 \\
ka & \cellcolor[HTML]{57BB8A}0.0 & \cellcolor[HTML]{E98A82}95.0 & \cellcolor[HTML]{C9E9D9}34.0 & \cellcolor[HTML]{7FCBA6}12.0 & \cellcolor[HTML]{82CCA8}13.0 & \cellcolor[HTML]{7FCBA6}12.0 \\
kk & \cellcolor[HTML]{EC9992}88.9 & \cellcolor[HTML]{E77F76}99.0 & \cellcolor[HTML]{EFAAA5}82.5 & \cellcolor[HTML]{58BB8B}0.5 & \cellcolor[HTML]{E67C73}100.0 & \cellcolor[HTML]{90D2B1}17.0 \\
ko & \cellcolor[HTML]{95D4B5}18.5 & \cellcolor[HTML]{E78078}98.5 & \cellcolor[HTML]{CAE9DA}34.5 & \cellcolor[HTML]{75C79F}9.0 & \cellcolor[HTML]{E4F4EC}42.0 & \cellcolor[HTML]{FAE3E1}60.9 \\
lt & \cellcolor[HTML]{81CCA7}12.7 & \cellcolor[HTML]{EEA6A0}84.2 & \cellcolor[HTML]{F7D1CE}67.6 & \cellcolor[HTML]{ABDDC4}25.2 & \cellcolor[HTML]{A2D9BE}22.5 & \cellcolor[HTML]{5FBE8F}2.5 \\
ml & \cellcolor[HTML]{6EC49A}7.1 & \cellcolor[HTML]{F4C1BD}73.8 & \cellcolor[HTML]{F3C1BC}74.0 & \cellcolor[HTML]{6CC399}6.5 & \cellcolor[HTML]{C4E7D6}32.5 & \cellcolor[HTML]{96D4B6}19.0 \\
mr & \cellcolor[HTML]{57BB8A}0.0 & \cellcolor[HTML]{EFA9A3}83.0 & \cellcolor[HTML]{EFAAA4}82.8 & \cellcolor[HTML]{E67C73}100.0 & \cellcolor[HTML]{EEF8F3}45.0 & \cellcolor[HTML]{EDA099}86.4 \\
ms & \cellcolor[HTML]{64C093}4.0 & \cellcolor[HTML]{E88279}97.9 & \cellcolor[HTML]{EEA6A0}84.0 & \cellcolor[HTML]{69C296}5.5 & \cellcolor[HTML]{95D4B5}18.5 & \cellcolor[HTML]{58BB8B}0.5 \\
my & \cellcolor[HTML]{E77F76}99.0 & \cellcolor[HTML]{EDA19B}86.0 & \cellcolor[HTML]{EFA9A3}83.0 & \cellcolor[HTML]{76C7A0}9.5 & \cellcolor[HTML]{EA8E86}93.5 & \cellcolor[HTML]{57BB8A}0.0 \\
nl & \cellcolor[HTML]{86CEAA}14.1 & \cellcolor[HTML]{EB928A}91.9 & \cellcolor[HTML]{ACDDC5}25.4 & \cellcolor[HTML]{5ABC8C}1.0 & \cellcolor[HTML]{A7DBC2}24.1 & \cellcolor[HTML]{7DCAA4}11.3 \\
pa & \cellcolor[HTML]{E77E75}99.5 & \cellcolor[HTML]{FCEFEE}56.3 & \cellcolor[HTML]{E67C73}100.0 & \cellcolor[HTML]{57BB8A}0.0 & \cellcolor[HTML]{D6EEE2}38.0 & \cellcolor[HTML]{7AC9A2}10.6 \\
pl & \cellcolor[HTML]{9DD7BA}20.9 & \cellcolor[HTML]{E67C73}100.0 & \cellcolor[HTML]{9CD7BA}20.8 & \cellcolor[HTML]{57BB8A}0.0 & \cellcolor[HTML]{7BC9A3}11.0 & \cellcolor[HTML]{62BF92}3.5 \\
pt & \cellcolor[HTML]{FFFBFB}51.8 & \cellcolor[HTML]{E77F76}99.0 & \cellcolor[HTML]{F9DFDD}62.4 & \cellcolor[HTML]{59BC8B}0.8 & \cellcolor[HTML]{A4DABF}23.0 & \cellcolor[HTML]{6EC49A}7.1 \\
qu & \cellcolor[HTML]{EB928A}91.9 & \cellcolor[HTML]{E67C73}100.0 & \cellcolor[HTML]{E67C73}100.0 & \cellcolor[HTML]{E78279}98.0 & \cellcolor[HTML]{E98880}95.5 & \cellcolor[HTML]{E8847C}97.0 \\
ru & \cellcolor[HTML]{C2E6D4}32.0 & \cellcolor[HTML]{E98A82}95.0 & \cellcolor[HTML]{C8E9D9}33.8 & \cellcolor[HTML]{6EC49A}7.1 & \cellcolor[HTML]{A7DBC2}24.0 & \cellcolor[HTML]{61BF91}3.0 \\
sw & \cellcolor[HTML]{E67C73}100.0 & \cellcolor[HTML]{E67C73}100.0 & \cellcolor[HTML]{EA8C84}94.0 & \cellcolor[HTML]{6BC398}6.0 & \cellcolor[HTML]{E67C73}100.0 & \cellcolor[HTML]{E78279}98.0 \\
ta & \cellcolor[HTML]{FBE8E6}59.0 & \cellcolor[HTML]{F1B4AF}78.7 & \cellcolor[HTML]{F8D7D4}65.5 & \cellcolor[HTML]{7DCAA4}11.5 & \cellcolor[HTML]{84CDA9}13.5 & \cellcolor[HTML]{83CCA8}13.2 \\
te & \cellcolor[HTML]{CDEADC}35.2 & \cellcolor[HTML]{EB928A}92.0 & \cellcolor[HTML]{F7D0CD}68.0 & \cellcolor[HTML]{B5E1CB}28.0 & \cellcolor[HTML]{CFEBDE}36.0 & \cellcolor[HTML]{FDF2F1}55.0 \\
th & \cellcolor[HTML]{EB938C}91.4 & \cellcolor[HTML]{F1B5B0}78.4 & \cellcolor[HTML]{E67C73}100.0 & \cellcolor[HTML]{E67C73}100.0 & \cellcolor[HTML]{E67C73}100.0 & \cellcolor[HTML]{E0F2E9}41.0 \\
tl & \cellcolor[HTML]{57BB8A}0.0 & \cellcolor[HTML]{E67C73}100.0 & \cellcolor[HTML]{F7D4D1}66.5 & \cellcolor[HTML]{81CCA7}12.5 & \cellcolor[HTML]{FBE9E8}58.5 & \cellcolor[HTML]{84CDA9}13.6 \\
tr & \cellcolor[HTML]{E67C73}100.0 & \cellcolor[HTML]{EEA59F}84.4 & \cellcolor[HTML]{F4C5C1}72.5 & \cellcolor[HTML]{58BB8B}0.5 & \cellcolor[HTML]{9DD7BB}21.0 & \cellcolor[HTML]{86CEAB}14.1 \\
uk & \cellcolor[HTML]{8DD1AF}16.2 & \cellcolor[HTML]{E98A82}94.9 & \cellcolor[HTML]{DCF1E7}39.8 & \cellcolor[HTML]{B1DFC8}26.9 & \cellcolor[HTML]{FFFCFB}51.5 & \cellcolor[HTML]{86CEAB}14.1 \\
ur & \cellcolor[HTML]{EB948D}90.9 & \cellcolor[HTML]{FCECEB}57.5 & \cellcolor[HTML]{F2BBB6}76.1 & \cellcolor[HTML]{8ED1B0}16.5 & \cellcolor[HTML]{E67C73}100.0 & \cellcolor[HTML]{95D4B5}18.7 \\
vi & \cellcolor[HTML]{84CDA9}13.6 & \cellcolor[HTML]{E77F76}99.0 & \cellcolor[HTML]{F0B1AB}80.0 & \cellcolor[HTML]{78C8A1}10.0 & \cellcolor[HTML]{E67C73}100.0 & \cellcolor[HTML]{5FBE8F}2.6 \\
wo & \cellcolor[HTML]{E67C73}100.0 & \cellcolor[HTML]{E67C73}100.0 & \cellcolor[HTML]{E67C73}100.0 & \cellcolor[HTML]{E67C73}100.0 & \cellcolor[HTML]{E67C73}100.0 & \cellcolor[HTML]{E67C73}100.0 \\
yo & \cellcolor[HTML]{E67C73}100.0 & \cellcolor[HTML]{E67C73}100.0 & \cellcolor[HTML]{E67C73}100.0 & \cellcolor[HTML]{E67C73}100.0 & \cellcolor[HTML]{E67C73}100.0 & \cellcolor[HTML]{E77E75}99.5 \\
zh & \cellcolor[HTML]{E98B83}94.4 & \cellcolor[HTML]{E67C73}100.0 & \cellcolor[HTML]{C5E7D7}33.0 & \cellcolor[HTML]{E67C73}100.0 & \cellcolor[HTML]{E98B83}94.5 & \cellcolor[HTML]{F5C8C4}71.2 \\
\midrule
Avg & \cellcolor[HTML]{F5FBF8}47.3 & \cellcolor[HTML]{EB948D}90.9 & \cellcolor[HTML]{F8D9D6}64.8 & \cellcolor[HTML]{B0DFC8}26.7 & \cellcolor[HTML]{FFFBFB}51.6 & \cellcolor[HTML]{C5E7D6}32.8 \\
\bottomrule
\end{tabular}%
}
\caption{Error rate of XLM-R fine-tuned on English SQuAD v1.1 on 6 \textsc{CheckList} QA tests across all 50 languages.}
\label{tab:checklist_xmlr_full}
\end{table}

\begin{table}[t!]
\centering
\resizebox{\columnwidth}{!}{%
\begin{tabular}{ccccccc}
\toprule
\multirow{2}{*}{Lang.} & \multirow{2}{*}{Comparisons} & \multirow{2}{*}{Intensifiers} & \multirow{2}{*}{Properties} & Job vs & Animal vs & Animal vs \\
& & & & Nationality & Vehicles & Vehicles 2\\ \midrule

af & \cellcolor[HTML]{EA9088}92.5 & \cellcolor[HTML]{E67C73}100.0 & \cellcolor[HTML]{EB938B}91.5 & \cellcolor[HTML]{BBE3D0}30.0 & \cellcolor[HTML]{EDA099}86.5 & \cellcolor[HTML]{FEFAFA}52.0 \\
ar & \cellcolor[HTML]{EB928A}91.9 & \cellcolor[HTML]{E67C73}100.0 & \cellcolor[HTML]{E67C73}100.0 & \cellcolor[HTML]{8ED1B0}16.5 & \cellcolor[HTML]{E67C73}100.0 & \cellcolor[HTML]{F1B2AD}79.4 \\
az & \cellcolor[HTML]{E67C73}100.0 & \cellcolor[HTML]{E67C73}100.0 & \cellcolor[HTML]{EA8C84}94.0 & \cellcolor[HTML]{6CC399}6.5 & \cellcolor[HTML]{E98880}95.5 & \cellcolor[HTML]{E78078}98.5 \\
bg & \cellcolor[HTML]{E77F76}99.0 & \cellcolor[HTML]{E67C73}100.0 & \cellcolor[HTML]{EA8F88}92.9 & \cellcolor[HTML]{57BB8A}0.0 & \cellcolor[HTML]{F0B1AB}80.0 & \cellcolor[HTML]{F3BFBB}74.5 \\
bn & \cellcolor[HTML]{E67C73}100.0 & \cellcolor[HTML]{E67C73}100.0 & \cellcolor[HTML]{E67C73}100.0 & \cellcolor[HTML]{E67C73}100.0 & \cellcolor[HTML]{E67C73}100.0 & \cellcolor[HTML]{EC9B94}88.4 \\
de & \cellcolor[HTML]{E8837B}97.5 & \cellcolor[HTML]{E67C73}100.0 & \cellcolor[HTML]{EB968F}90.2 & \cellcolor[HTML]{78C8A1}10.0 & \cellcolor[HTML]{E67C73}100.0 & \cellcolor[HTML]{8DD0AF}16.2 \\
el & \cellcolor[HTML]{E9877F}95.9 & \cellcolor[HTML]{E67C73}100.0 & \cellcolor[HTML]{E98880}95.5 & \cellcolor[HTML]{FAE2E0}61.3 & \cellcolor[HTML]{E77F76}99.0 & \cellcolor[HTML]{EFA9A3}83.0 \\
en & \cellcolor[HTML]{E67C73}100.0 & \cellcolor[HTML]{E67C73}100.0 & \cellcolor[HTML]{E8867D}96.5 & \cellcolor[HTML]{57BB8A}0.0 & \cellcolor[HTML]{ABDDC4}25.0 & \cellcolor[HTML]{89CFAD}15.1 \\
es & \cellcolor[HTML]{E67C73}100.0 & \cellcolor[HTML]{E67C73}100.0 & \cellcolor[HTML]{ED9C96}87.8 & \cellcolor[HTML]{57BB8A}0.0 & \cellcolor[HTML]{B5E1CB}28.0 & \cellcolor[HTML]{78C8A1}10.1 \\
et & \cellcolor[HTML]{F1B4AF}78.9 & \cellcolor[HTML]{E67C73}100.0 & \cellcolor[HTML]{E8837A}97.6 & \cellcolor[HTML]{F2BAB5}76.5 & \cellcolor[HTML]{E67C73}100.0 & \cellcolor[HTML]{61BF91}3.0 \\
eu & \cellcolor[HTML]{E67C73}100.0 & \cellcolor[HTML]{E67C73}100.0 & \cellcolor[HTML]{E78078}98.5 & \cellcolor[HTML]{EB938B}91.5 & \cellcolor[HTML]{F1B4AE}79.0 & \cellcolor[HTML]{F2B8B3}77.4 \\
fa & \cellcolor[HTML]{E98880}95.5 & \cellcolor[HTML]{E67C73}100.0 & \cellcolor[HTML]{E77E75}99.5 & \cellcolor[HTML]{EFF8F4}45.5 & \cellcolor[HTML]{E77E75}99.5 & \cellcolor[HTML]{F6CBC8}69.8 \\
fi & \cellcolor[HTML]{EB958D}90.8 & \cellcolor[HTML]{E67C73}100.0 & \cellcolor[HTML]{F3BFBB}74.6 & \cellcolor[HTML]{8BD0AE}15.5 & \cellcolor[HTML]{EC9790}89.7 & \cellcolor[HTML]{E6F4ED}42.6 \\
fr & \cellcolor[HTML]{E67C73}100.0 & \cellcolor[HTML]{E67C73}100.0 & \cellcolor[HTML]{EFA8A1}83.6 & \cellcolor[HTML]{97D5B7}19.3 & \cellcolor[HTML]{90D2B1}17.0 & \cellcolor[HTML]{8BD0AE}15.5 \\
gu & \cellcolor[HTML]{E67C73}100.0 & \cellcolor[HTML]{E67C73}100.0 & \cellcolor[HTML]{E67C73}100.0 & \cellcolor[HTML]{EB918A}92.0 & \cellcolor[HTML]{E67C73}100.0 & \cellcolor[HTML]{E77F76}99.0 \\
ha & \cellcolor[HTML]{EB948D}91.0 & \cellcolor[HTML]{E67C73}100.0 & \cellcolor[HTML]{E67C73}100.0 & \cellcolor[HTML]{E67C73}100.0 & \cellcolor[HTML]{E67C73}100.0 & \cellcolor[HTML]{EA8C84}94.0 \\
he & \cellcolor[HTML]{E67C73}100.0 & \cellcolor[HTML]{E67C73}100.0 & \cellcolor[HTML]{E67C73}100.0 & \cellcolor[HTML]{E67C73}100.0 & \cellcolor[HTML]{E67C73}100.0 & \cellcolor[HTML]{F3BCB8}75.6 \\
hi & \cellcolor[HTML]{E77F76}99.0 & \cellcolor[HTML]{E67C73}100.0 & \cellcolor[HTML]{EA8D85}93.8 & \cellcolor[HTML]{81CCA7}12.5 & \cellcolor[HTML]{EA8C84}94.0 & \cellcolor[HTML]{F0AFA9}80.8 \\
ht & \cellcolor[HTML]{F5C6C2}72.0 & \cellcolor[HTML]{E67C73}100.0 & \cellcolor[HTML]{E67C73}100.0 & \cellcolor[HTML]{E67C73}100.0 & \cellcolor[HTML]{E67C73}100.0 & \cellcolor[HTML]{F6D0CD}68.2 \\
hu & \cellcolor[HTML]{EC9790}89.8 & \cellcolor[HTML]{E67C73}100.0 & \cellcolor[HTML]{EA8C84}94.0 & \cellcolor[HTML]{F9E0DE}62.0 & \cellcolor[HTML]{F0AEA9}81.0 & \cellcolor[HTML]{FCECEB}57.3 \\
id & \cellcolor[HTML]{E77F76}99.0 & \cellcolor[HTML]{E67C73}100.0 & \cellcolor[HTML]{F0B1AB}80.0 & \cellcolor[HTML]{71C59C}8.0 & \cellcolor[HTML]{EA8C84}94.0 & \cellcolor[HTML]{C2E6D4}32.0 \\
it & \cellcolor[HTML]{E77E75}99.5 & \cellcolor[HTML]{E67C73}100.0 & \cellcolor[HTML]{E98880}95.7 & \cellcolor[HTML]{9AD6B9}20.2 & \cellcolor[HTML]{DBF0E6}39.5 & \cellcolor[HTML]{C0E5D3}31.3 \\
ja & \cellcolor[HTML]{E67C73}100.0 & \cellcolor[HTML]{E67C73}100.0 & \cellcolor[HTML]{E77F76}99.0 & \cellcolor[HTML]{E67C73}100.0 & \cellcolor[HTML]{E67C73}100.0 & \cellcolor[HTML]{F1B5B0}78.4 \\
jv & \cellcolor[HTML]{E98B83}94.5 & \cellcolor[HTML]{E67C73}100.0 & \cellcolor[HTML]{E77E75}99.5 & \cellcolor[HTML]{FFFFFF}50.0 & \cellcolor[HTML]{E67C73}100.0 & \cellcolor[HTML]{EB938B}91.5 \\
ka & \cellcolor[HTML]{E67C73}100.0 & \cellcolor[HTML]{E67C73}100.0 & \cellcolor[HTML]{E77E75}99.5 & \cellcolor[HTML]{FBE9E8}58.5 & \cellcolor[HTML]{F6FBF9}47.5 & \cellcolor[HTML]{FEF8F7}53.0 \\
kk & \cellcolor[HTML]{E88279}98.0 & \cellcolor[HTML]{E67C73}100.0 & \cellcolor[HTML]{E8837A}97.5 & \cellcolor[HTML]{69C296}5.5 & \cellcolor[HTML]{E67C73}100.0 & \cellcolor[HTML]{F3C1BC}74.0 \\
ko & \cellcolor[HTML]{E8837A}97.5 & \cellcolor[HTML]{E67C73}100.0 & \cellcolor[HTML]{F3BEB9}75.0 & \cellcolor[HTML]{87CEAB}14.5 & \cellcolor[HTML]{E77E75}99.5 & \cellcolor[HTML]{FDF2F1}55.3 \\
lt & \cellcolor[HTML]{EA8F87}92.9 & \cellcolor[HTML]{E67C73}100.0 & \cellcolor[HTML]{E77E75}99.5 & \cellcolor[HTML]{D3EDE0}37.0 & \cellcolor[HTML]{EA8C84}94.0 & \cellcolor[HTML]{D3EDE0}37.0 \\
ml & \cellcolor[HTML]{E88279}98.0 & \cellcolor[HTML]{E67C73}100.0 & \cellcolor[HTML]{EB918A}92.0 & \cellcolor[HTML]{E67C73}100.0 & \cellcolor[HTML]{E67C73}100.0 & \cellcolor[HTML]{E8847C}97.0 \\
mr & \cellcolor[HTML]{E67C73}100.0 & \cellcolor[HTML]{E67C73}100.0 & \cellcolor[HTML]{E67C73}100.0 & \cellcolor[HTML]{E67C73}100.0 & \cellcolor[HTML]{E67C73}100.0 & \cellcolor[HTML]{E77E75}99.5 \\
ms & \cellcolor[HTML]{E67C73}100.0 & \cellcolor[HTML]{E67C73}100.0 & \cellcolor[HTML]{E77E75}99.5 & \cellcolor[HTML]{71C59C}8.0 & \cellcolor[HTML]{E67C73}100.0 & \cellcolor[HTML]{C0E5D3}31.3 \\
my & \cellcolor[HTML]{E67C73}100.0 & \cellcolor[HTML]{E67C73}100.0 & \cellcolor[HTML]{EA8E86}93.5 & \cellcolor[HTML]{EEA59F}84.5 & \cellcolor[HTML]{E67C73}100.0 & \cellcolor[HTML]{FEF8F8}52.8 \\
nl & \cellcolor[HTML]{E8867D}96.5 & \cellcolor[HTML]{E67C73}100.0 & \cellcolor[HTML]{EA9189}92.2 & \cellcolor[HTML]{EFF8F4}45.5 & \cellcolor[HTML]{E2F3EB}41.5 & \cellcolor[HTML]{74C69E}8.8 \\
pa & \cellcolor[HTML]{E67C73}100.0 & \cellcolor[HTML]{E67C73}100.0 & \cellcolor[HTML]{E67C73}100.0 & \cellcolor[HTML]{FDF5F4}54.0 & \cellcolor[HTML]{E67C73}100.0 & \cellcolor[HTML]{E67C73}100.0 \\
pl & \cellcolor[HTML]{E77F76}99.0 & \cellcolor[HTML]{E67C73}100.0 & \cellcolor[HTML]{EC9A93}88.9 & \cellcolor[HTML]{C7E8D8}33.5 & \cellcolor[HTML]{F0AEA9}81.0 & \cellcolor[HTML]{FEF6F5}53.8 \\
pt & \cellcolor[HTML]{E88279}98.0 & \cellcolor[HTML]{E67C73}100.0 & \cellcolor[HTML]{E8837A}97.6 & \cellcolor[HTML]{5FBE8F}2.5 & \cellcolor[HTML]{F9DCDA}63.5 & \cellcolor[HTML]{B0DFC8}26.8 \\
qu & \cellcolor[HTML]{E67C73}100.0 & \cellcolor[HTML]{E67C73}100.0 & \cellcolor[HTML]{E67C73}100.0 & \cellcolor[HTML]{E67C73}100.0 & \cellcolor[HTML]{E8867D}96.5 & \cellcolor[HTML]{E78078}98.5 \\
ru & \cellcolor[HTML]{E8877F}96.0 & \cellcolor[HTML]{E67C73}100.0 & \cellcolor[HTML]{EA8F87}93.0 & \cellcolor[HTML]{F4C3BF}73.1 & \cellcolor[HTML]{F1B4AE}79.0 & \cellcolor[HTML]{90D2B1}17.0 \\
sw & \cellcolor[HTML]{E67C73}100.0 & \cellcolor[HTML]{E67C73}100.0 & \cellcolor[HTML]{E67C73}100.0 & \cellcolor[HTML]{FDF1F0}55.5 & \cellcolor[HTML]{E67C73}100.0 & \cellcolor[HTML]{EDA099}86.5 \\
ta & \cellcolor[HTML]{EB978F}90.0 & \cellcolor[HTML]{E67C73}100.0 & \cellcolor[HTML]{E67C73}100.0 & \cellcolor[HTML]{9DD7BB}21.0 & \cellcolor[HTML]{E67C73}100.0 & \cellcolor[HTML]{E98880}95.4 \\
te & \cellcolor[HTML]{EC9B94}88.4 & \cellcolor[HTML]{E67C73}100.0 & \cellcolor[HTML]{E98880}95.5 & \cellcolor[HTML]{EB948D}91.0 & \cellcolor[HTML]{E67C73}100.0 & \cellcolor[HTML]{E78078}98.5 \\
th & \cellcolor[HTML]{E77E75}99.5 & \cellcolor[HTML]{E67C73}100.0 & \cellcolor[HTML]{E67C73}100.0 & \cellcolor[HTML]{E67C73}100.0 & \cellcolor[HTML]{E67C73}100.0 & \cellcolor[HTML]{E67C73}100.0 \\
tl & \cellcolor[HTML]{E77E75}99.5 & \cellcolor[HTML]{E67C73}100.0 & \cellcolor[HTML]{E67C73}100.0 & \cellcolor[HTML]{E8867D}96.5 & \cellcolor[HTML]{E67C73}100.0 & \cellcolor[HTML]{E67C73}100.0 \\
tr & \cellcolor[HTML]{E67C73}100.0 & \cellcolor[HTML]{E67C73}100.0 & \cellcolor[HTML]{E67C73}100.0 & \cellcolor[HTML]{F5CAC6}70.5 & \cellcolor[HTML]{F2B9B4}77.0 & \cellcolor[HTML]{F7D1CE}67.7 \\
uk & \cellcolor[HTML]{E9877F}95.8 & \cellcolor[HTML]{E67C73}100.0 & \cellcolor[HTML]{EEA29C}85.6 & \cellcolor[HTML]{F0ADA7}81.5 & \cellcolor[HTML]{F8D8D5}65.0 & \cellcolor[HTML]{D6EEE2}37.9 \\
ur & \cellcolor[HTML]{E67C73}100.0 & \cellcolor[HTML]{E67C73}100.0 & \cellcolor[HTML]{E67C73}100.0 & \cellcolor[HTML]{E77E75}99.5 & \cellcolor[HTML]{E67C73}100.0 & \cellcolor[HTML]{E67C73}100.0 \\
vi & \cellcolor[HTML]{E8837B}97.5 & \cellcolor[HTML]{E67C73}100.0 & \cellcolor[HTML]{EB958E}90.5 & \cellcolor[HTML]{70C59B}7.5 & \cellcolor[HTML]{E67C73}100.0 & \cellcolor[HTML]{BFE5D2}31.1 \\
wo & \cellcolor[HTML]{FBE7E5}59.5 & \cellcolor[HTML]{E67C73}100.0 & \cellcolor[HTML]{E67C73}100.0 & \cellcolor[HTML]{E67C73}100.0 & \cellcolor[HTML]{E67C73}100.0 & \cellcolor[HTML]{EB938B}91.5 \\
yo & \cellcolor[HTML]{E77F76}99.0 & \cellcolor[HTML]{E67C73}100.0 & \cellcolor[HTML]{E77F76}99.0 & \cellcolor[HTML]{E67C73}100.0 & \cellcolor[HTML]{E67C73}100.0 & \cellcolor[HTML]{F8D7D5}65.3 \\
zh & \cellcolor[HTML]{E67C73}100.0 & \cellcolor[HTML]{E67C73}100.0 & \cellcolor[HTML]{E67C73}100.0 & \cellcolor[HTML]{E67C73}100.0 & \cellcolor[HTML]{E67C73}100.0 & \cellcolor[HTML]{EA9089}92.4 \\
Avg & \cellcolor[HTML]{E9877F}95.8 & \cellcolor[HTML]{E67C73}100.0 & \cellcolor[HTML]{E98981}95.3 & \cellcolor[HTML]{FDF2F1}55.1 & \cellcolor[HTML]{ED9E98}87.0 & \cellcolor[HTML]{F8DBD8}64.1 \\
\bottomrule
\end{tabular}%
}
\caption{Error rate of mBERT fine-tuned on English SQuAD v1.1 on 6 \textsc{CheckList} QA tests across all 50 languages.}
\label{tab:checklist_mbert_full}
\end{table}

\noindent \textbf{Challenges of template translation} $\:$ We highlight some of the challenges and linguistic phenomena we encountered during the process of creating \multichecklist in the following. Unless specified otherwise, we create separate templates to disambiguate each phenomenon.
\begin{itemize}
    \itemsep0em
    \item \textbf{Gender agreement}: Adjectives and nationalities need to be declined to match the gender of their referring expression. To keep the translation effort manageable and avoid creating separate templates for each gender, we control for gender and restrict fill-in values to male names for the affected tests (3/6). We sample genders equally for the other tests. We welcome dedicated test suites analyzing multilingual gender bias as future extensions.
    \item \textbf{Declination}: In Russian, animal and vehicle names require Accusative and Nominative in different cases.
    \item \textbf{Normalization}: For appropriate substitution, fill-in values often need to include articles. We normalize answers and predictions by removing language-specific articles in order to ensure a consistent comparison.
    \item \textbf{Names}: Our use of names based on data in Wikidata leads to certain biases. Names that are more common in Wikidata are more likely to be chosen. In some cases, names in Wikidata are not written in the native script. Japanese names from Wikidata are often written in hiragana or katakana rather than kanji. Our choice of using the first name is also not applicable to all languages. In Japanese, people are usually not referred to by their first name, e.g. Masanori Suzuki would be called Suzuki-san instead of Masanori.
    \item \textbf{Declension of names}: In some languages, a suffix is appended to a name depending on its spelling. For instance, in Turkish the suffix changes based on the vowel of the last syllable, e.g. Ahmet'in, Ali'nin, Umut'un, Şeyma'nın, Özge'nin, etc., and Ahmet'ten, Ali'den, Umut'tan, Şeyma'dan, Özge'den, etc. In Finnish, names are appended with a variation of ``lla'', e.g. Peterillä, Lisalla, Mattilla, etc.
    \item \textbf{Professions}: Words for certain professions are gendered (e.g. \textit{waiter}/\textit{waitress}), so they only occur with male or female names.
    \item \textbf{Question syntax}: In some languages, the syntax of the question changes depending on the property or adjective one asks about.
    \item \textbf{Syntax of adjectives}: In some languages, the syntax changes depending on what adjective is used. In German, the translations of ``happy'', ``excited'', and ``passionate'' require different prepositions.
    \item \textbf{Stilted language}: Some text appears stilted when values are filled into the translated templates. For instance, the question \begin{CJK}{UTF8}{min}``どちらの方が冷静でないですか。''\end{CJK} is an unusual way to do negation in Japanese; if directly translated to English, it would mean ``Who is more not calm?''.
\end{itemize}

We tried to address most of these challenges by instructing translators to create additional templates to disambiguate linguistic phenomena and by consolidating different templates programmatically.
% \footnote{We restrict selection to male names to avoid creating separate templates necessary for disambiguating gender agreement for names. We aim to create balanced tests in the future.} 
However, as this process was relatively labor-intensive, we recommend the usage of morphologically aware templates similar to \citet{jiang2020x-factr} for future work. Note that morphologically aware templates may not be able to resolve some of the finer linguistic differences. For this reason, we also advise working closely with native speakers to design tests that reflect natural language as closely as possible.

\noindent \textbf{Full results} $\:$ We show the full results of XLM-R and mBERT on the \multichecklist tests in Tables \ref{tab:checklist_xmlr_full} and \ref{tab:checklist_mbert_full} respectively. mBERT only shows limited amounts of cross-lingual taxonomic knowledge. While it is able to distinguish between job and nationality and animals and vehicles in some languages, it fails to do this consistently across all languages. In addition, it completely fails to distinguish between different properties and intensifiers and is not able to perform comparisons. In contrast, while XLM-R struggles with intensifiers, it demonstrates the other capabilities much more consistently across languages.

\paragraph{Example failure cases} We provide example failure cases of XLM-R on a subset of languages in Table \ref{tab:checklist_failure_cases}. We will publicly release a comprehensive list of failure cases for XLM-R and mBERT, the complete tests and model outputs for further analysis.

\begin{table*}[]
\centering
\includegraphics[width=\textwidth]{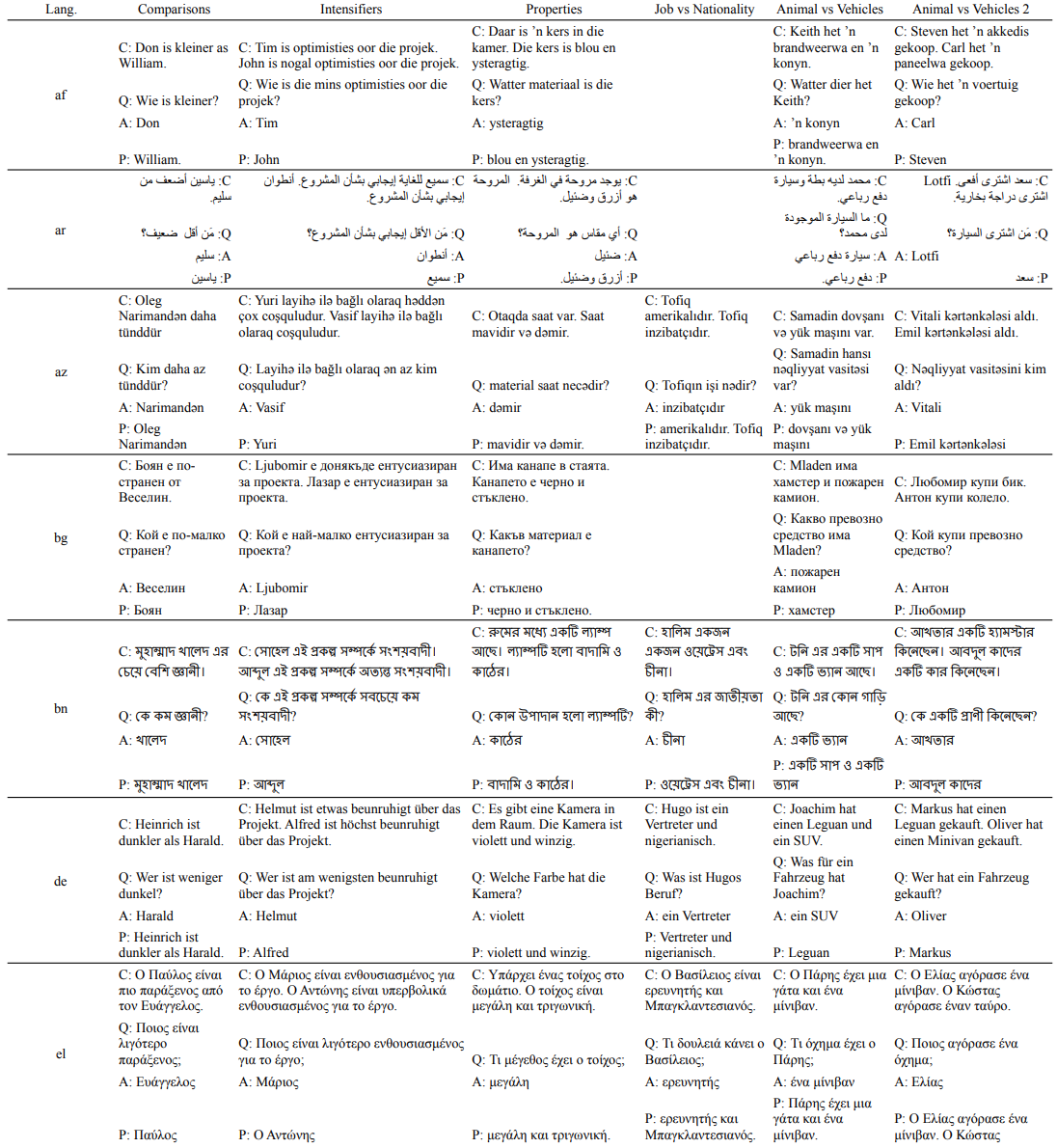}
\caption{Example failure cases of XLM-R on a subset of languages. Each failure case consists of a context (C), a question (Q), an answer (A), and XLM-R's prediction (P).}
\label{tab:checklist_failure_cases}
\end{table*}

\section{Hyper-parameters}

\noindent \textbf{mBERT} $\:$ We use the cased version, which covers 104 languages, has 12 layers, 768 hidden units per layer, 12 attention heads, a 110k shared WordPiece vocabulary, and 110M parameters.\footnote{\url{https://github.com/google-research/bert/blob/master/multilingual.md}} The model was trained using Wikipedia data in all 104 languages, oversampling low-resource languages with an exponential smoothing factor of 0.7. We generally fine-tune mBERT for two epochs, with a training batch size of 32 and a learning rate of 2e-5. We build on the Transformers library \cite{wolf2019huggingface} for training on each task.

\noindent \textbf{XLM-R} $\:$ We use the XLM-R Large version that covers 100 languages, uses a 200k shared BPE vocabulary, and has been trained with masked language modelling.\footnote{\url{https://github.com/facebookresearch/XLM}} We fine-tune XLM-R generally for two epochs with a learning rate of 3e-5 and an effective batch size of 16. We use the Transformers library for training XLM-R on all tasks.

\noindent \textbf{mT5} $\:$ We use the publicly released mT5-XXL version that has nearly 13 billion parameters with a vocabulary size 250k \cite{Xue2020mt5}. It has been trained on multilingual C4 (mC4) corpus which has 6.3 trillion tokens spanning 101 languages\footnote{\url{https://www.tensorflow.org/datasets/catalog/c4\#c4multilingual}}. For all downstream tasks, we fine-tune mT5-XXL for 10k steps with a constant learning rate of 0.001, dropout rate of 0.1 and a batch size of 2\textsuperscript{17} tokens. For early stopping, we save checkpoints
every 200 steps and choose the checkpoint with the
highest performance on the validation set.

\section{Metadata} \label{app:metadata}

\begin{table*}[ht]
\centering
\begin{tabular}{lccc}
\toprule
\multicolumn{1}{c}{\multirow{2}{*}{Model}} & \multirow{2}{*}{\begin{tabular}[c]{@{}c@{}}Number of parameters \\ (in millions)\end{tabular}} & \multicolumn{2}{c}{Pre-training data}                                    \\
\multicolumn{1}{c}{}                       &  & \multicolumn{1}{l}{Monolingual data} & \multicolumn{1}{l}{Parallel data} \\ \midrule
mBERT  & 178 & 85 GB & N/A  \\
XLM-R (large)  & 559    & 6.3T tokens & N/A \\
MMTE   & 190 & N/A & 25B pairs \\
mT5 & 13,000 & 1T tokens  & N/A \\
RemBERT & 575 & 1.8T tokens & N/A \\
X-STILTS & 559 & 6.3T tokens & N/A\\
FILTER & 559 & 6.3T tokens & N/A\\
VECO & 559 & 1.3 TB & 6.4M pairs \\
T-URLv2 + StableTune & 559 & 2.1 TB & 42 GB \\
ERNIE-M & 559 & 1.5 TB & 69 GB \\
\bottomrule
\end{tabular}
\caption{Metadata for the current submissions to XTREME. Note that monolingual pre-training data is reported in either number of tokens or size of the data (in GB/TB). The amount of parallel data is reported in either number of pairs or size of the data (in GB/TB).}
\label{app:metadata-table}
\end{table*}
We intend to ask each submission to \name for relevant metadata. Such metadata includes the number of parameters, amount of pre-training data, amount of fine-tuning data, etc. We are doing this to enhance transparency and to increase utility of our benchmark for practitioners with varying needs. As a first step in this direction, we provide information about the number of parameters and the amount of monolingual and parallel pre-training data used by all submissions to \xtreme in Table~\ref{app:metadata-table}. Note that the different systems report their training data in different ways (e.g. number of tokens, number of examples, size of the data). We plan to standardize this by asking submissions to \name to report training data in terms of number of tokens seen.
% \mj{Maybe talk about other metadata we want like fine-tuning data, inference latency, etc.}
% \mj{Add info about whether additional data was used during fine-tuning and whether translations were used.}

\section{Language-agnostic Retrieval Results} \label{app:retrieval-results}

The multiway cross-language nature of Mewsli-X and LAReQA enables closer analysis of model performance by input and target \emph{language pairs}.
Mewsli-X can directly be split by language pair as it has a single correct target per input mention.
For LAReQA, we follow the ``Limit to One Target'' strategy of \citet{roy-etal-2020-lareqa}: instead of asking the model to retrieve all correct answers in one pass, we evaluate on each target separately, with all the other correct answers removed from the candidate pool, allowing us to report splits by language pair.

%to filter the candidate set for a given target language in turn to recalculate results and obtain a similar split by language pairs.

\begin{figure*}[]
\centering
\begin{subfigure}{0.5\textwidth}
  \centering
  \includegraphics[height=0.75\textheight]{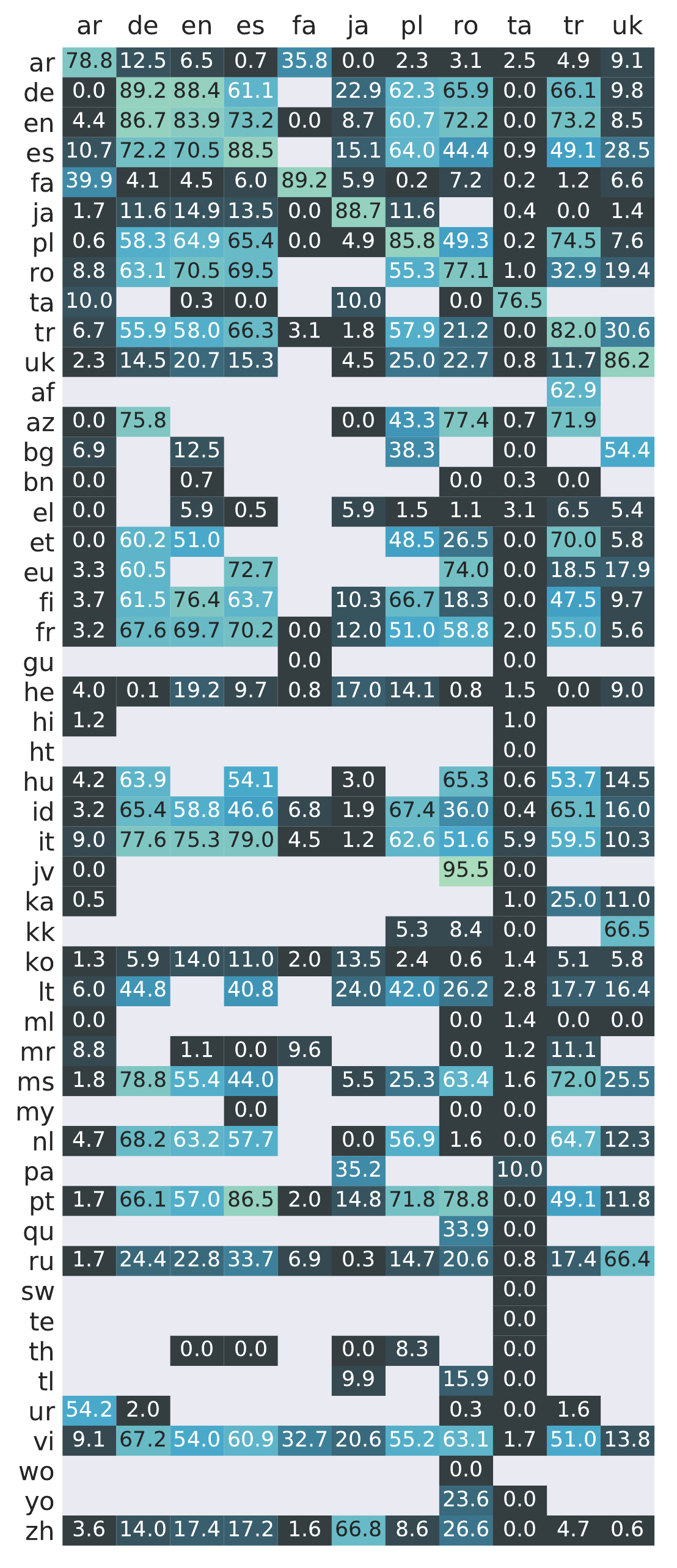}
  \caption{mBERT}
\end{subfigure}%
\begin{subfigure}{.5\textwidth}
  \centering
  \includegraphics[height=0.75\textheight]{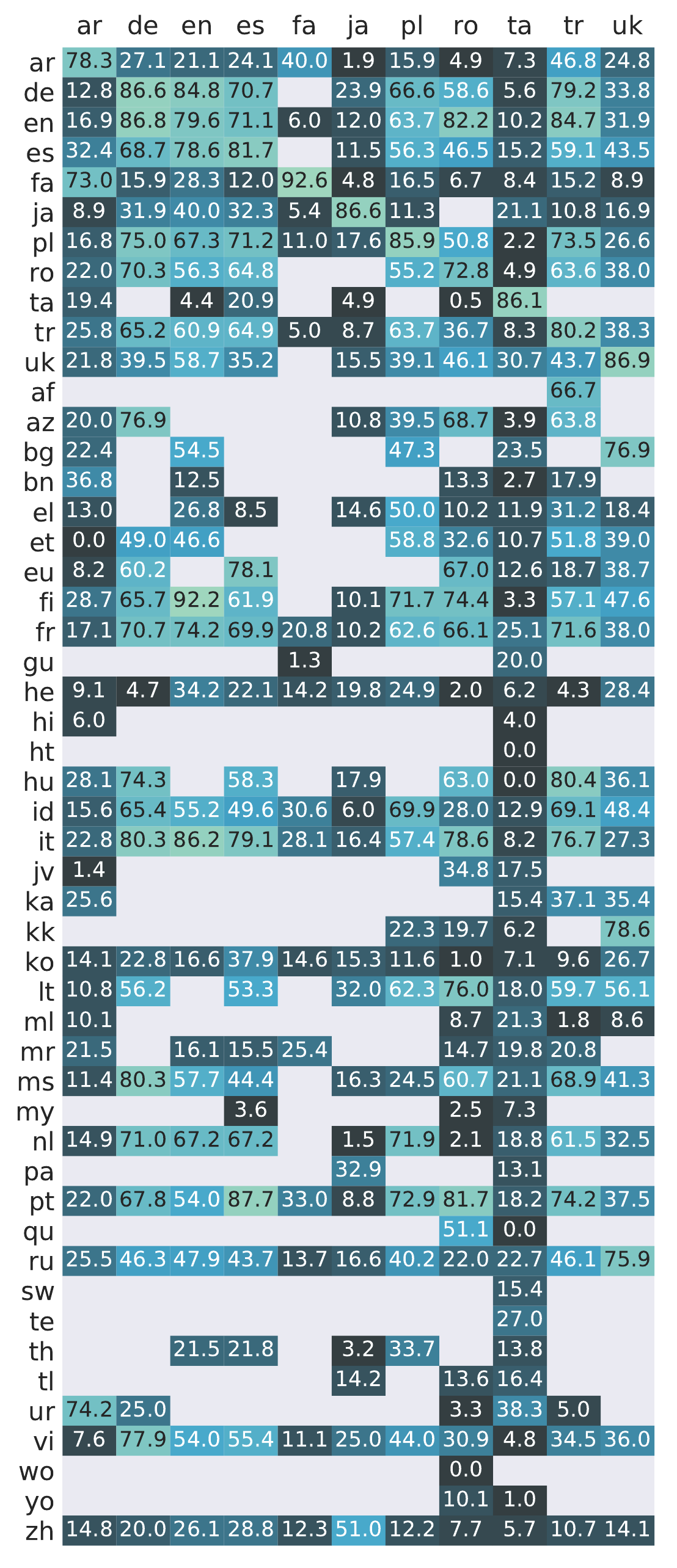}
  \caption{XLM-R Large}
\end{subfigure}
\caption{Mewsli-X results broken down by mention language (columns) and entity description language (rows), only showing combinations with at least 10 test items.}
\label{fig:mewslix_result_heatmaps}
\end{figure*}

\begin{figure*}[]
\centering
\begin{subfigure}{0.5\textwidth}
  \centering
  \includegraphics[width=\linewidth]{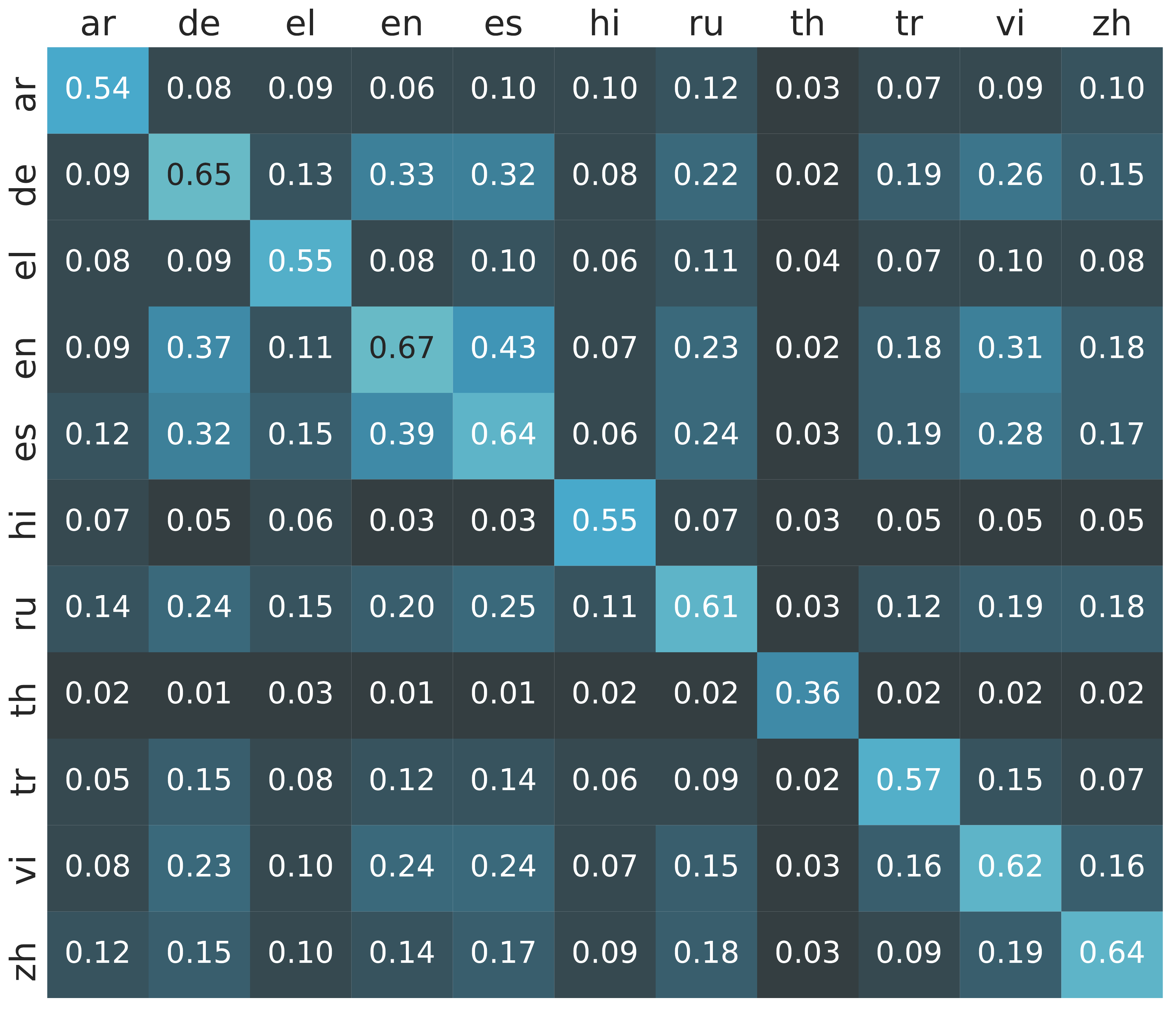}
  \caption{mBERT}
\end{subfigure}%
\begin{subfigure}{.5\textwidth}
  \centering
  \includegraphics[width=\linewidth]{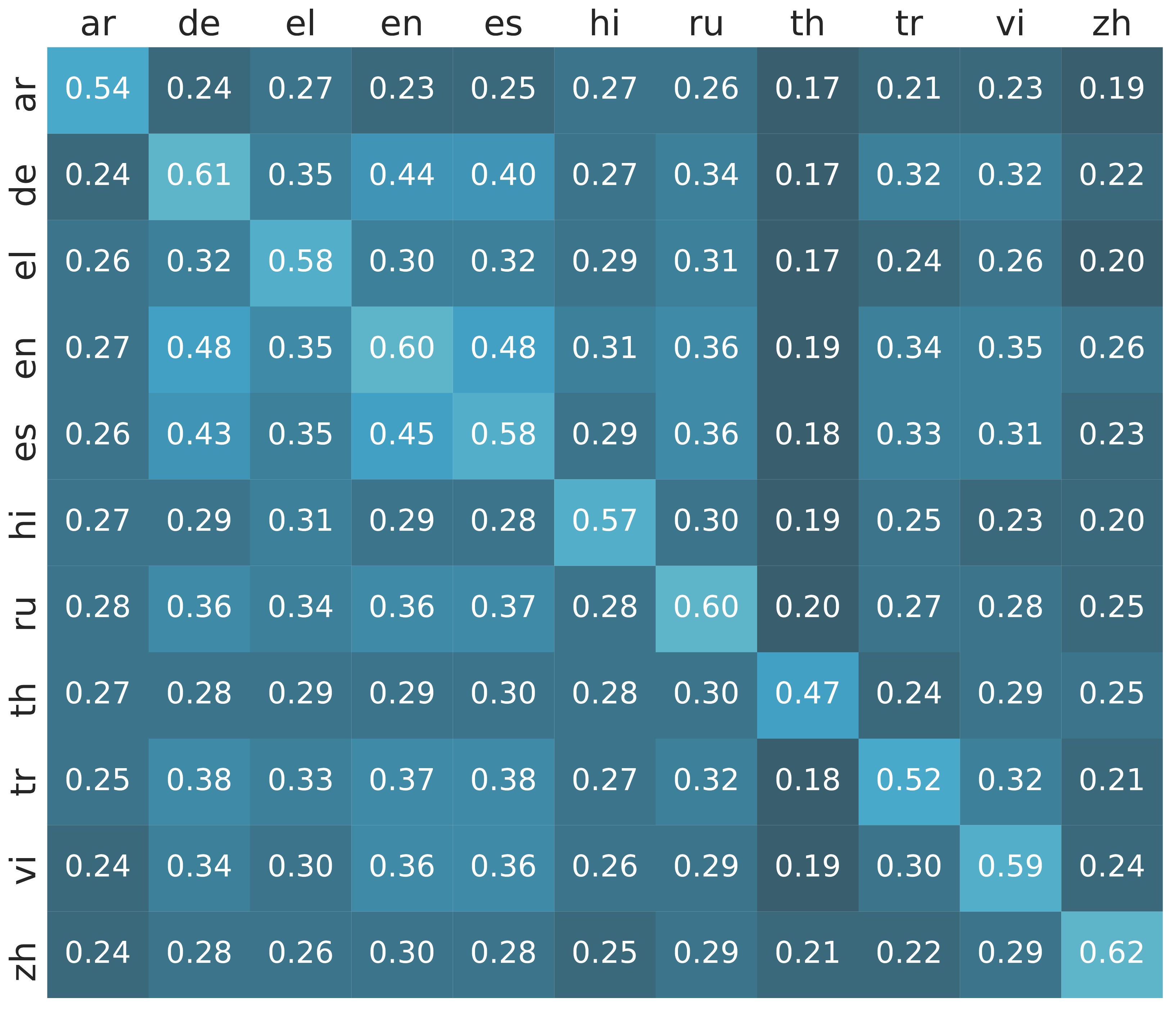}
  \caption{XLM-R Large}
\end{subfigure}
\caption{LAReQA results broken down by question language (rows) and answer language (columns) for ``Limit to One Target'' inference. }
\label{fig:lareqa_result_heatmaps}
\end{figure*}

\autoref{app:retrieval_same_diff} summarizes these pairwise
mAP@20 scores (here, micro-averaged), showing
that XLM-R Large improves substantially over mBERT on the cross-lingual case (+38\% on Mewsli-X and +137\% for LAReQA) in exchange for a slight drop for the same-language case.
Even so, performance on the cross-lingual case is still low at 29--36 mAP@20, and 
remains a challenging area for future work.
Figures~\ref{fig:mewslix_result_heatmaps} and \ref{fig:lareqa_result_heatmaps} show the detailed breakdowns.

\begin{table*}[]
\centering
\begin{tabular}{@{}lcccc@{}} \toprule
 & \multicolumn{2}{c}{Mewsli-X} & \multicolumn{2}{c}{LAReQA} \\ \cmidrule(l){2-3} \cmidrule(l){4-5}
Subset \textbackslash{} Model & mBERT & XLM-R Large & mBERT & XLM-R Large \\ \midrule
All & 40.2 & 47.1 & 16.3 & 31.3 \\
Same language & 85.7 & 83.6 & 58.2 & 57.1 \\
Different languages & 25.8 & 35.5 & 12.1 & 28.7 \\ \bottomrule
\end{tabular}
\caption{Language-agnostic retrieval results (mAP@20) broken down by whether the query and answer languages are the same or different.}
\label{app:retrieval_same_diff}
\end{table*}

\section{Detailed results}

We show the detailed results for each task and language in Tables \ref{tab:xnli-results} (XNLI), \ref{tab:xcopa-results} (XCOPA), \ref{tab:pos-results} (UD-POS), \ref{tab:ner-results} (WikiANN-NER), \ref{tab:xquad-results} (XQuAD), \ref{tab:mlqa-results} (MLQA), \ref{tab:tydiqa-results} (TyDiQA-GoldP), \ref{tab:tatoeba-results} (Tatoeba), \ref{tab:mewslix-results} (Mewsli-X), and \ref{tab:lareqa-results} (LAReQA).

\begin{table*}[]
\centering
\resizebox{\textwidth}{!}{
\begin{tabular}{lccccccccccccccc|c}
\toprule
Model & en & ar & bg & de & el & es & fr & hi & ru & sw & th & tr & ur & vi & zh & \textbf{avg} \\ \midrule
mBERT & 81.7 & 66.0 & 69.2 & 71.1 & 66.8 & 74.9 & 74.2 & 60.7 & 70.2 & 49.3 & 54.7 & 61.2 & 58.2 & 70.5 & 69.0 & 65.4 \\
XLM-R & {88.7} & {77.2} & {83.0} & {82.5} & {80.8} & {83.7} & {82.2} & {75.6} & {79.1} & {71.2} & {77.4} & {78.0} & {71.7} & {79.3} & {78.2} & {79.2} \\
mT5 & 92.6 & 84.5 & 87.0 & 87.3 & 86.9  & 88.4 & 87.4 & 82.3 & 84.3 & 80.6 & 81.2 & 83.4 & 79.8 & 84.0 & 83.0 & 84.8 \\
\textit{\begin{tabular}[c]{@{}l@{}}mBERT translate-train\end{tabular}} & 80.8 & 73.6 & 76.6 & 77.4 & 75.7 & 78.1 & 77.4 & {71.9} & 75.2 & 69.4 & {70.9} & {75.3} & 67.2 & 75.0 & 74.1 & 74.6 \\
\textit{\begin{tabular}[c]{@{}l@{}}mBERT translate-train-all\end{tabular}} & 81.9 & {73.8} & {77.6} & {77.6} & {75.9} & {79.1} & {77.8} & 70.7 & {75.4} & {70.5} & 70.0 & 74.3 & {67.4} & {77.0} & {77.6} & {75.1} \\ \bottomrule
\end{tabular}
}
\caption{XNLI results (accuracy) across languages.}
\label{tab:xnli-results}
\end{table*}

\begin{table*}[]
\centering
\resizebox{\textwidth}{!}{%
\begin{tabular}{lccccccccccc|c}
\toprule
Model & et & ht & id & it & qu & sw & ta & th & tr & vi & zh & Avg \\ \midrule
mBERT & 54.6 & 51.8 & 55.4 & 57.6 & 54.0 & 52.2 & 55.6 & 52.0 & 55.4 & 60.4 & 67.6 & 56.1 \\
XLM-R & 68.2 & 52.6 & 80.6 & 71.0 & 52.8 & 61.8 & 73.8 & 74.4 & 72.0 & 76.2 & 78.0 & 69.2 \\
mT5 & 77.5 & 72.1 & 81.1 & 75.9 & 54.4 & 74.1 & 75.9 & 78.3 & 78.1 & 76.9 & 79.5 & 74.9 \\
\textit{\begin{tabular}[c]{@{}l@{}}mBERT translate-train\end{tabular}} & 57.4 & 55.6 & 60.6 & 63.4 & 47.8 & 51.6 & 54.8 & 56.4 & 58.8 & 59.6 & 64.0 & 57.3 \\
\textit{\begin{tabular}[c]{@{}l@{}}mBERT translate-train-all\end{tabular}} & 56.0 & 54.4 & 59.0 & 60.4 & 51.0 & 56.8 & 55.4 & 58.2 & 56.8 & 63.6 & 65.4 & 57.9 \\
\bottomrule
\end{tabular}
}
\caption{XCOPA results (accuracy) across languages.}
\label{tab:xcopa-results}
\end{table*}

\begin{table*}[]
\centering
\resizebox{\textwidth}{!}{%
\begin{tabular}{lcccccccccccccccccccc}
\toprule
Model & af & ar & bg & de & el & en & es & et & eu & fa & fi & fr & he & hi & hu & id & it & ja & kk & ko \\ \midrule
mBERT & 86.1 & 54.0 & 85.2 & 85.6 & 80.4 & 95.4 & 85.8 & 79.5 & 59.3 & 66.2 & 78.0 & 83.4 & 55.3 & 67.1 & 78.3 & 80.6 & 88.2 & 49.7 & 69.8 & 49.2 \\
XLM-R Large & 89.7 & 68.1 & 88.6 & 88.5 & 86.5 & 96.1 & 89.1 & 87.2 & 74.3 & 73.9 & 86.3 & 88.4 & 68.2 & 74.4 & 83.4 & 82.9 & 89.8 & 29.4 & 79.3 & 54.0 \\ \midrule
 & mr & nl & pt & ru & ta & te & th & tl & tr & ur & vi & yo & zh & lt & pl & uk & wo & ro & Avg &  \\ \midrule
mBERT & 66.6 & 88.6 & 86.8 & 85.1 & 68.6 & 75.2 & 39.1 & 68.5 & 68.3 & 58.0 & 53.4 & 53.4 & 61.2 & 77.9 & 79.9 & 80.4 & 28.9 & 77.3 & 70.9 &  \\
XLM-R Large & 85.3 & 89.5 & 89.3 & 89.7 & 77.3 & 85.3 & 47.7 & 75.0 & 75.4 & 67.2 & 56.8 & 22.8 & 40.8 & 84.5 & 84.8 & 85.8 & 27.6 & 85.5 & 75.0 & \\ \bottomrule
\end{tabular}
}
\caption{UD-POS results (F1 score) across languages.}
\label{tab:pos-results}
\end{table*}

\begin{table*}[]
\centering
\resizebox{\textwidth}{!}{
\begin{tabular}{lccccccccccccccccccccccccc}
\toprule
Model & ar & he & vi & id & jv & ms & tl & eu & ml & ta & te & af & nl & en & de & el & bn & hi & mr & ur & fa & fr & it & pt & es \\ \midrule
mBERT & 43.9 & 56.6 & 72.1 & 62.9 & 64.6 & 70.9 & 73.7 & 65.0 & 53.5 & 53.0 & 47.4 & 76.0 & 81.9 & 84.5 & 78.0 & 68.8 & 70.3 & 66.7 & 57.3 & 35.6 & 50.1 & 79.1 & 81.3 & 79.8 & 72.2 \\
XLM-R large & 43.7 & 54.1 & 77.2 & 52.3 & 58.7 & 69.8 & 72.2 & 62.1 & 65.8 & 56.9 & 52.3 & 77.7 & 84.3 & 84.6 & 78.0 & 77.2 & 76.3 & 71.0 & 64.2 & 54.1 & 61.1 & 79.1 & 81.1 & 79.6 & 68.8 \\ \midrule
 & bg & ru & ja & ka & ko & th & sw & yo & my & zh & kk & tr & et & fi & hu & qu & pl & uk & az & lt & pa & gu & ro & Avg &  \\ \midrule
mBERT & 77.2 & 63.5 & 28.4 & 65.0 & 59.1 & 2.3 & 72.7 & 49.4 & 49.1 & 43.3 & 49.0 & 72.1 & 77.0 & 77.4 & 75.0 & 57.4 & 79.9 & 69.3 & 65.6 & 74.3 & 37.0 & 47.3 & 72.1 & 62.7 &  \\
XLM-R large & 81.2 & 71.5 & 18.3 & 68.9 & 58.0 & 1.5 & 69.9 & 41.8 & 50.9 & 25.8 & 49.9 & 78.9 & 78.0 & 78.6 & 79.3 & 50.4 & 81.3 & 73.1 & 69.2 & 76.9 & 46.8 & 60.7 & 79.6 & 64.4 & \\ \bottomrule
\end{tabular}
}
\caption{WikiANN-NER results (F1 score) across languages.}
\label{tab:ner-results}
\end{table*}

\begin{table*}[]
\centering
\resizebox{\textwidth}{!}{%
\begin{tabular}{lccccccccccc|c}
\toprule 
Model & en & es & de & el & ru & tr & ar & vi & th & zh & hi & avg \\ \midrule
mBERT & 84.5 & 75.1 & 73.2 & 62.9 & 71.3 & 53.7 & 62.2 & 70.1 & 43.5 & 59.6 & 59.5 & 65.1 \\ 
XLM-R Large & 87.4 & 82.7 & 80.9 & 80.7 & 80.5 & 76.1 & 74.4 & 79.2 & 75.4 & 55.0 & 77.0 & 77.2 \\
mT5 & 90.2 & 84.6 & 82.3 & 82.8 & 78.8 & 76.5 & 80.3 & 83.3 & 74.7 & 81.7 & 81.7 & 81.5 \\
\textit{\begin{tabular}[c]{@{}l@{}}mBERT translate-train\end{tabular}} & 83.5 & 80.2 & 75.6 & 70.0 & 75.0 & 68.9 & 68.0 & 75.6 & 36.9 & 66.2 & 69.6 & 70.0 \\
\textit{\begin{tabular}[c]{@{}l@{}}mBERT translate-train-all\end{tabular}} & 86.0 & 82.4 & 78.8 & 74.2 & 78.1 & 70.6 & 71.0 & 78.5 & 38.1 & 67.7 & 71.3 & 72.4 \\
\bottomrule
\end{tabular}
}
\caption{XQuAD results (F1) across languages.}
\label{tab:xquad-results}
\end{table*}

\begin{table*}[]
\centering
% \resizebox{\textwidth}{!}{%
\begin{tabular}{lccccccc | c}
\toprule
Model & en & es & de & ar & hi & vi & zh & avg \\ \midrule
mBERT & 80.7 & 66.2 & 60.2 & 51.6 & 49.9 & 60.2 & 60.3 & 61.3 \\
XLM-R Large & 83.9 & 74.4 & 70.3 & 67.0 & 70.8 & 74.2 & 68.4 & 72.7 \\
mT5 & 86.4 & 76.2 & 73.1 & 70.2 & 75.3 & 76.5 & 71.4 & 75.6 \\
\textit{\begin{tabular}[c]{@{}l@{}}mBERT translate-train\end{tabular}} & 80.2 & 70.0 & 64.4 & 55.0 & 60.1 & 65.7 & 63.9 & 65.6 \\
\textit{\begin{tabular}[c]{@{}l@{}}mBERT translate-train-all\end{tabular}}  & 80.7 & 71.3 & 66.0 & 58.9 & 62.4 & 67.9 & 66.0 & 67.6
\\ \bottomrule
\end{tabular}%
% }
\caption{MLQA results (F1) across languages.}
\label{tab:mlqa-results}
\end{table*}

\begin{table*}[]
\centering
% \resizebox{\textwidth}{!}{%
\begin{tabular}{lccccccccc|c}
\toprule
Model & en & ar & bn & fi & id & ko & ru & sw & te & avg \\ \midrule
mBERT & 69.4 & 61.7 & 53.5 & 57.4 & 63.2 & 57.6 & 56.5 & 59.7 & 46.2 & 58.4 \\
XLM-R Large & 69.2 & 66.1 & 59.1 & 64.7 & 73.8 & 58.0 & 62.2 & 66.4 & 59.1 & 64.3 \\
mT5 & 83.1 & 82.4 & 83.6 & 81.2 & 84.5 & 73.2 & 78.7 & 87.2 & 83.6 & 81.9 \\
\textit{\begin{tabular}[c]{@{}l@{}}mBERT translate-train\end{tabular}} & 75.3 & 61.5 & 31.9 & 62.6 & 68.6 & 53.2 & 53.1 & 61.9 & 27.4 & 55.1 \\
\textit{\begin{tabular}[c]{@{}l@{}}mBERT translate-train-all\end{tabular}} & 73.2 & 71.8 & 49.7 & 68.1 & 72.3 & 58.6 & 64.3 & 66.8 & 53.3 & 64.2 \\
\bottomrule
\end{tabular}%
% }
\caption{TyDiQA-GoldP results (F1) across different languages.}
\label{tab:tydiqa-results}
\end{table*}

\begin{table*}[]
\centering
\resizebox{\textwidth}{!}{%
\begin{tabular}{lccccccccccccccccccccc}
\toprule
Model & ar & he & vi & id & jv & tl & eu & ml & ta & te & af & nl & de & el & bn & hi & mr & ur & fa & fr & it \\ \midrule
mBERT & 32.7 & 46.8 & 66.2 & 57.7 & 18.5 & 17.5 & 31.0 & 19.2 & 23.5 & 26.1 & 49.6 & 66.1 & 79.7 & 29.1 & 20.7 & 38.6 & 23.7 & 38.4 & 49.9 & 67.1 & 65.6 \\
XLM-R & 68.3 & 77.6 & 91.0 & 88.4 & 28.8 & 60.8 & 58.6 & 83.6 & 65.8 & 80.8 & 74.9 & 90.0 & 96.6 & 76.6 & 67.6 & 88.9 & 70.4 & 76.5 & 85.3 & 87.5 & 82.4 \\ \midrule
 & pt & es & bg & ru & ja & ka & ko & th & sw & zh & kk & tr & et & fi & hu & az & lt & pl & uk & ro & Avg \\ \midrule
mBERT & 74.4 & 71.8 & 52.8 & 65.1 & 50.8 & 21.1 & 43.9 & 15.0 & 12.1 & 74.9 & 30.4 & 38.4 & 34.0 & 42.7 & 44.2 & 37.2 & 35.4 & 54.1 & 55.8 & 54.8 & 41.3 \\
XLM-R & 89.8 & 89.5 & 84.4 & 86.1 & 75.5 & 66.2 & 81.9 & 80.5 & 31.3 & 79.5 & 63.8 & 84.5 & 73.4 & 86.5 & 83.1 & 74.7 & 77.5 & 88.0 & 82.6 & 89.6 & 76.2 \\
\bottomrule
\end{tabular}%
}
\caption{Tatoeba results (accuracy) across different languages.}
\label{tab:tatoeba-results}
\end{table*}

\begin{table*}[]
\centering
\begin{tabular}{@{}lccccccccccc|c@{}}
\toprule
Model & ar & de & en & es & fa & ja & pl & ro & ta & tr & uk & avg \\ \midrule
mBERT & 15.3 & 61.0 & 54.8 & 59.4 & 13.5 & 44.2 & 57.7 & 27.5 & 4.1 & 49.9 & 37.0 & 38.6 \\
XLM-R Large & 28.7 & 64.8 & 59.7 & 62.0 & 24.6 & 45.0 & 62.1 & 30.8 & 14.9 & 59.4 & 51.2 & 45.7 \\ \bottomrule
\end{tabular}
\caption{Mewsli-X results (mean average precision@20) across different input languages.}
\label{tab:mewslix-results}
\end{table*}

\begin{table*}[]
\centering
\begin{tabular}{@{}lccccccccccc|c@{}}
\toprule
Model & ar & de & el & en & es & hi & ru & th & tr & vi & zh & avg \\ \midrule
mBERT & 17.0 & 29.3 & 16.3 & 31.3 & 30.8 & 12.3 & 27.2 & 5.8 & 17.9 & 25.3 & 24.2 & 21.6 \\
XLM-R Large & 34.6 & 42.8 & 38.8 & 46.2 & 43.7 & 38.2 & 42.5 & 39.5 & 41.3 & 40.9 & 39.8 & 40.7 \\ \bottomrule
\end{tabular}
\caption{LAReQA results (mean average precision@20) across different question languages.}
\label{tab:lareqa-results}
\end{table*}

\section{Nuanced Multilingual Evaluation} \label{app:nuanced-evaluation}

We perform nuanced multilingual evaluations by categorizing testing examples into different attribute buckets and measuring the system performance on each attribute bucket. In the following, we describe the available attributes for tasks in \name and provide additional analysis on different attributes.

\subsection{Attribute Definition}
\label{sec:attribute}
\noindent \textbf{QA} $\:$ We denote $(\mathbf{X}_c, \mathbf{X}_q, \mathbf{X}_a)$ as a tuple of the corresponding context, question and answer, and refer to \texttt{cLen}, \texttt{qLen}, \texttt{aLen} as their lengths (i.e., the number of tokens).
We use BLEU \cite{papineni-etal-2002-bleu} to measure the lexical overlap between $(\mathbf{X}_a, \mathbf{X}_q)$ and  $(\mathbf{X}_q, \mathbf{X}_c)$ as \texttt{BLEU-AQ} and \texttt{BLEU-QC}. We classify questions based on their first tokens and report the top 5 most frequent question types as \texttt{qType} (i.e., what, how, when, where, which), which cover 85\% of questions in the training set. We list the six attributes as follows.

\begin{itemize}
%  \item $\phi_{\texttt{aLen}}(\mathbf{X}_c, \mathbf{X}_q, \mathbf{X}_a) = \mathrm{len}(a)$,
%  \item $\phi_{\texttt{qLen}}(\mathbf{X}_c, \mathbf{X}_q, \mathbf{X}_a) = \mathrm{len}(q)$,
%  \item $\phi_{\texttt{cLen}}(\mathbf{X}_c, \mathbf{X}_q, \mathbf{X}_a) = \mathrm{len}(c)$,
%  \item $\phi_{\texttt{BLEU}_{aq}}(\mathbf{X}_a, \mathbf{X}_q) = \texttt{BLEU}(\mathbf{X}_a, \mathbf{X}_q)$,       
%  \item $\phi_{\texttt{BLEU}_{qc}}(\mathbf{X}_q, \mathbf{X}_c) = \texttt{BLEU}(\mathbf{X}_q, \mathbf{X}_c)$,      
%  \item $\phi_{\texttt{type}}(\mathbf{X}_q) = \mathrm{type}(\textbf{X}_q)$~~ \textit{question type}, 
 \item $\phi_{\texttt{aLen}}(\mathbf{X}_a) = \mathrm{len}(\mathbf{X}_a)$,
 \item $\phi_{\texttt{qLen}}(\mathbf{X}_q) = \mathrm{len}(\mathbf{X}_q)$,
 \item $\phi_{\texttt{cLen}}(\mathbf{X}_c) = \mathrm{len}(\mathbf{X}_c)$,
 \item $\phi_{\texttt{BLEU}_{aq}}(\mathbf{X}_a, \mathbf{X}_q) = \texttt{BLEU}(\mathbf{X}_a, \mathbf{X}_q)$,       
 \item $\phi_{\texttt{BLEU}_{qc}}(\mathbf{X}_q, \mathbf{X}_c) = \texttt{BLEU}(\mathbf{X}_q, \mathbf{X}_c)$,      
 \item $\phi_{\texttt{type}}(\mathbf{X}_q) = \mathrm{type}(\textbf{X}_q)$,~~ \textit{question type}. 
\end{itemize}

\noindent \textbf{Structured Prediction} $\:$ Given a sentence $\mathbf{X}$, we define the $i$-th word token as $x_i$ and a span of words in the range of $[i,j)$ as $\mathbf{X}_{i:j}$ in the sentence. We then define five attributes including the label of a span (\texttt{tag}), the token length of a sentence (\texttt{sLen}), the token length of an entity span (\texttt{eLen}), the character length of an entity span (\texttt{tLen}) and the relative token position of an entity (\texttt{rPos}) in the sentence as follows.

% We define a token $x_i$ or span $\mathrm{x}$ and its ground truth or prediction label $y = \mathrm{lab}(\cdot)$ in a sentence $\mathbf{X} = \mathrm{sent}(\mathbf{x})$.

\begin{itemize}
    \item $\phi_{\texttt{tag}}(\mathbf{X}_{i:j}) = \mathrm{label}(\mathbf{X}_{i:j})$
    \item $\phi_{\texttt{sLen}}(\mathbf{X}) = \mathrm{len}(\mathbf{X})$
    \item $\phi_{\texttt{eLen}}(\mathbf{X}_{i:j}) = \mathrm{len}(\mathbf{X}_{i:j})$
    \item $\phi_{\texttt{tLen}}(\mathbf{X}_{i:j}) = |\mathbf{X}_{i:j}|$
    \item $\phi_{\texttt{rPos}}(\mathbf{X}_{i:j}) = i/\phi_{\texttt{sLen}(\mathbf{X})}$, ~~ \textit{relative position}
\end{itemize} 
where $|x|$ represents the number of characters. %while $|\mathbf{x}|$ denotes the number of tokens. $i$ is the starting index of span $\mathbf{x}$ in the sentence $\mathbf{X}$.

\subsection{Attribute Buckets}
We bucket all test examples into different attribute buckets for a given attribute. Specifically, for an attribute defined for a task, we measure the attribute value of the test examples (see Section~\ref{sec:attribute}), then determine $N$ attribute buckets for all test examples ($N=4$ by default), and finally we measure the system performance on the test examples falling in each attribute interval to observe the performance change over different attribute buckets. Since the attribute values can be either continuous (e.g., answer length \texttt{aLen}) or discrete (e.g., question type \texttt{qType}), we perform different strategies for creating attribute buckets for them.

\noindent \textbf{Continuous Attribute Values} $\:$ For the attributes with continuous attribute values (e.g., $\phi_{\texttt{aLen}}$, $\phi_{\texttt{qLen}}$, $\phi_{\texttt{cLen}}$, $\phi_{\texttt{BLEU}_{aq}}$,
$\phi_{\texttt{BLEU}_{qc}}$), we divide the test examples into different intervals where the numbers of the test samples in all attribute intervals are equal.
%the by dividing the test sample equally. 

\noindent \textbf{Discrete Attribute Values} $\:$ For the attribute with discrete attribute values (e.g., $\phi_{\texttt{type}}$), test samples with the same type are put into the same attribute bucket.

Table~\ref{tab:interval-xquad} and~\ref{tab:interval-ner} show the detailed attribute intervals for each category on the XQuAD task and WikiANN-NER task, respectively.

\begin{table*}[htb]
  \centering  \scriptsize  %\footnotesize
  \renewcommand\tabcolsep{2pt}
    \begin{tabular}{ccccccccccc}
    \toprule
    \multirow{2}[4]{*}{Bucket} & aLen  & qLen  & cLen  & BLEU\_AQ & BLEU\_QC & aLen  & qLen  & cLen  & BLEU\_AQ & BLEU\_QC \\
\cmidrule{2-11}          & \multicolumn{5}{c}{en}                & \multicolumn{5}{c}{zh} \\
\cmidrule(lr){1-1} \cmidrule(lr){2-6} \cmidrule(lr){7-11} 
    XS    & [1, 2] & [3, 8] & [25, 91] & 0     & [0, 1.64e-08] & 1     & [1, 13] & [3, 8] & 0     & [0, .0075] \\
    S     & (2, 5] & (8, 11] & (91, 111] & (0, .016] & (1.64e-08, 2.8e-06] & (1, 6] & (13, 17] & (8, 11] & (0, .0118] & (.0075, .01] \\
    L     & (5, 13] & (11, 16] & (111, 156] & (.016, .128] & (2.8e-06, 7.92e-05] & (6, 15] & (17, 22] & (11, 17] & (.0118, .0542] & (.01, .013] \\
    XL    & (13, 25] & (16, 29] & (156, 509] & (.128, .146] & (7.92e-05, .0589] & (15, 72] & (22, 52] & (17, 144] & (.0542, .328] & (.013, .097 \\
    \midrule
          & \multicolumn{5}{c}{hi}                & \multicolumn{5}{c}{el} \\
    \cmidrule(lr){1-1} \cmidrule(lr){2-6} \cmidrule(lr){7-11} 
    XS    & [1, 2] & [3, 8] & [28, 104] & 0     & [0, 2.02e-08] & [1, 2] & [3, 8] & [29, 97] & 0     & [4.99e-39, 9.58e-09] \\
    S     & (2, 5] & (8, 11] & (104, 131] & (0, .015] & (2.02e-08, 2.91e-06] & (2, 5] & (8, 11] & (97, 120] & (0, .014] & (9.58e-09, 1.32e-06] \\
    L     & (5, 13] & (11, 15] & (131, 171] & (.015, .19] & (2.91e-06, 7.29e-05] & (5, 11] & (11, 16] & (120, 162] & (.014, .019] & (1.32e-06, 4.37e-05] \\
    XL    & (13, 29] & (15, 53] & (171, 575] & (.019, .24] & (7.29e-05, .0628] & (11, 26] & (16, 30] & (162, 554] & (.019, .12] & (4.37e-05, .033] \\
    \midrule
          & \multicolumn{5}{c}{ru}                & \multicolumn{5}{c}{tr} \\
    \cmidrule(lr){1-1} \cmidrule(lr){2-6} \cmidrule(lr){7-11} 
    XS    & [1, 2] & [3, 6] & [25, 86] & 0     & [1.34e-39, 3.24e-09] & [1, 2] & [2, 6] & [18, 75] & 0     & [0, 6.37e-09] \\
    S     & (2, 5] & (6, 8] & (86, 104] & (0, .017] & (3.24e-09, 6.2e-07] & (2, 5] & (6, 8] & (75, 94] & (0, .016] & (6.37e-09, 9.33e-07] \\
    L     & (5, 12] & (8, 11] & (104, 140] & (.017, .023] & (6.2e-07, 2.07e-05] & (5, 11] & (8, 11] & (94, 130] & (.016, .057] & (9.33e-07, 2.67e-05] \\
    XL    & (12, 23] & (11, 26] & (140, 470] & (.023, .287] & (2.07e-05, .0294] & (11, 20] & (11, 20] & (130, 414] & (.057, .143] & (2.67e-05, .0408] \\
    \midrule
          & \multicolumn{5}{c}{ar}                & \multicolumn{5}{c}{vi} \\
    \cmidrule(lr){1-1} \cmidrule(lr){2-6} \cmidrule(lr){7-11} 
    XS    & [1, 2] & [2, 7] & [24, 83] & 0     & [0, 4.72e-09] & [1, 2] & [3, 10] & [39, 127] & 0     & [0, 1.47e-08] \\
    S     & (2, 4] & (7, 9] & (83, 104] & (0, .0243] & (4.72e-09, 8.51e-07] & (2, 4] & (10, 13] & (127, 160] & (0, .0139] & (1.47e-08, 2.43e-06] \\
    L     & (4, 9] & (9, 12] & (104, 138] & (.0243, .293] & (8.51e-07, 3.39e-05] & (4, 8] & (13, 17] & (160, 213] & (.0139, .135] & (2.43e-06, 7.76e-05] \\
    XL    & (9, 22] & (12, 38] & (138, 432] & (.293, .246] & (3.39e-05, .0266] & (8, 34] & (17, 40] & (213, 763] & (.135, .189] & (7.76e-05, .0546] \\
    \bottomrule
    \end{tabular}%
    \caption{Detailed attribute intervals for each category on the XQuAD task.}
  \label{tab:interval-xquad}%
\end{table*}%

\begin{table*}[htb]
  \centering \scriptsize
  \renewcommand\tabcolsep{2pt}
    \begin{tabular}{ccccccccccccc}
    \toprule
    \multirow{2}[2]{*}{Bucket} & \multicolumn{4}{c}{en}        & \multicolumn{4}{c}{es}        & \multicolumn{4}{c}{fr} \\
        \cmidrule(lr){2-13}
          & sLen  & eLen  & rPos  & eDen  & sLen  & eLen  & rPos  & eDen  & sLen  & eLen  & rPos  & eDen \\
    \cmidrule(lr){1-1}\cmidrule(lr){2-5} \cmidrule(lr){6-9} \cmidrule(lr){10-13}
    XS    & [3,5] & 1     & 0     & [0,0.33] & [3,5] & 1     & 0     & [0,0.33] & [1,5] & 1     & 0     & [0.027,0.33] \\
    S     & (5,8] & 2     & (0,0.33] & (0.33,0.56] & (5,7] & 2     & (0,0.25] & (0.33,0.67] & (5,7] & 2     & (0,0.33] & (0.33,0.64] \\
    L     & (8,16] & (2,4] & (0.33,0.62] & (0.56,1] & (7,18] & (2,4] & (0.25,0.6] & (0.67,0.86] & (7,16] & (2,4] & (0.33,0.67] & (0.64,1] \\
    XL    & (16,1e+03) & (4,1e+03] & (0.62,1] &   -    & (18,1e+03] & (4,1e+03] & (0.6,1] & (0.86,1] & (16,1e+03] & (4,1e+03] & (0.67,1] &  -\\
    \bottomrule
    \end{tabular}%
     \caption{Detailed attribute intervals for each category on WikiANN-NER task in \texttt{en}, \texttt{es}, and \texttt{fr}.}
  \label{tab:interval-ner}%
\end{table*}%

\subsection{Additional Nuanced Analysis}

% \subsubsection{WikiANN-NER}

\subsubsection{WikiANN-NER}
Table~\ref{tab:ner_ernie_M} and \ref{tab:ner_xlm_r} illustrate the single system diagnosis of ERNIE-M and XLM-R  respectively on the WikiANN-NER task in three languages (i.e., \texttt{en, es, fr}). We make the following observations.

\noindent \textbf{ERNIE-M} $\:$ In Table~\ref{tab:ner_ernie_M}, first, we observe that the effects of some attributes for \texttt{ERNIE-M} are language-independent. For example, based on the attribute \texttt{rPos}, the system is good at predicting entities located within the first 1/3 part of the English sentences, while it is relatively bad at predicting entities within the first 1/3 part of the sentences for other languages. Second, the system favors long sentences based on the attribute \texttt{sLen}. We even observe that performance increases as the sentence length increases on \texttt{es} and \texttt{fr}. Third, across all languages, the system performs relatively bad at predicting long entities (\texttt{eLen}) and entities belonging to the organization class (\texttt{tag}). Finally, the system is good at predicting sentences with fewer entities based on the attribute for entity density (\texttt{eDen}).
%that are longer and whose proportion of entities is larger.
%Moreover, long entities and entities with organization category are difficult to identify.
% Another interesting finding is, for samples of Romance languages, entities that are at the beginning of sentences are easier to identify.
%, which accounts for more than $40\%$. 

\noindent \textbf{XLM-R} $\:$ In Table~\ref{tab:ner_xlm_r}, we observe that the influence of some attributes such as \texttt{sLen},\texttt{eLen}, \texttt{eDen} with respect to the system performance are similar between \texttt{ERNIE-M} and \texttt{XLM-R}, although \texttt{ERNIE-M} performs significantly better than \texttt{XLM-R} at generalizing its predictions on \texttt{es, fr}.
% \paragraph{XLM-R}
% The influence of these attributes \texttt{sLen},\texttt{eLen}, \texttt{eDen} show similar influence on the system \texttt{XLM-R}.

\begin{table*}[!ht]
  \centering
  \renewcommand\tabcolsep{0.75pt}
\renewcommand\arraystretch{0.8}
    \begin{tabular}{crrrrr}
    \toprule
    Lang. & \multicolumn{1}{c}{sLen} & \multicolumn{1}{c}{eLen} & \multicolumn{1}{c}{rPos} & \multicolumn{1}{c}{eDen} & \multicolumn{1}{c}{tag}  \\
    \midrule
    \multicolumn{6}{l}{English} \\ 
    \midrule
    \multirow{2}[1]{*}{en}  &  \multirow{4}[1]{*}{\includegraphics[scale=0.4]{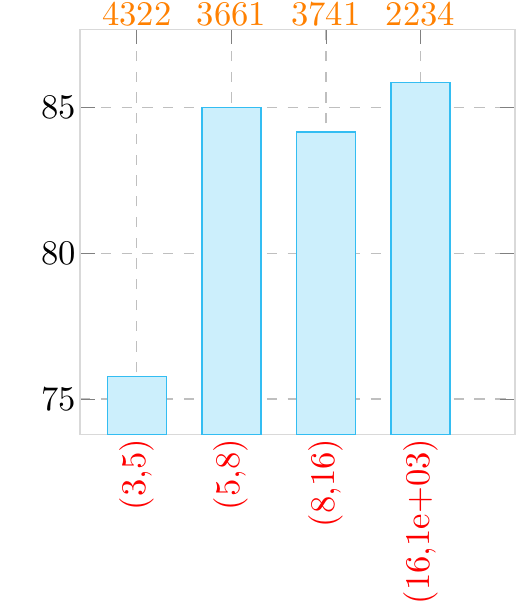}}     &   \multirow{4}[1]{*}{\includegraphics[scale=0.4]{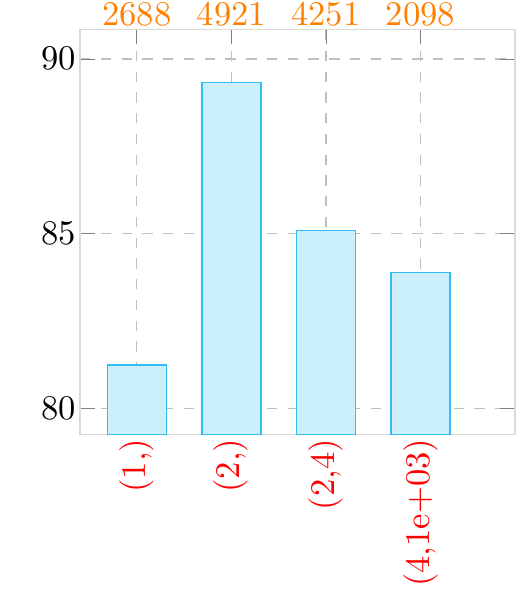}}       &  \multirow{4}[1]{*}{\includegraphics[scale=0.4]{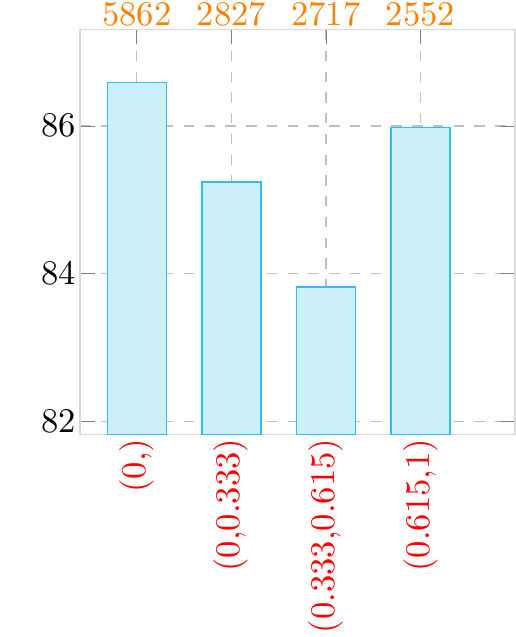}}       &  \multirow{4}[1]{*}{\includegraphics[scale=0.4]{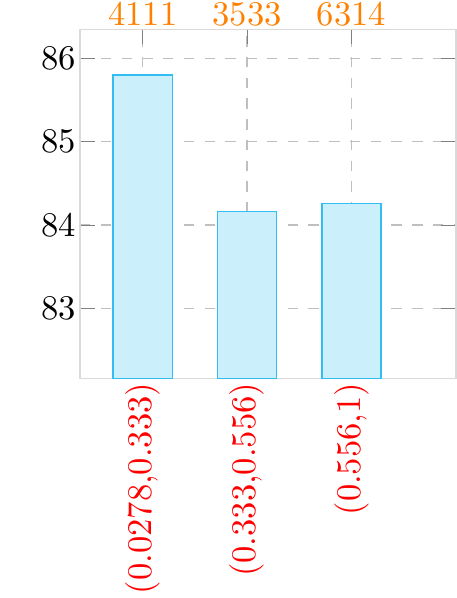}}    &  \multirow{4}[1]{*}{\includegraphics[scale=0.4]{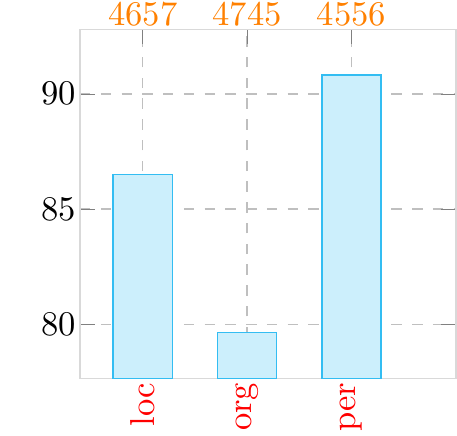}}     
  \\ \\ 
  \multirow{2}[1]{*}{F1: 85.66}   &&&&
  \\ \\ \\ \\ \\
  \midrule
%     \multicolumn{6}{l}{Sino-Tibetan} \\ 
%     \midrule
%     \multirow{3}[1]{*}{zh}    &  \multirow{5}[1]{*}{\includegraphics[scale=0.4]{fig/panx/baidu/zh-f1-sLen.pdf}}     &   \multirow{5}[1]{*}{\includegraphics[scale=0.4]{fig/panx/baidu/zh-f1-eLen.pdf}}       &  \multirow{5}[1]{*}{\includegraphics[scale=0.4]{fig/panx/baidu/zh-f1-rPos.pdf}}       &  \multirow{5}[1]{*}{\includegraphics[scale=0.4]{fig/panx/baidu/zh-f1-eDen.pdf}}    &  \multirow{5}[1]{*}{\includegraphics[scale=0.4]{fig/panx/baidu/zh-f1-tag.pdf}}     
%   \\ \\ \\
%   \\ \\ \\ \\
%   \midrule
    \multicolumn{6}{l}{IE: Romance} \\ 
    %  &&&&&
    \midrule
  \multirow{3}[1]{*}{es}    &  \multirow{5}[1]{*}{\includegraphics[scale=0.4]{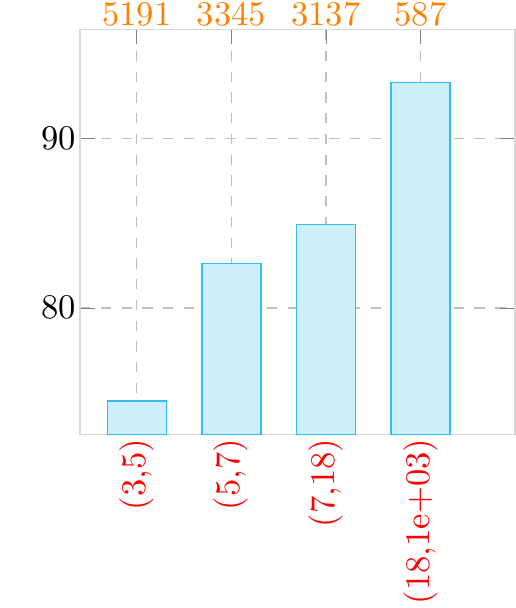}}     &   \multirow{5}[1]{*}{\includegraphics[scale=0.4]{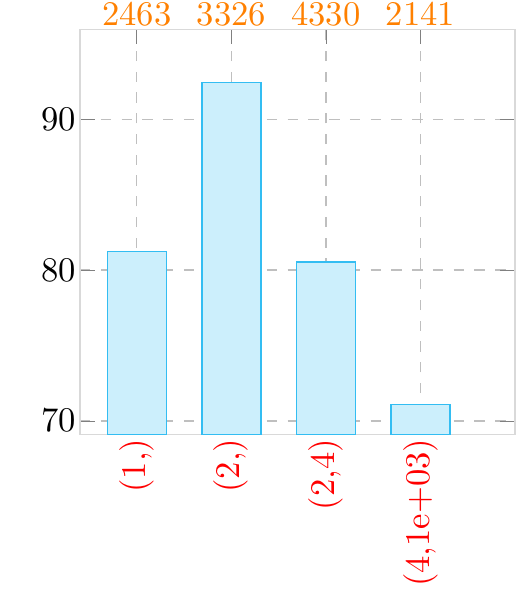}}       &  \multirow{5}[1]{*}{\includegraphics[scale=0.4]{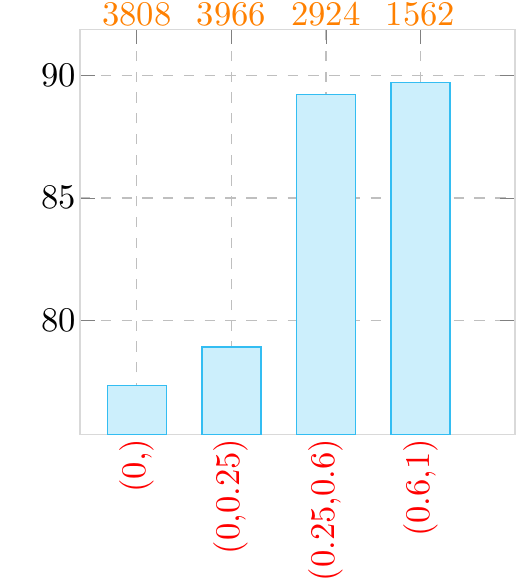}}       &  \multirow{5}[1]{*}{\includegraphics[scale=0.4]{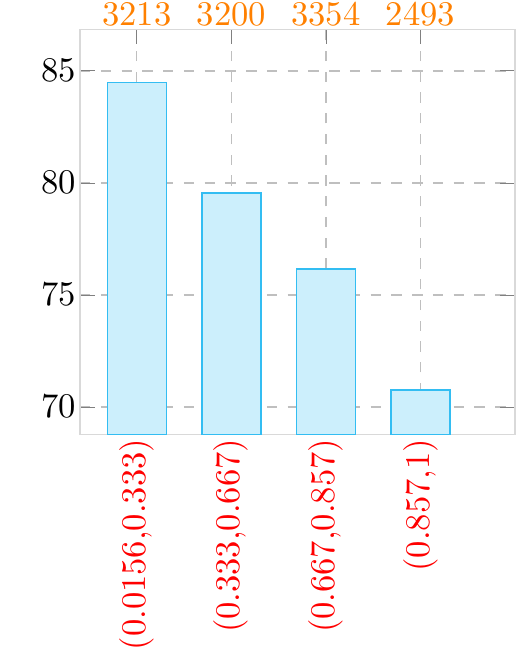}}    &  \multirow{5}[1]{*}{\includegraphics[scale=0.4]{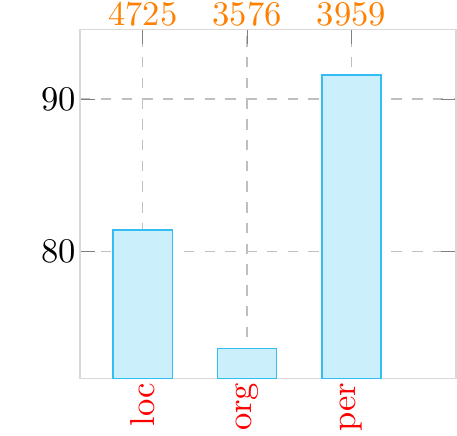}}     
  \\ \\ 
  \multirow{2}[1]{*}{F1: 82.25}   &&&&
  \\
  \\ \\ \\ \\
  \midrule
     \multirow{3}[1]{*}{fr}    &  \multirow{5}[1]{*}{\includegraphics[scale=0.4]{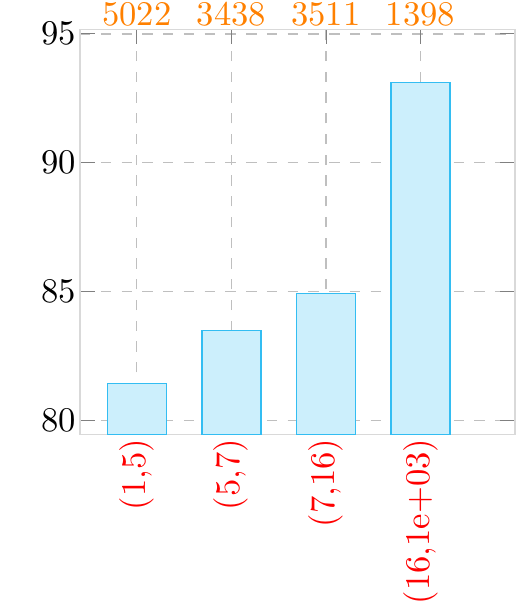}}     &   \multirow{5}[1]{*}{\includegraphics[scale=0.4]{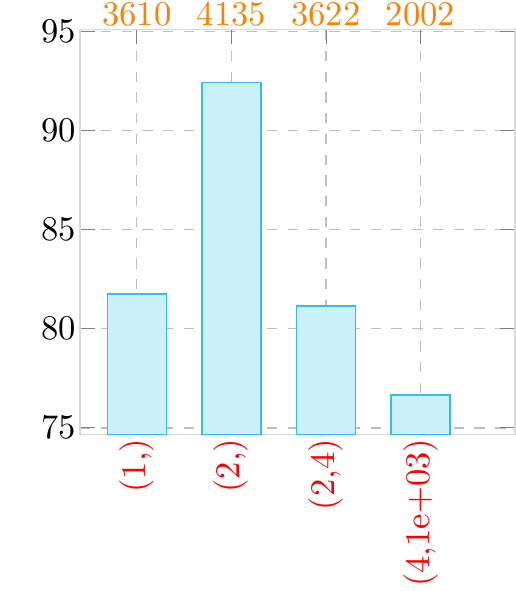}}       &  \multirow{5}[1]{*}{\includegraphics[scale=0.4]{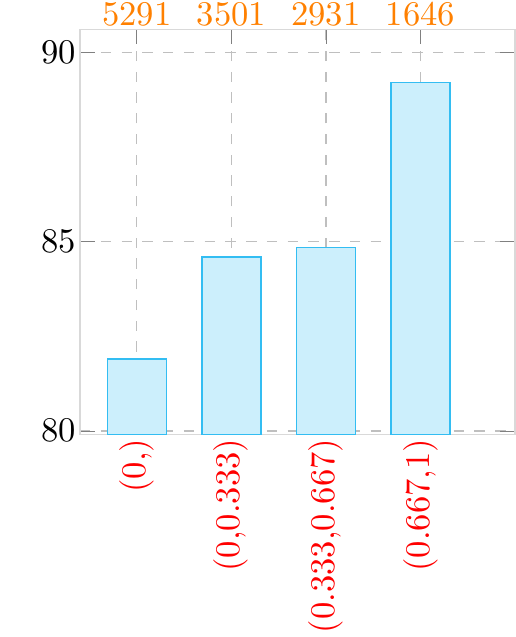}}       &  \multirow{5}[1]{*}{\includegraphics[scale=0.4]{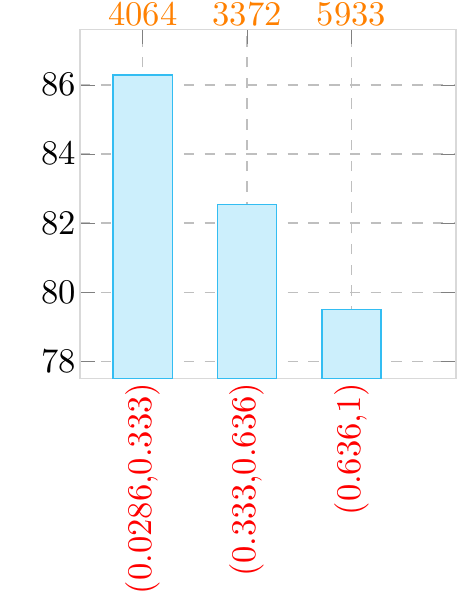}}    &  \multirow{5}[1]{*}{\includegraphics[scale=0.4]{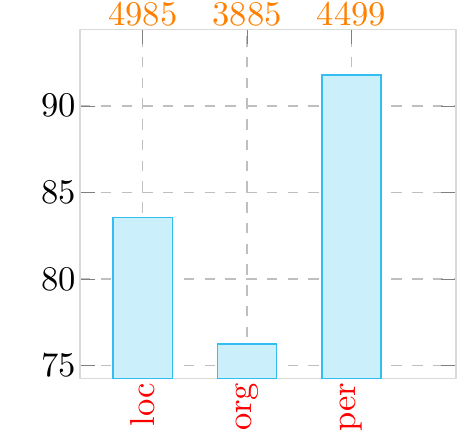}}     
  \\ \\ 
  \multirow{2}[1]{*}{F1: 84.12}   &&&&
  \\
  \\ \\ \\ \\
    \bottomrule
    \end{tabular}%
  \caption{The single system  diagnosis of ERNIE-M on WikiANN-NER task in \texttt{en}, \texttt{es} and \texttt{fr}. The first row shows the fine-grained performance on English test data, and the second and third rows show the zero-shot transfer fine-grained performance on languages in the IE: Romance language family.   }
  \label{tab:ner_ernie_M}
\end{table*}%

\begin{table*}[!ht]
  \centering
  \renewcommand\tabcolsep{0.75pt}
\renewcommand\arraystretch{0.8}
    \begin{tabular}{crrrrr}
    \toprule
    Lang. & \multicolumn{1}{c}{sLen} & \multicolumn{1}{c}{eLen} & \multicolumn{1}{c}{rPos} & \multicolumn{1}{c}{eDen} & \multicolumn{1}{c}{tag}  \\
    \midrule
    \multicolumn{6}{l}{English} \\ 
    \midrule
    \multirow{3}[1]{*}{en}    &  \multirow{5}[1]{*}{\includegraphics[scale=0.4]{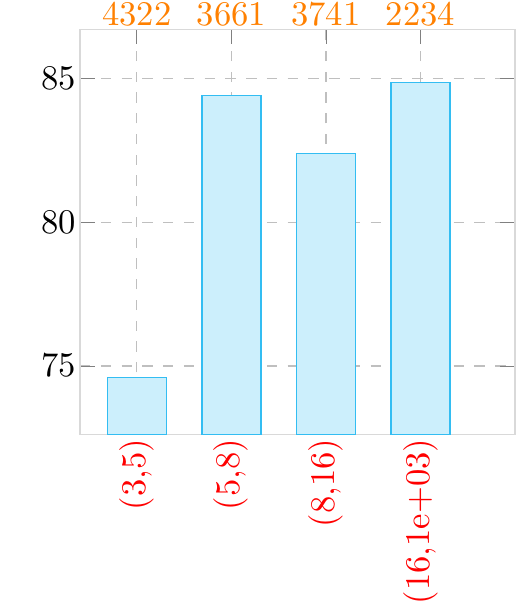}}     &   \multirow{5}[1]{*}{\includegraphics[scale=0.4]{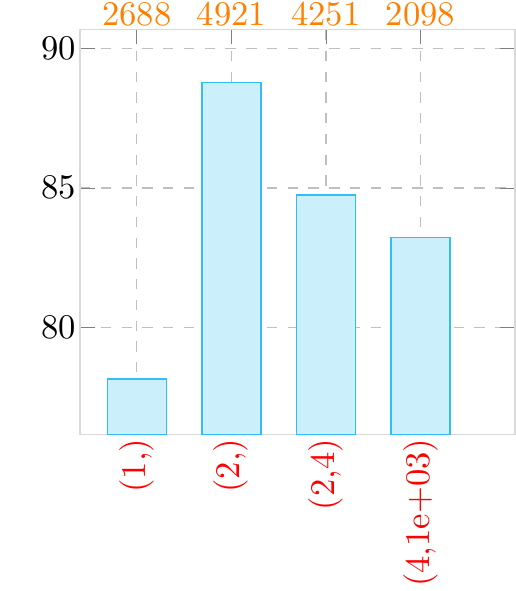}}       &  \multirow{5}[1]{*}{\includegraphics[scale=0.4]{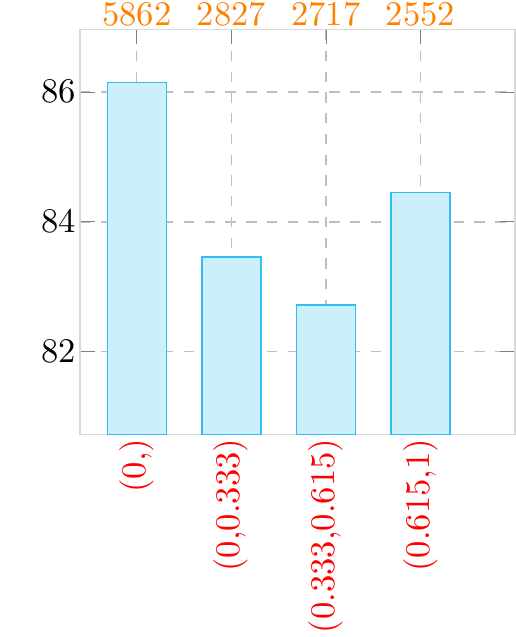}}       &  \multirow{5}[1]{*}{\includegraphics[scale=0.4]{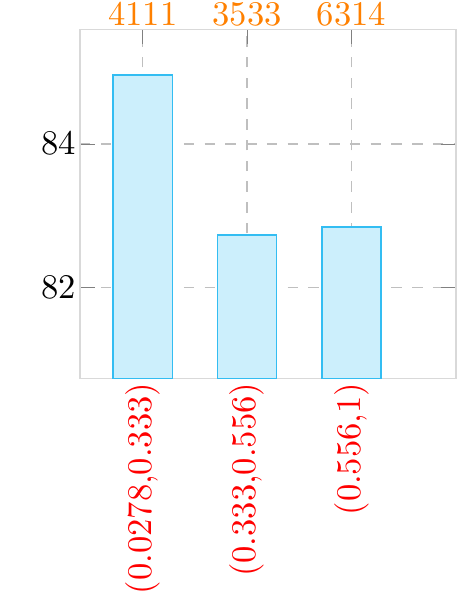}}    &  \multirow{5}[1]{*}{\includegraphics[scale=0.4]{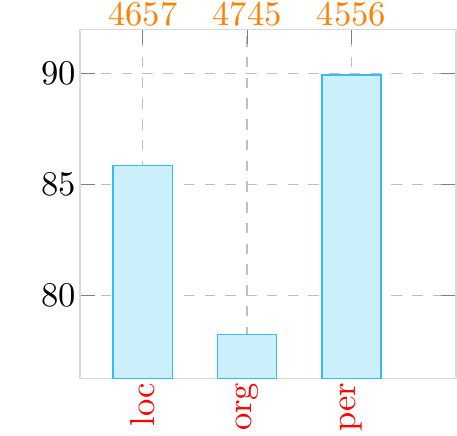}}     
  \\ \\ 
  \multirow{2}[1]{*}{F1: 84.62}   &&&&
  \\
  \\ \\ \\ \\
  \midrule
%     \multicolumn{6}{l}{Sino-Tibetan} \\ 
%     \midrule
%     \multirow{3}[1]{*}{zh}    &  \multirow{5}[1]{*}{\includegraphics[scale=0.4]{fig/panx/XLMR/zh-f1-sLen.pdf}}     &   \multirow{5}[1]{*}{\includegraphics[scale=0.4]{fig/panx/XLMR/zh-f1-eLen.pdf}}       &  \multirow{5}[1]{*}{\includegraphics[scale=0.4]{fig/panx/XLMR/zh-f1-rPos.pdf}}       &  \multirow{5}[1]{*}{\includegraphics[scale=0.4]{fig/panx/XLMR/zh-f1-eDen.pdf}}    &  \multirow{5}[1]{*}{\includegraphics[scale=0.4]{fig/panx/XLMR/zh-f1-tag.pdf}}     
%   \\ \\ \\
%   \\ \\ \\ \\
%   \midrule
    \multicolumn{6}{l}{IE: Romance} \\ 
    %  &&&&&
    \midrule
  \multirow{3}[1]{*}{es}    &  \multirow{5}[1]{*}{\includegraphics[scale=0.4]{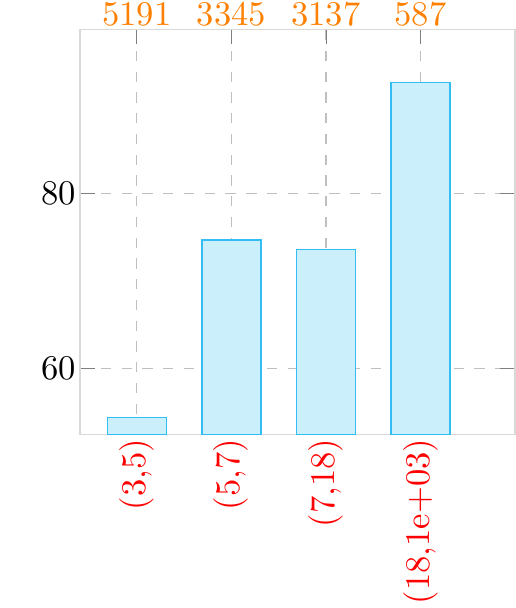}}     &   \multirow{5}[1]{*}{\includegraphics[scale=0.4]{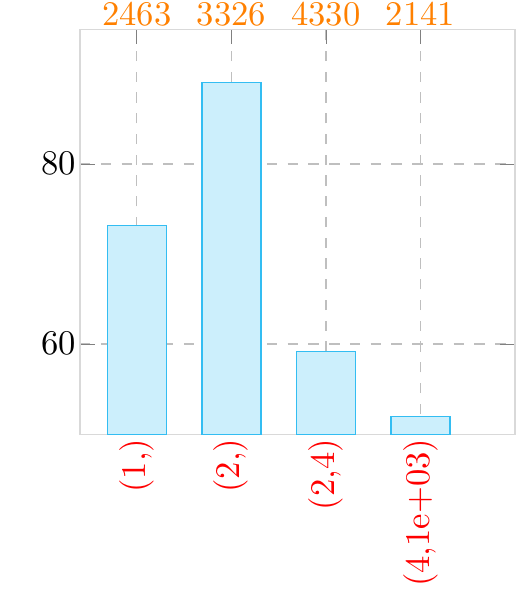}}       &  \multirow{5}[1]{*}{\includegraphics[scale=0.4]{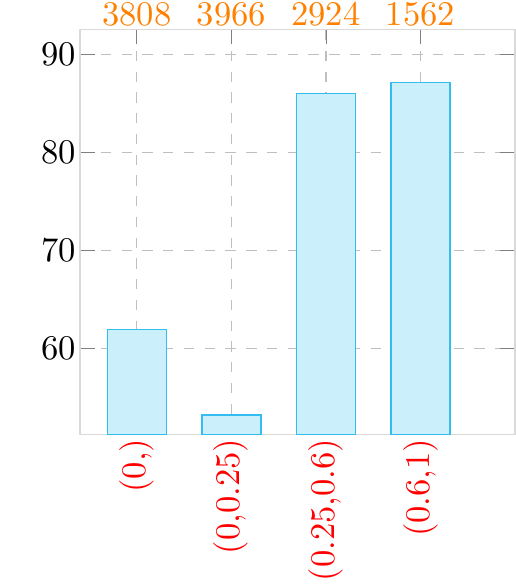}}       &  \multirow{5}[1]{*}{\includegraphics[scale=0.4]{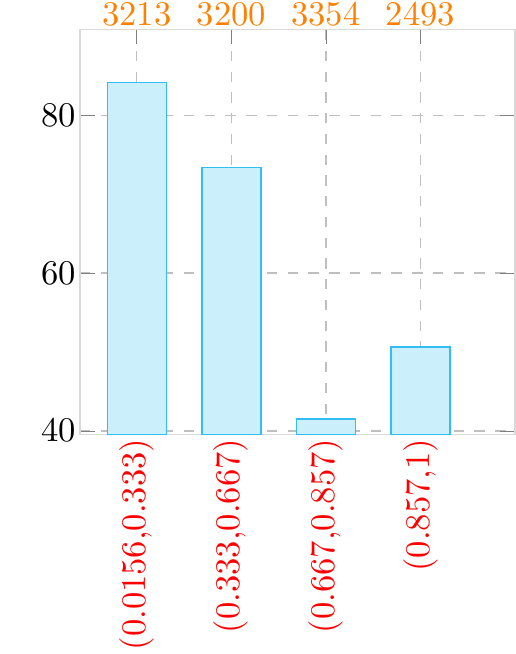}}    &  \multirow{5}[1]{*}{\includegraphics[scale=0.4]{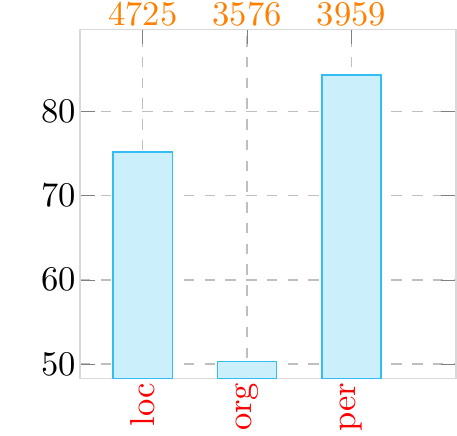}}     
  \\ \\ 
  \multirow{2}[1]{*}{F1: 68.76}   &&&&
  \\
  \\ \\ \\ \\
  \midrule
     \multirow{3}[1]{*}{fr}    &  \multirow{5}[1]{*}{\includegraphics[scale=0.4]{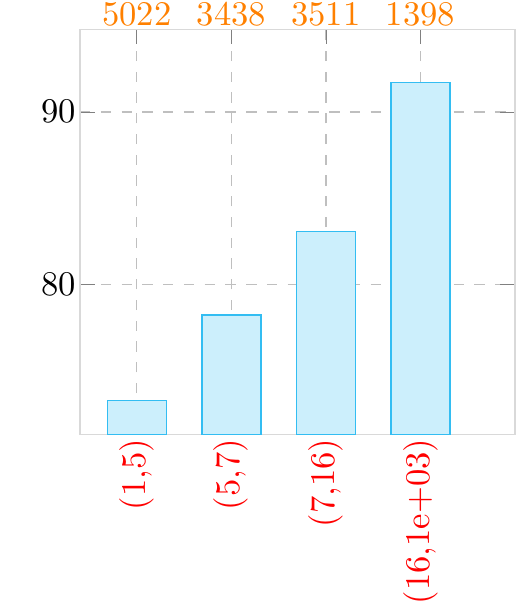}}     &   \multirow{5}[1]{*}{\includegraphics[scale=0.4]{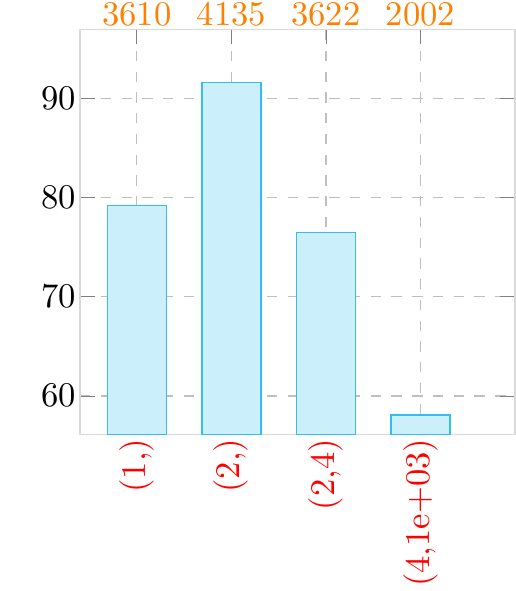}}       &  \multirow{5}[1]{*}{\includegraphics[scale=0.4]{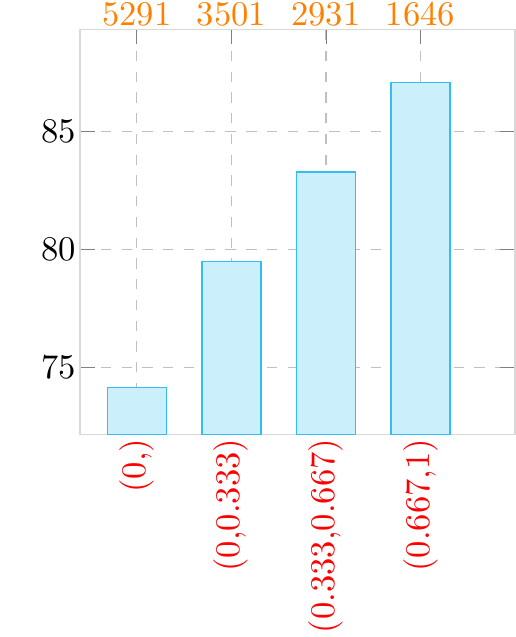}}       &  \multirow{5}[1]{*}{\includegraphics[scale=0.4]{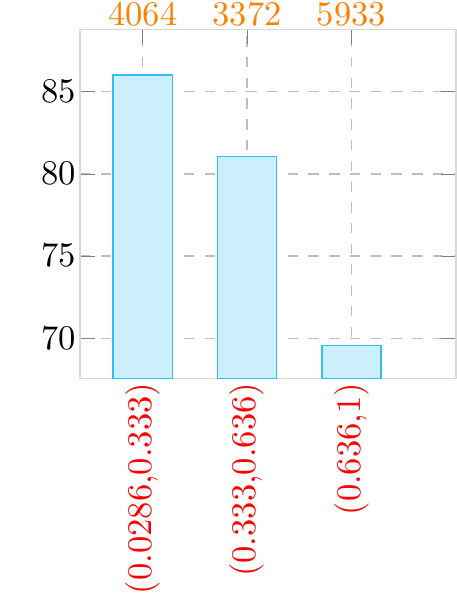}}    &  \multirow{5}[1]{*}{\includegraphics[scale=0.4]{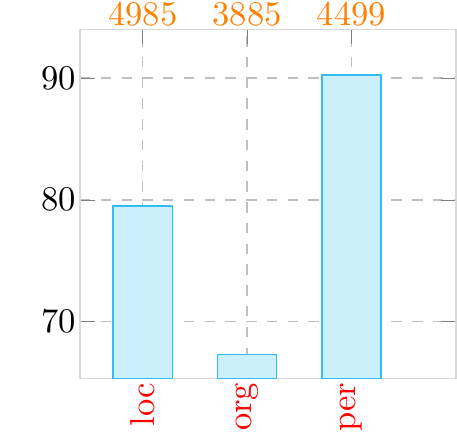}}     
  \\ \\ 
  \multirow{2}[1]{*}{F1: 79.07}   &&&&
  \\
  \\ \\ \\ \\
    \bottomrule
    \end{tabular}%
    \caption{The single system  diagnosis of XLM-R on WikiANN-NER task in \texttt{en}, \texttt{es} and \texttt{fr}.  The first row shows the fine-grained performance on English test data, and the second and third rows show the zero-shot transfer fine-grained performance on languages in the IE: Romance language family. } 
     \label{tab:ner_xlm_r}
\end{table*}%

\subsubsection{QA}
Table~\ref{tab:bucket-wise-appendix} shows the pairwise system analysis of {ERNIE-M} and {T-URLv2} for the XQuAD task. 
We find that although the overall performance of {T-URLv2} outperforms {ERNIE-M}, it is surpassed by {ERNIE-M} on a few buckets. 
For example, in \texttt{zh}, {ERNIE-M} is better at dealing with samples that have long answers, long questions, and a high lexical overlap between questions and answers.  
In \texttt{ru}, {ERNIE-M} is better at dealing with samples with long answers, long questions, and lower lexical overlap between questions and answers, questions and contexts.

% In summary, {ERNIE-M} surpassed {T-URLv2} on those samples with the higher lexical similarity between question and answer, higher lexical similarity between question and context, short context, long question, long answer, which holds for more than 5/8 languages.

\begin{table*}[!htb]
  \centering \tiny
    % \addtolength{\tabcolsep}{-1pt} 
    % \setlength{\tabcolsep}{-1pt}
    \renewcommand\tabcolsep{0.72pt}
\renewcommand\arraystretch{0.77}  
    \begin{tabular}{cccccccc cccccccc cccccccc cccccccc cccccccc cccccccc cccccccc cccccccc cccccc cc cccccccc cccccccc ccccccc cccccccccccc} 
    \toprule
          & \multicolumn{17}{c}{\texttt{aLen}}                           & \multicolumn{17}{c}{\texttt{qLen}}                            & \multicolumn{17}{c}{\texttt{cLen}}                             &
          \multicolumn{17}{c}{\texttt{BLEU-AQ}}  &
          \multicolumn{17}{c}{\texttt{BLEu-QC}}  & \multicolumn{17}{c}{\texttt{qType}}\\
    \midrule
    % \cmidrule(lr){2-107}
    % \cmidrule(lr){1-1}
    & &&&&&&& 
    en & zh & hi & el & ru &  tr & ar &  vi &
    % &&&&&&&
     &&&&&&& && 
      en & zh & hi & el & ru &  tr & ar &  vi &
    %  \multicolumn{1}{l}{\rotatebox{90}{en}} & 
    % \multicolumn{1}{l}{\rotatebox{90}{es}} & 
    % \multicolumn{1}{l}{\rotatebox{90}{de}} & 
    % \multicolumn{1}{l}{\rotatebox{90}{el}} & 
    % \multicolumn{1}{l}{\rotatebox{90}{ru}} & 
    % \multicolumn{1}{l}{\rotatebox{90}{tr}} & 
    % \multicolumn{1}{l}{\rotatebox{90}{ar}} & 
    % \multicolumn{1}{l}{\rotatebox{90}{vi}} &
    % &&&&&&&
    &&&&&&& &&
  en & zh & hi & el & ru &  tr & ar &  vi &
    &&&&&&& &&
     en & zh & hi & el & ru &  tr & ar &  vi &
    &&&&&&& &&
     en & zh & hi & el & ru &  tr & ar &  vi &
    &&&&&&& &&
    en & zh & hi & el & ru &  tr & ar &  vi &
    \\
    \midrule
    & \multicolumn{17}{l}{\multirow{12}[2]{*}{\includegraphics[scale=0.28]{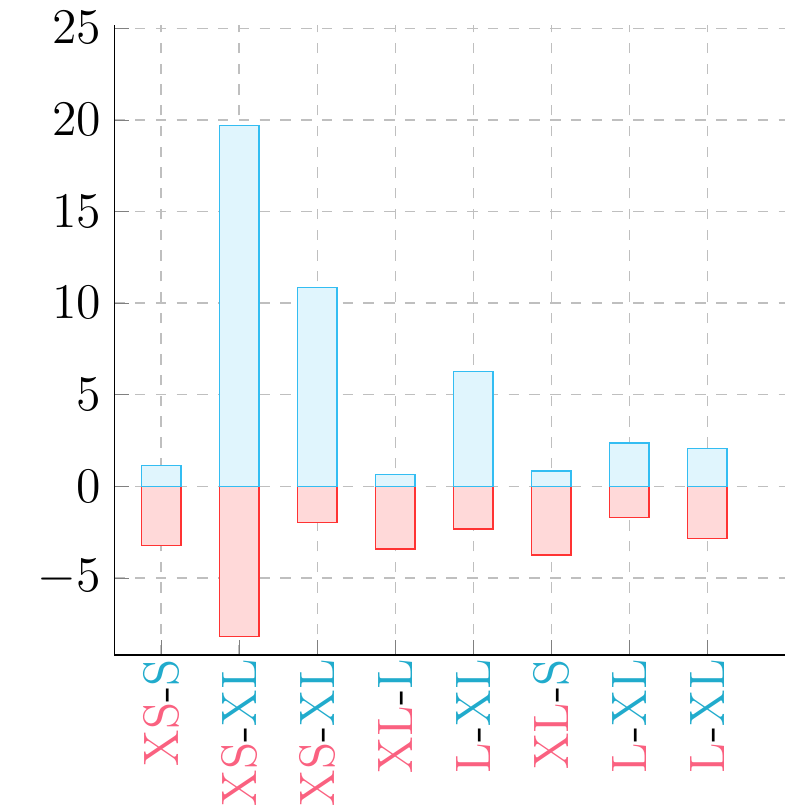}}}              
    & \multicolumn{17}{l}{\multirow{12}[2]{*}{\includegraphics[scale=0.28]{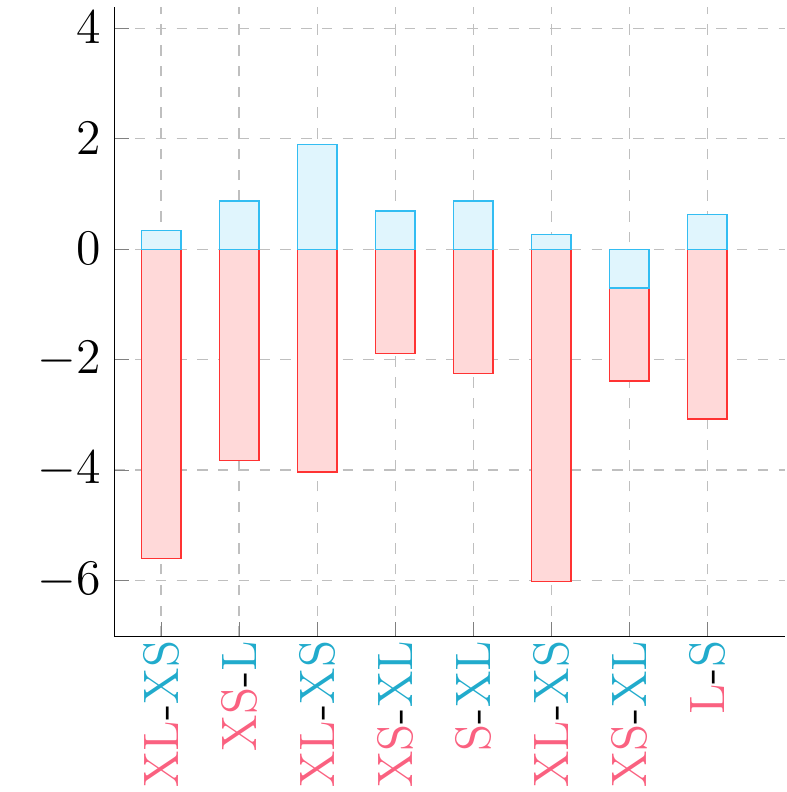}}}              
    & \multicolumn{17}{l}{\multirow{12}[2]{*}{\includegraphics[scale=0.28]{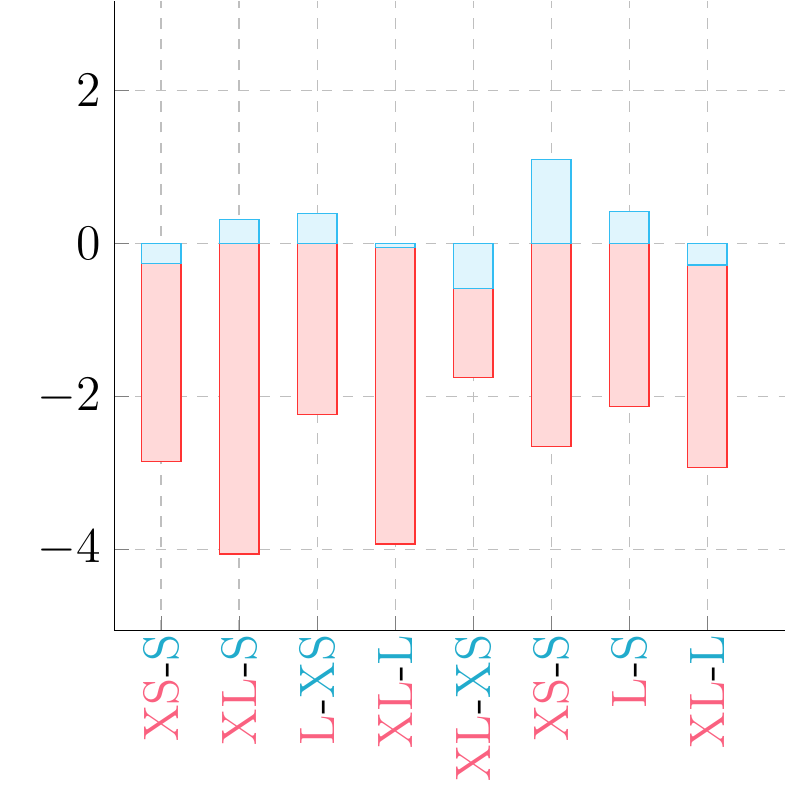}}}              
    &\multicolumn{17}{l}{\multirow{12}[2]{*}{\includegraphics[scale=0.28]{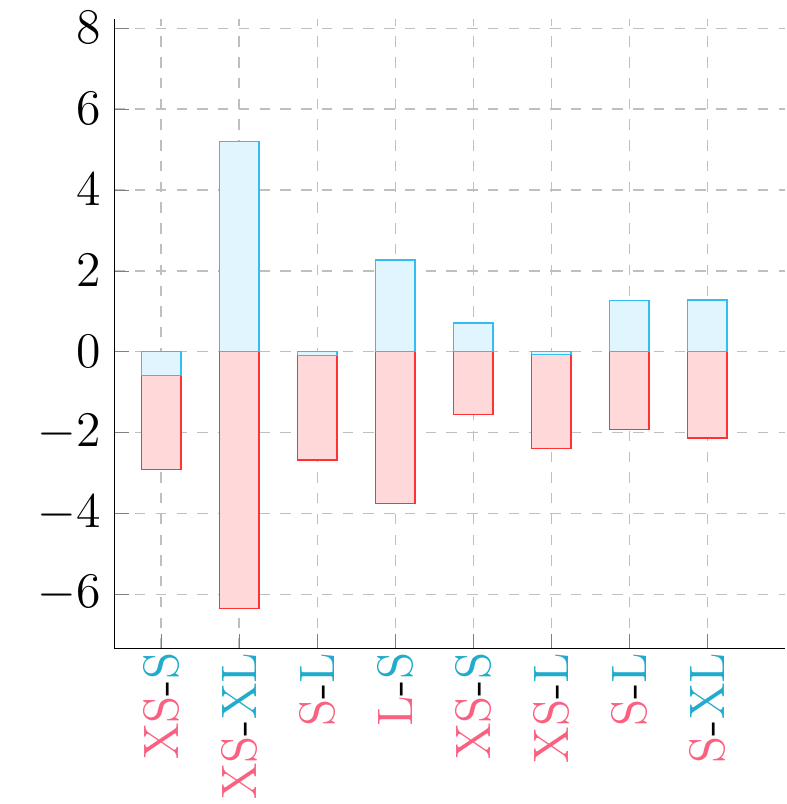}}} 
    &\multicolumn{17}{l}{\multirow{12}[2]{*}{\includegraphics[scale=0.28]{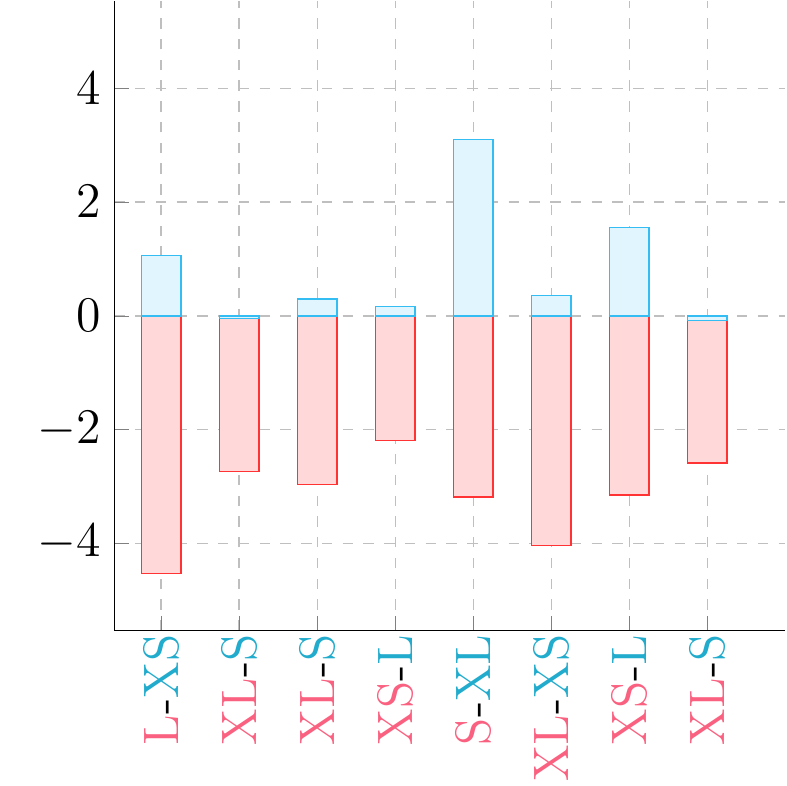}}} 
    & \multicolumn{17}{l}{\multirow{12}[2]{*}{\includegraphics[scale=0.28]{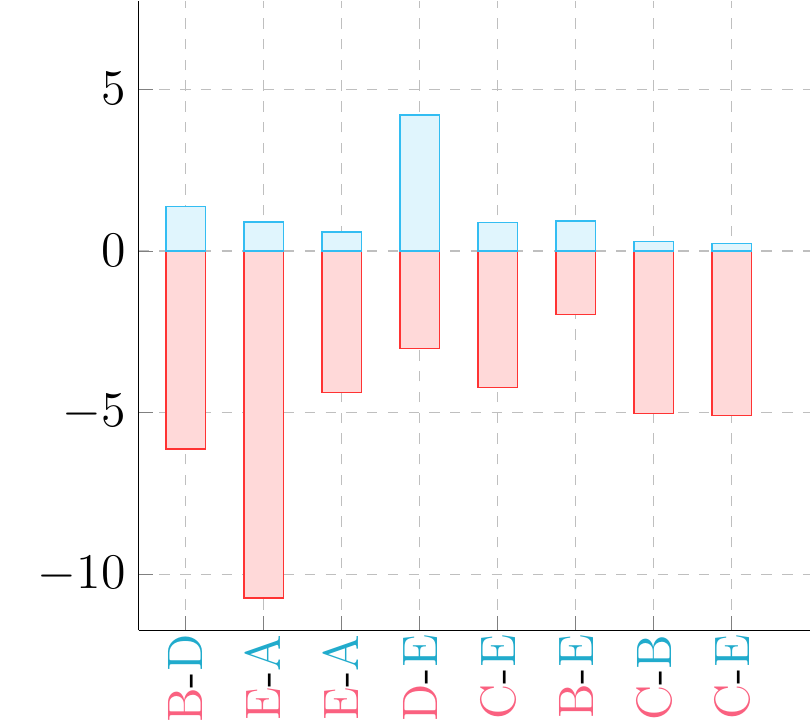}}} 
    \\ \\ 
    \multicolumn{1}{l}{M1: \textit{ERNIE-M}} &  \\
    \multicolumn{1}{l}{(Overall: 75.46)} &  \\ \\
    \multicolumn{1}{l}{M2: \textit{T-URLv2}} &  \\
    \multicolumn{1}{l}{(Overall: 76.68)} &  \\ \\
    \multicolumn{1}{c}{\textbf{Pairwise Sys.}} & 
    \\ 
    \multicolumn{1}{c}{\textbf{(M1$-$M2)}} & 
    \\  \\ \\   

    \bottomrule
    \end{tabular}%
      \caption{Pairwise system diagnosis  of {ERNIE-M} and {T-URLv2} for XQuAD task.  ``M1$-$M2'' represents the performance difference between M1 and M2.
    We classify the attribute values into four \textit{categories}: extra-small (XS), small (S), large (L) and extra-large (XL) values. The top 5 most frequent question types (\texttt{qType}) include what (A), how (B), who (C), when (D) and which (E), ranked by their frequencies in training set. 
In the \textit{pairwise system diagnosis} histogram, \textcolor{ballblue}{blue} (\textcolor{brinkpink}{red}) $x$ ticklabels represents the bucket value of a specific attribute on which system M1 surpasses (under-performs) M2 by the largest margin that is illustrated by a \textcolor{ballblue}{blue} (\textcolor{brinkpink}{red}) bin. The \textcolor{ballblue}{blue}-only $x$ ticklabels (e.g., \textcolor{ballblue}{-D}) indicate that M1 outperforms M2 in all categories of an attribute. 
 }
    \vspace{-5pt} 
  \label{tab:bucket-wise-appendix}%
\end{table*}%

\subsection{\explainaboard Demonstration}
\begin{figure*}
    \centering
    \includegraphics[width=0.75\linewidth]{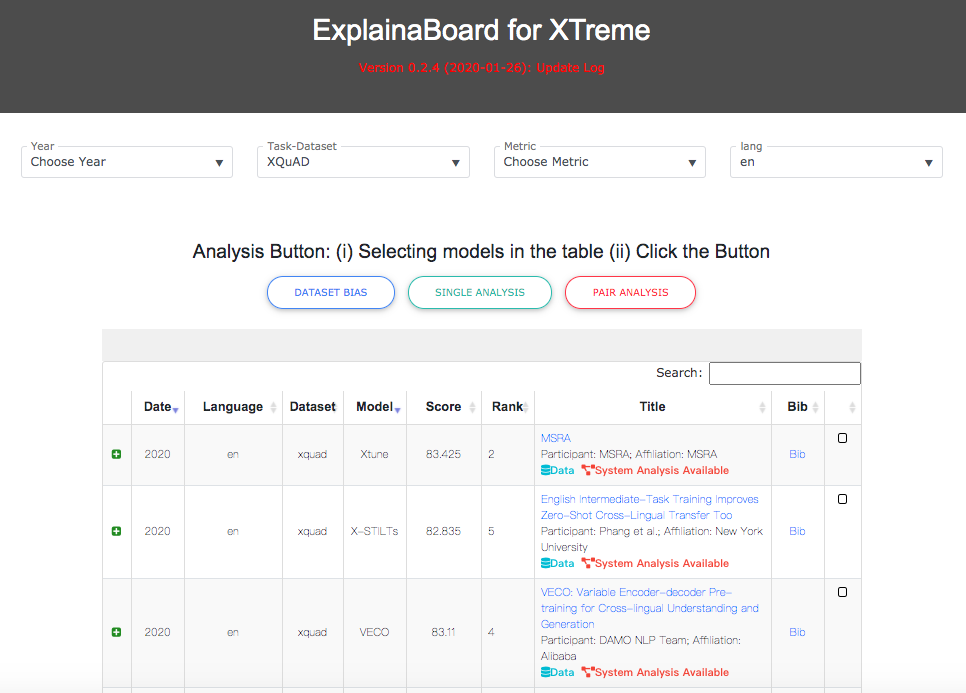}
    \vspace{-6pt}
    \caption{Overall user interface. Four top-down lists at the top are used to filter the entries in the table by the \textit{publication year}, \textit{task}, \textit{metric} and \textit{languages}. Three analysis buttons (e.g., \textit{DATASET BIAS}, \textit{SINGLE ANALYSIS}, \textit{PAIRWISE ANALYSIS}) are used to perform three different fine-grained evaluations. Each row in the table represents the performance of a system on a specific dataset and a specific language. Relevant pieces of information such as the paper title also are provided. }
    \label{fig:intro_nerr1}
\end{figure*}

\begin{figure*}
    \centering
    \includegraphics[width=0.75\linewidth]{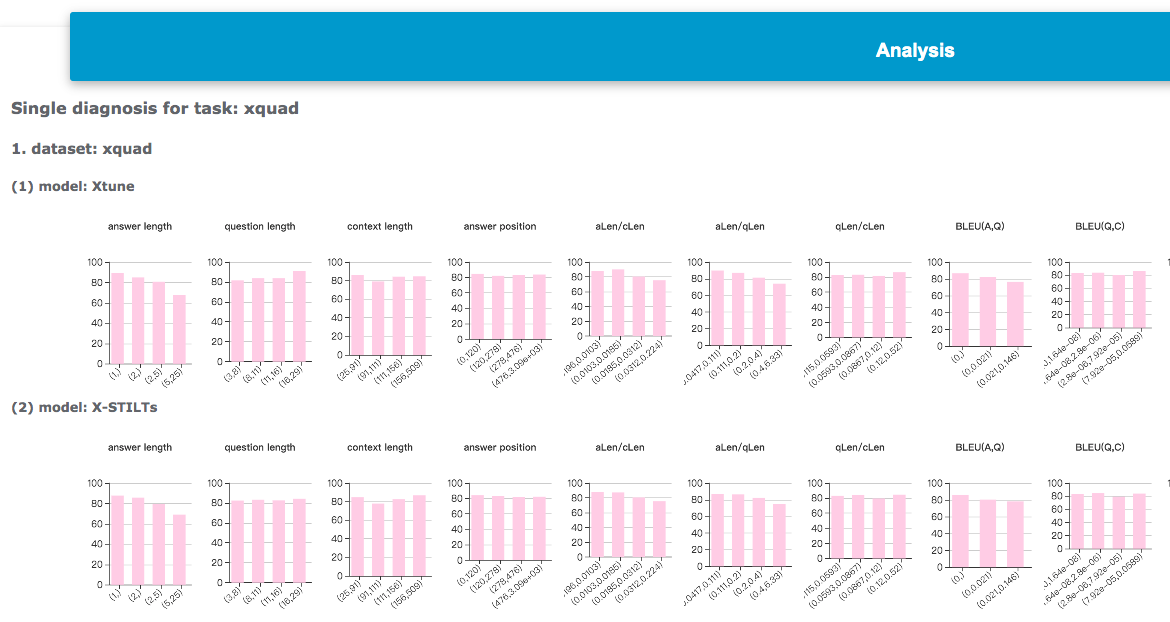}
    \vspace{-6pt}
    \caption{Single system analysis. Each histogram represents the fine-grained evaluation results of a given system, which are broken down based on a pre-defined attribute (e.g., \texttt{answer length} ). }
    \label{fig:intro_nerr2}
\end{figure*}

\begin{figure*}
    \centering
    \includegraphics[width=0.75\linewidth]{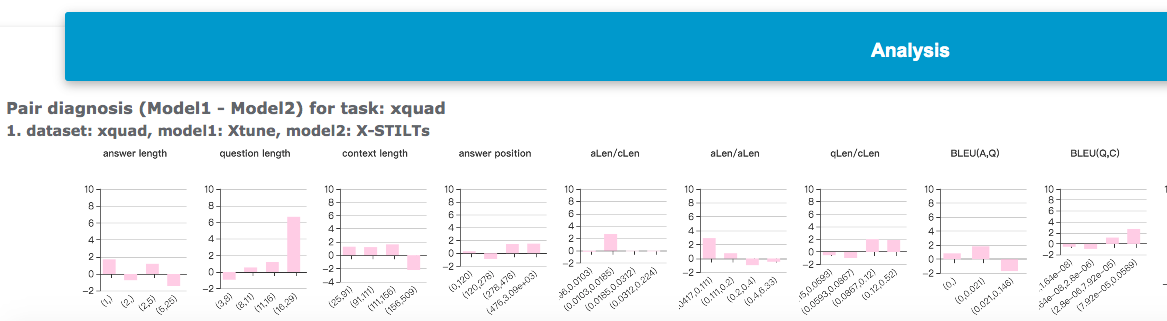}
    \vspace{-6pt}
    \caption{Pairwise system analysis. Each histogram illustrates the fine-grained  performance gap between system 1 and system 2, which has been broken down by different pre-defined attributes (e.g., \texttt{answer length}).}
    \label{fig:intro_nerr3}
\end{figure*}

Figure~\ref{fig:intro_nerr1} shows the interface of \explainaboard containing possible selection options to observe the fine-grained analysis for submitted systems on \xtreme. We also demonstrate how to perform \textit{Single System} and \textit{Pair Systems} analysis on Figure~\ref{fig:intro_nerr2} and \ref{fig:intro_nerr3} respectively. %show the screenshots of \explainaboard for \name that can help users to perform the fine-grained evaluation.

Specifically, to generate a fine-grained overview, we first select models in the table, then click one of the three \textit{Analysis Buttons}, which generates a fine-grained analysis such as in Figure~\ref{fig:intro_nerr2} (\textit{single system analysis}) and Figure~\ref{fig:intro_nerr3} (\textit{pair-wise system analysis}).

\end{document}